%% file: 00main.tex
\documentclass{article}

\PassOptionsToPackage{compress,numbers}{natbib}
\usepackage[preprint]{neurips_2024}
\usepackage{amsmath}

\usepackage{mathrsfs}
\usepackage{yk}
\usepackage{dsfont}

\usepackage{float}
\usepackage{wrapfig}
\usepackage{algorithm}
\usepackage{algpseudocode}
\usepackage[utf8]{inputenc} 
\usepackage[T1]{fontenc}    
\usepackage{hyperref}       
\usepackage{url}            
\usepackage{booktabs}       
\usepackage{amsfonts}       
\usepackage{nicefrac}       
\usepackage{microtype}      
\usepackage{xcolor}         
\usepackage{multirow}
\usepackage[pdftex]{graphicx} 
\usepackage{subcaption}

\newcommand{\blue}[1]{\textcolor{blue}{#1}}

\title{Transfer Learning Under High-Dimensional Graph Convolutional Regression Model for Node Classification}

\author{%
Jiachen Chen$^{1}$ \quad Danyang Huang$^{2}$ \quad Liyuan Wang$^2$ \quad Kathryn L. Lunetta$^1$ \\
\textbf{Debarghya Mukherjee}$^{1}\thanks{Corresponding authors.}$  \quad  \textbf{Huimin Cheng}$^{1}\footnotemark[1]$ \\
$^1$Boston University \quad $^2$Renmin University of China\\
\texttt{\{chenjc,klunetta,mdeb,huimin23\}@bu.edu}\\
\texttt{\{dyhuang,wangly2023\}@ruc.edu.cn}\\
}

\def \logit{\mathrm{logistic}}
\def \bA{\mathbf{A}}
\def \bX{\mathbf{X}}
\def \bY{\mathbf{Y}}
\def \bZ{\mathbf{Z}}
\def \dims{d}
\def \br{\mathbb{R}}
\def \bdelta{\boldsymbol{\delta}}
\def \bu{\boldsymbol{u}}

\def \our{\mbox{GCR}}

\def \degree{\mathbf{D}}
\def \half{\frac{1}{2}}

\def \bS{\mathbf{S}}
\def \mA{\mathcal{A}}

\def \one{\textbf{1}}
\def \zero{\textbf{0}}
\def \s{s}
\def \bbeta{\boldsymbol{\beta}}

\begin{document}

\maketitle

\begin{abstract}
 Node classification is a fundamental task, but obtaining node classification labels can be challenging and expensive in many real-world scenarios.  Transfer learning has emerged as a promising solution to address this challenge by leveraging knowledge from source domains to enhance learning in a target domain.  Existing transfer learning methods for node classification primarily focus on integrating Graph Convolutional Networks (GCNs) with various transfer learning techniques. While these approaches have shown promising results, they often suffer from a lack of theoretical guarantees, restrictive conditions, and high sensitivity to hyperparameter choices. To overcome these limitations, we propose a \underline{G}raph \underline{C}onvolutional Multinomial Logistic \underline{R}egression (GCR) model and a transfer learning method based on the GCR model, called Trans-GCR.
We provide theoretical guarantees of the estimate obtained under $\our$  model  in high-dimensional
settings. 
Moreover, Trans-GCR demonstrates superior empirical performance, has a low computational cost, and requires fewer hyperparameters than existing methods.
 
\end{abstract}

\section{Introduction}

Network (a.k.a graph) data is ubiquitous in various domains, including social networks  \citep{barabasi2013network}, citation networks \citep{ji2022co}, and biological networks \citep{zitnik2018modeling, han2019gcn}. A fundamental task in network analysis is node classification \citep{kipf2016semi}, which aims to predict the class labels of a node based on its own features and its neighboring nodes' features. Usually, the node features are high-dimensional \citep{hamilton2017inductive}. Thus, we focus on high-dimensional settings in this paper. 
Nevertheless, obtaining node classification labels can be challenging and expensive in many real-world scenarios \citep{dai2022graph}. For example, classifying genes into disease categories using a gene-gene interaction network faces a scarcity of disease labels, as experimentally annotating genes is expensive and limited \citep{guney2016network}. 

To address the challenge of limited labeled data, transfer learning, which uses knowledge from source domains to enhance learning in a target domain, has emerged as a promising solution \citep{dai2022graph}.  
Continuing the aforementioned example, despite the scarcity of disease labels, 
abundant functional annotations of genes exist in curated databases like KEGG pathways \citep{kanehisa2012kegg}. 
In the existing literature, 
various transfer learning methods based on Graph Convolutional Networks (GCNs)  \citep{sperduti1997supervised,bruna2013spectral, defferrard2016convolutional,kipf2017semi} have been proposed to enhance node classification accuracy.
These  GCN-based transfer learning methods can be broadly summarized into three main areas. 

 First, pre-training and fine-tuning approaches \citep{hu2019pre, hu2019strategies, lu2021learning, yang2022self,   kooverjee2022investigating, xu2023better},  which usually pre-train a GCN on a large-scale dataset to learn transferable representations and then fine-tune the pre-trained model on a target task. While effective, these methods usually lack theoretical guarantees.
Second, theoretical transferability analysis. Researchers have investigated the theoretical transferability properties of GCNs for graphs sampled from the same underlying space or graphon model \citep{nilsson2020experimental, ruiz2020graphon, levie2021transferability,  ruiz2023transferability}.   Despite valuable theoretical insights, they often assume that the source and target domain
are drawn from exactly the same underlying model, which usually does not hold in practical scenarios where domain shifts occur. 
Third, to address the domain shifts challenge, various domain adaptation techniques have been proposed, such as unsupervised adaptation \citep{wu2020unsupervised}, local structure transfer \citep{zhu2021transfer}, adversarial domain alignment \citep{dai2022graph}, and noise-resistant transfer \citep{yuan2023alex}. Despite promising results,  they often lack theoretical guarantees or can be sensitive to hyperparameter choices.
In summary, existing methods suffer from a lack of theoretical guarantees,  restrictive conditions, and high sensitivity to hyperparameters.




 To address these limitations, we propose a novel statistical transfer learning framework based on a \underline{G}raph \underline{C}onvolutional Multinomial Logistic \underline{R}egression ($\our$) model. The $\our$ model assumes that the classification label depends on the graph-aggregated node features (obtained through multiple graph aggregation layers),  followed by a multinomial logistic regression model.
Our proposed $\our$ model is motivated by the observation in \cite{wu2019simplifying}, which suggests that removing nonlinear activation functions (e.g., ReLU) in GCN's hidden layers achieves comparable performance to the original GCN architecture. 
Under the $\our$ model, we develop a two-step transfer learning method Trans-$\our$. 
Specifically,  we let $\bbeta^{s}$ and $\bbeta^{t}=\bbeta^{s}+\delta$ denote the $\our$'s high-dimensional sparse 
model parameters in source and target data respectively, where $\delta$ measures the domain shift.
In the first step, we obtain the estimate of source domain parameters,  denoted as $\hat{\bbeta}^s$,
by minimizing the $l_1$-regularized negative likelihood function of the $\our$ model using source data.  In the second step, we estimate the shift term $\delta$  by substituting $\bbeta^{t}$ with $\hat{\bbeta}^s + \delta$ in the $\our$ negative likelihood function and minimizing it using the target data. This step leverages the knowledge learned from $\hat{\bbeta}^s$.
Finally, our estimate of the target domain parameters is given by $\hat{\bbeta}^t = \hat{\bbeta}^s + \hat{\delta}$. 

Our method enjoys the following advantages: (1)  We demonstrate through extensive empirical studies that our proposed method achieves superior or comparable performance compared with existing complicated GCN-based transfer learning approaches for node classification.
 (2) 
 We provide theoretical guarantees of the estimate obtained under our proposed $\our$  model  in high-dimensional
settings under mild conditions. 
 (3) By leveraging the simplified model $\our$, our method involves fewer parameters to be trained than more complex GCN-based models, resulting in reduced computational cost. (4) Our framework has only two hyperparameters, i.e., the number of graph aggregation layers and the $l_1-$norm penalty strength.

\section{Graph Convolutional Regression Model}

A graph  with $n$ nodes is represented by an adjacency matrix $\bA=(\bA_{ij}) \in \{0,1\}^{n \times n}$, $i,j=1,\ldots,n$, where $\bA_{ij}=1$  if  there is an edge between nodes $i$ and $j$, and $\bA_{ij}=0$  otherwise. We only consider a graph with no self-loops, so all diagonal entries of $\bA$ are 0.
In addition, each node is associated with a $\dims$-dimensional covariates and a $C$-dimensional one-hot class label. The entire covariate matrix is  $\bX \in \mathbb{R}^{n \times \dims}$, and the entire classification label matrix is  $\bY \in \{0,1\}^{n \times C}$. Node classification aims to predict $\bY$ based on $\bX$ and $\bA$. The normalized adjacency matrix is 
\begin{equation}\label{eq:norm_adj}
    \bS =  \tilde{\degree}^{-\half} \tilde{\bA} \tilde{\degree}^{-\half}, \mbox{ where }  \tilde{\bA}=\bA+\mathbf{I}_n,
\end{equation}
where $\mathbf{I}_n$ is an $n \times n$ identity matrix, $\tilde{\bA}$ denote the adjacency matrix with added self-connections. 
 Here,  $\tilde{\degree}$ is the degree matrix of $\tilde{\bA}$, with diagonal entry  $\tilde{\degree}_{ii}$ representing the degree of node $i$, and all off-diagonal elements being zero.



\subsection{Related Work: Graph Convolutional Networks}
GCNs and their variants have gained increasing popularity for node classification tasks \citep{hamilton2017inductive, kipf2017semi}. Among these variants, Simple Graph Convolution (SGC) \citep{wu2019simplifying} is a popular and powerful simplified variant of  GCNs.
The main idea behind SGC is to remove the nonlinear activation functions in the hidden layers of a GCN and only keep the softmax activation in the final layer to obtain probabilistic outputs. 
An $M$-layer SGC trains a classifier 
    $\hat{\bY} = \mbox{Softmax}(\bS^M \bX \hat{\bbeta})$, 
where $\hat{\bbeta} \in \br^{d \times C}$ is the learned weight matrix.  SGC enjoys the following advantages: (1) SGC reduces model complexity and computational burden, making SGC more efficient and scalable.
(2) SGC has only one hyperparameter, the number of layers $M$, reducing tuning complexity and hyperparameter sensitivity and thus preventing overfitting.
(3) Despite the simplicity, extensive experiments have shown that SGC maintains competitive or even slightly better performance than GCN \citep{wu2019simplifying}.


\def \Multi{\mbox{Multi}}
\def \bP{\mathbf{P}}

\subsection{High-Dimensional Graph Convolutional Multinomial Logistic Regression Model}
Motivated by the success of SGC, we propose the $\our$ model to model the relationship between $\bY$, $\bX$, and $\bA$.  $\our$ assumes that the classification labels are not only affected by their own, but also their neighbors' features based on the graph structure captured by the adjacency matrix. Usually, a normalized adjacency matrix is used; a common choice is  $\bS$ in Eq. \ref{eq:norm_adj}. In what follows, we present the $\our$ model using $\bS$. 
In $\our$, there is a  vector of coefficients $\bbeta_c=(\bbeta_{1c},\ldots,\bbeta_{\dims c})^T$ for each category $c=1, \ldots, C$. To ensure the identifiability of $\bbeta_c$, 
we take the last category as the reference category, i.e., $\bbeta_C=\zero_\dims$. 
In $\our$ model, given predictors $\bX$  and  $\bA$, for each node $i = 1 \ldots, n $, the probability of node $i$'s classification label 
$\bY_i$ belonging to category $c$  is 
\begin{equation}
\textstyle
   \bP_{ic} = \mathbb{P}(\bY_{i} = c| \bX, \bA )  = \frac{\exp^{ \sum_{j=1}^n\bS^M_{ij}\bX_j \bbeta_c}}{1+\sum_{c=1}^{C-1} \exp^{\sum_{j=1}^n\bS^M_{ij}\bX_j\bbeta_c } }, c=1,\ldots, C, 
\end{equation}\label{eq:GCMLR}
where $\bX_j=(\bX_{j1}, \ldots,\bX_{j\dims})$ is the $j$th row of $\bX$, 
$\bS^M_{ij}$ is the $(i,j)$th element in $\bS^M$, $\bS$ is defined in Eq. \ref{eq:norm_adj},  and $M$ is the number of convolution layer. The sum $\sum_{j=1}^n\bS^M_{ij}\bX_j$ aggregates neighboring nodal features, encouraging neighboring nodes to have similar aggregated features, thus having similar classification labels. A larger value of $M$ enables the model to capture higher-order neighborhood dependencies.
In the high-dimensional setting, where $\dims$ can be larger than $n$, we assume $\bbeta_c$ to be sparse such that the number of nonzero elements of $\bbeta_c$, denoted by $s <<\dims$.

\textbf{{{Parameter estimation. }}}  Given observed data $(\bA,\bX,\bY)$,   the parameter estimate $\hat{\bbeta}=(\hat{\bbeta_1}, 
\ldots, \hat{\bbeta}_{C-1} )$ is obtained by  minimizing the following negative log-likelihood with $l_1$ regularization, 
$l(\bbeta; \bY, \bX, \bA) = 
   - \sum_{i=1}^{n} \sum_{c=1}^{C} (\bY_{ic} \log \bP_{ic} + 
   (1-\bY_{ic}) \log(1-\bP_{ic} )
   ) +
   \lambda \sum_{c=1}^{C-1} || \bbeta_c||_1,$
where  $\bP_{ic}$ is defined in Eq. \ref{eq:GCMLR}, 
$\bbeta_c$ is the $c$th column of $\bbeta$,
$||\cdot||_1$ is the $l_1-$norm, $\lambda \geq 0$ is a regularization hyperparameter controlling the trade-off between the log-likelihood and the penalty.
 $l(\bbeta; \bY, \bX, \bA)$ can be further simplified as 
\begin{equation}
\label{eq:mle}
\textstyle
l(\bbeta; \bY, \bX, \bA) = 
   - \sum_{i=1}^{n} \sum_{c=1}^{C} (\bY_{ic} (\sum_{j=1}^n \bS_{ij}^M \bX_j) \bbeta_c  - \log(1+e^{\sum_{j=1}^n \bS_{ij}^M \bX_j \bbeta_c} ) ) +
   \lambda \sum_{c=1}^{C-1} || \bbeta_c||_1
\end{equation}
To minimize Eq. \ref{eq:mle}, we use the standard coordinate descent algorithm \citep{wright2015coordinate}, which iteratively minimizes the objective function with respect to each coordinate of $\bbeta$, while keeping the others fixed.

\section{Transfer Learning}
\label{sec:method}

In this paper, we consider the following transfer learning problem. Let $(\bA^{(0)}, \bX^{(0)},\bY^{(0)})$ denote the target data,  where  $\bA^{(0)} \in \{0,1\}^{n_0 \times n_0}$, $\bX^{(0)} \in \mathbb{R}^{n_0 \times \dims}$,  $\bY^{(0)} \in \{0,1\}^{n_0 \times C}$.  Let  $\bbeta^{(0)} = (\bbeta_1^{(0)}, \ldots, \bbeta_{C-1}^{(0)}) \in \mathbb{R}^{\dims \times (C-1)}$  denote the  true coefficient matrix associated with target data  under $\our$ model.
Let $(\bA^{(k)}, \bX^{(k)},\bY^{(k)})$ denote the $k$th source data, 
 $k=1,\ldots, K$,  where  $\bA^{(k)} \in \{0,1\}^{n_k \times n_k}$, $\bX^{(k)} \in \mathbb{R}^{n_k \times \dims}$,  $\bY^{(k)} \in \{0,1\}^{n_k \times C}$.  
 Let $\bbeta^{(k)} = (\bbeta_1^{(k)}, \ldots, \bbeta_{C-1}^{(k)}) \in \mathbb{R}^{\dims \times (C-1)}$ denote the  true coefficient matrix   associated with the $k$th source data.
 
The difference between the  $k$th source domain's coefficient and the target domain's coefficient is $\delta^{(k)} = \bbeta^{(0)} - \bbeta^{(k)}$. Let $(C-1)^{-1}\sum_{c=1}^{C-1} ||\delta_c^{(k)}||_1$ measure the domain shift level of the $k$th source data, where  $\delta_c^{(k)}$ is the $c$th column of $\delta^{(k)}$. A source sample is defined as $h-$level transferable if its domain shift level is lower than a threshold $h$.  The set of $h-$level transferable source data is $\mA_h = \{k :  (C-1)^{-1}\sum_{c=1}^{C-1} ||\delta_c^{(k)}||_1 \leq h \}$.  
 To ensure that transferring sources within the set $\mA_h$ is beneficial, $h$ should be reasonably small.
In the subsequent sections, we will abbreviate the notation $\mA_h$ as $\mathcal{A}$ for brevity without special emphasis.


\def \alg{\mbox{Trans-$\our$}}
\subsection{Transfer Learning When the Transferable Source Set is Known}\label{sec:method1}

\begin{figure} 
    \centering
    \includegraphics[scale=0.29]{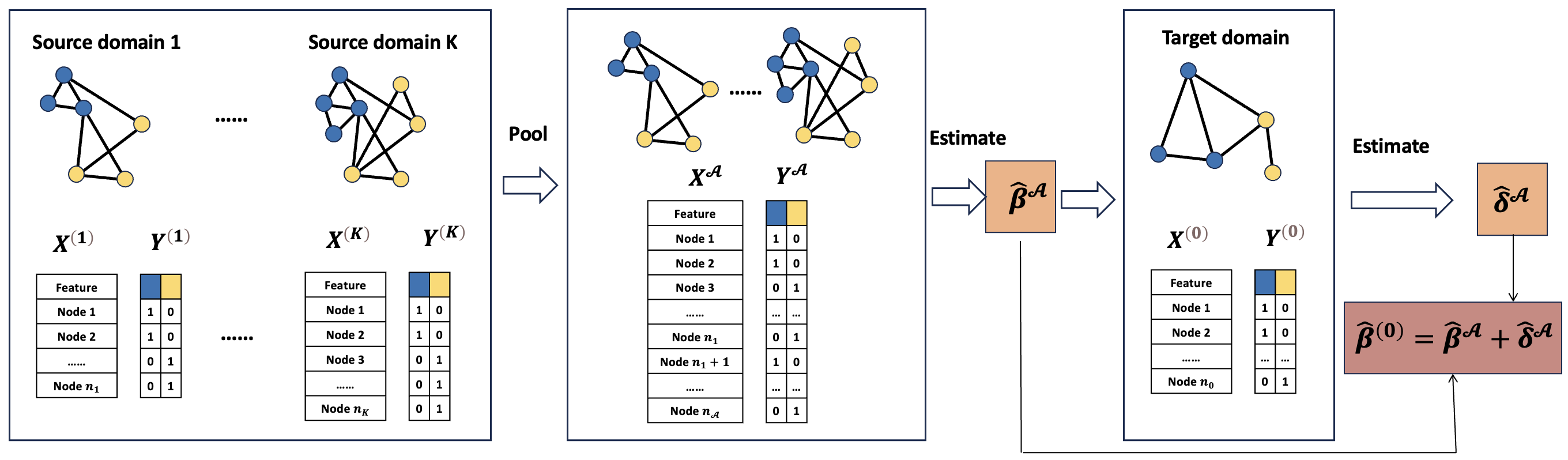}
    \caption{Workflow of Trans-$\our$. We first pool all source domains to get an estimate $\hat{\bbeta}^\mA$. We then use target data and the knowledge from $\hat{\bbeta}^\mA$ to estimate domain shift $\hat{\delta}^{\mA}$. The final estimate for the target data is $\hat{\bbeta}^{(0)}=\hat{\bbeta}^\mA+\hat{\delta}^{\mA}$. 
    }\label{fig:workflow}
\end{figure}

In this subsection, we present a transfer learning method under $\our$ (abbreviated as 
$\alg$) for the node classification task when the transferable source set $\mA$ is known, i.e., we have prior knowledge of which source data to utilize. Our method Trans-$\our$ is motivated by the transfer learning literature in the conventional regression model (without considering graph structure) \citep{bastani2018predicting, li2022transfer, tian2023transfer}.
Figure \ref{fig:workflow} shows the workflow of Trans-$\our$, which works in the following steps.   We first preprocess $\bA^{(k)}$ to obtain  the normalized adjacency matrices $\bS^{(k)}$, $k \in \{0,\mA\}$. 


\textbf{Source sample pooling.} We then pooled all source domain data in $\mA$  into a single source sample with pooled normalized adjacency matrix $\bS^{\mA} \in \mathbb{R}^{n_{\mA} \times n_{\mA}}$, node features $\bX^{\mA} \in \mathbb{R}^{n_{\mA} \times \dims}$, and labels $\bY^{\mathcal{\mA}} \in \{0,1\}^{n_{\mA} \times C}$, where $n_{\mA} =\sum_{k \in \mA}n_k$. $\bS^{\mA}$ has a block structure with diagonal blocks corresponding to the individual $\bS^{(k)}$ matrices, $k \in \mA$. The pooled node features $\bX^{\mA}$ and labels $\bY^{\mA}$ are obtained by concatenating the respective $\bX^{(k)}$ and $\bY^{(k)}$ matrices row-wise, $k \in \mA$. The rationale behind pooling the source samples in $\mA$ together is as follows. Since the source domains in $\mA$ are assumed to be similar to the target domain, with only small domain shifts, this naturally implies that the source domains in $\mA$ have similar underlying model parameters. By pooling the source samples, the algorithm gains access to a larger effective sample size from the same distribution, which leads to a more accurate and stable estimate.

\textbf{Source domain parameter estimation.}  Let $\bbeta^{\mA}$ denote the latent parameter 
in $\our$ that are used to generate $(\bY^\mA)$.
To obtain $\hat\bbeta^{\mA}$, we minimize the aforementioned 
 negative log-likelihood with $l_1-$norm penalty in Eq. \ref{eq:mle} using the pooled source samples. \textbf{Domain shift estimation.} Let $\delta^\mA$ denote the difference between the pooled source parameters $\bbeta^{\mA}$ and the target domain parameters $\bbeta^{(0)}$, i.e., $\bbeta^{(0)} = \bbeta^{\mA} + \delta^\mA$. 
We then estimate $\delta^\mA$ by minimizing the following loss function,  
 which essentially replaces $\bbeta^{(0)}$ with $\hat\bbeta^{\mA} + \delta^\mathcal{A}$ in the likelihood function of 
 $\our$ model using the target data, 
\begin{equation}
\label{eq:delta_esti}
\textstyle
     - \sum \limits_{i=1}^{n_0} \sum_{c=1}^{C} 
 [\bY^{(0)}_{ic} (\sum \limits_{j=1}^{n_0} s_{ij} \bX^{(0)}_j) (\hat{\bbeta}_c^\mA +
 \delta_c^\mathcal{A})  - \psi(\sum\limits_{j=1}^{n_0} s_{ij} \bX^{(0)}_j (\hat{\bbeta}_c^\mA +
 \delta_c^\mathcal{A}) )] 
  + 
  \lambda \sum\limits_{c=1}^{C-1} || \hat{\bbeta}_c^\mA +
 \delta_c^\mathcal{A}||_1,
\end{equation}
where $\psi(x) = \log{(1 + e^x)}, s_{ij}$ is the $(i,j)$th element of $(\bS^{(0)})^M$, $\hat{\bbeta}_c^\mA$ and $\delta_c^\mathcal{A}$ are the $c$th column of $\hat{\bbeta}^\mA$ and $\delta^\mathcal{A}$, respectively. This reformulation transforms the unknown parameter from $\bbeta^{(0)}$ to $\delta^\mathcal{A}$. 
Finally, the target domain parameter estimate is $\hat{\bbeta}^{(0)} =  \hat{\bbeta}^{\mathcal{A}} + \hat\delta^\mathcal{A}$, combining the estimated source domain parameters and the learned domain shift. 

We  summarize our procedure in 
Algorithm \ref{al:trans_alg} in Appendix \ref{sec:algorihtm_add}, which involves two hyperparameters: the graph convolution layers $M$ and the $l_1-$norm penalty strength $\lambda$.  When implementing Trans-$\our$, we apply cross-validation procedures to select these hyperparameters.

\subsection{Transferable Source Detection}
In section \ref{sec:method1}, we presented a method when the transferable set  $\mA$ is known. Nevertheless, in real applications, such prior knowledge might be unavailable. 
When the source domain differs significantly from the target domain, negative transfer can occur, leading to decreased performance on the target task \citep{li2022transfer, tian2023transfer}. To address this issue, we propose a data-driven cross-validation approach to select the transferable set $\mA$ automatically. 

 Our method begins by partitioning nodes in the target data into $V$ folds, i.e., $s_1, \ldots, s_V$. For each fold $v$, the labels of nodes in $s_v$  in the target data (denoted as $\bY_{s_v}^{(0)}$) are held out as the testing target data, while the labels in the remaining folds $\bY_{-s_v}^{(0)}$ serve as the training target data. 
   We  then apply the transfer learning Algorithm \ref{al:trans_alg} using the  $k$th source data $\{\bA^{(k)}, \bX^{(k)},  \bY^{(k)}\}$  and training target data
        $\{ \bA^{(0)}, \bX^{(0)}, \bY_{-s_v}^{(0)}\}$ to obtain the estimate $\hat{\bbeta}_{vk}^{(0)}$ for the target data. 
Note that we use the entire network structure of the target data $\bA^{(0)}$ and all node covariates $\bX^{(0)}$  to obtain aggregated features. However, we only use the classification labels of the training nodes $\bY_{-s_v}^{(0)}$ when minimizing the negative log-likelihood in Eq. \ref{eq:mle}  during the estimation process.
       
  Using the model estimate  $\hat{\bbeta}_{vk}^{(0)}$, we 
   predict labels for the testing nodes in the target data, 
  calculating $\hat{\mathbb{P}}(\bY_i^{(0)}=c)$, $i \in s_v$.
  The model's performance is assessed through the negative log-likelihood, $\mbox{NL}_v^{(k)}= -\sum_{i\in s_v}\sum_{c=1}^C (\bY^{(0)}_{ic} \log \hat{\mathbb{P}}(\bY^{(0)}_i=c) + (1-\bY^{(0)}_{ic})  \log (1- \hat{\mathbb{P}}(\bY^{(0)}_i=c))  $. 
 We then average  $\mbox{NL}_v^{(k)}$ across all $V$ folds, yielding $\mbox{NL}^{(k)} = V^{-1}\sum_{v=1}^{V}\mbox{NL}_v^{(k)}$. A lower score of $\mbox{NL}^{(k)}$ indicates higher transferability of the $k$th source data.  
 We rank the $K$ sources by their corresponding $\mbox{NL}^{(k)}$ values and select the top  $L$  sources with the lowest scores
 as the estimated transferable set $\hat{\mathcal{A}}$, where $L$ is a user-defined hyperparameter specifying the number of source data to include.  We summarize our procedure in Algorithm \ref{al:source_selection_alg} in the Appendix \ref{sec:algorihtm_add}. 


\input{Theory}

\section{Simulation Studies}


\textbf{Simulation setup.}
For the $k$th dataset, we generate simulation data in the following steps. 
We first generate  node features 
$\bX^{(k)} \in \mathbb{R}^{n_k \times \dims}$ from i.i.d. Gaussian with mean zero and identity covariance matrix.
 We then generate adjacency matrix $\bA^{(k)} \in \{0,1\}^{n_k \times n_k}$ using the ER random graph model with parameter $p_k$. 
We then generate a $C$-classes  response  $\bY^{(k)} \in \{0,1\}^{n_k \times C}$ using Eq. \ref{eq:GCMLR} with $M=1$.  

In this simulation, we consider a three-classes outcome, i.e., $C=3$.  For the target data, we set the target sample size $n_0=200$.
We consider high-dimensional sparse node covariates,  where the number of covariates is $d=500$, and the number of non-zero covariates in $\bbeta_c^{(0)}$ is $s=50$.  We set non-zero  model coefficients for the first $s$ covariates, i.e., 
$\bbeta_1^{(0)}=(0.4 \cdot \one_s, \zero_{\dims-\s} )$,  
$\bbeta_2^{(0)}=(0.5 \cdot \one_s, \zero_{\dims-\s} )$,  
where $\one_s$
has all $s$ elements 1 and $\zero_{\dims-\s}$ has all $\dims-\s$ elements 0.  For the $k$th source data, we set   
$\bbeta_1^{(k)}=\bbeta_1^{(0)} - (h_k \cdot \one_s, \zero_{\dims-\s} )$,  
$\bbeta_2^{(k)}=\bbeta_2^{(0)} + (h_k \cdot \one_s, \zero_{\dims-\s} )$, $k=1,\ldots, K$. For the first five sources, we set 
$h_k=h$ where $h$ is a small value and consider varying $h$ to study its effect.  For the subsequent sources, $h_k$  is set to be a large value, i.e., 10. We regard the first five sources as transferable datasets, that is, $\mA=\{1,\ldots,5\}$, and view the remaining sources as non-transferable datasets. We set the source sample sizes as the same value, i.e., $n_1=\ldots=n_K=n$.

\textbf{Evaluation metric and baselines.}  
We evaluate the performance by calculating the mean squared estimation errors (MSE) between the estimated coefficients and the true target coefficient, i.e., $\mbox{MSE}=\frac{1}{2d}(||\hat{\bbeta}_1^{(0)} - \bbeta_1^{(0)}||_F^2 + ||\hat{\bbeta}_2^{(0)} - \bbeta_2^{(0)}||_F^2)$. All experiments are replicated 100 times to calculate the averaged MSE. We compare our method Trans-$\our$ with two baselines: (1) $\our$ which uses target data only, and (2) Naive transfer learning (Naive TL), which 
 pools the source and target data together in a brute-force way and trains a single model on the combined dataset to obtain an estimation. 

Note that additional simulation results can be found in  Appendix \ref{sec:simu_add}. Specifically, in Figures \ref{fig:append-sbm} and \ref{fig:append-graphon}, we also show our method's superior performance when the target network is generated by other random graph models, including stochastic block model (SBM)  and graphon models. \textcolor{black}{In Figure \ref{fig:append-layer},} we also show our method's superior performance with multiple convolution layers, $M=2$. 

\subsection{Simulation Results When  the Transferable Source Set is Known}\label{sec:5.1}
We first show results when $\mA=\{1,\ldots,5\}$ is known, i.e., we will only use the first five source data to perform transfer learning. 

\begin{figure} 
    \centering
    \includegraphics[scale=0.395]{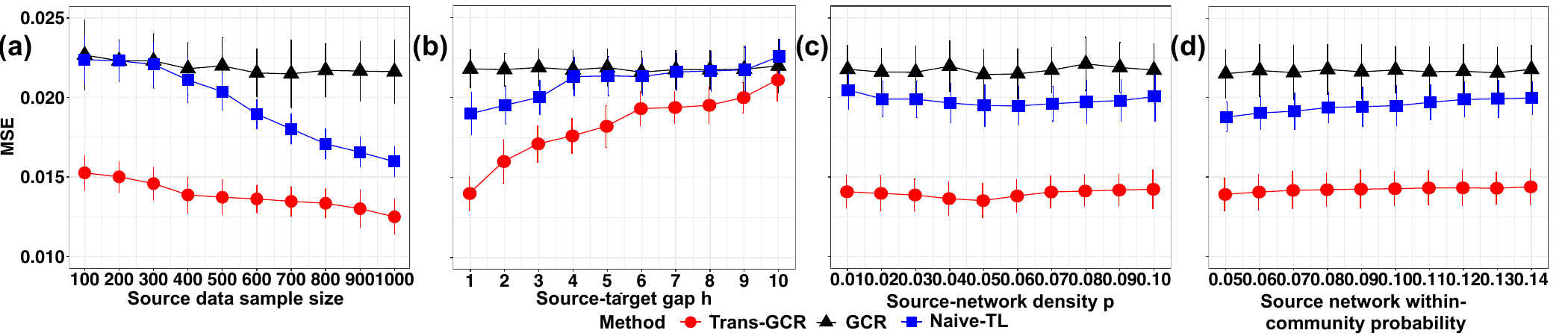}
    \caption{Performance comparison (MSE) of Trans-$\our$ (red), $\our$ (black), Naive TL (blue) across varying (a) Source sample size, (b) Source-target gap $h$, (c) Source network density (0.05 means identical densities) (d) Source network within-community probability (higher value means more discrepancy). }\label{fig:simu1}
\end{figure}

\textbf{Asymptotic performance.}  
 To investigate the asymptotic performance of the Trans-$\our$ method,  we vary the sample size of source data $n \in \{100,\ldots,1000\}$  while fixing source-target domain shift $h=1$, fixing the ER edge probability in the target data and source data $p_0=\ldots=p_5=0.05$.
 Figure \ref{fig:simu1}(a) shows that our method  Trans-$\our$ demonstrates a marked decrease in  MSE as the source data sample size increases. Naive TL also shows a decreasing MSE trend but at a significantly higher error rate compared with Trans-$\our$. 
 In contrast,    the $\our$ method has the highest MSE across all source sample sizes and remains unchanged since it does not utilize source data information. 


\textbf{Effect of source-target domain shift.} To investigate the impact of  $h$,  we vary $h \in \{1,\ldots, 10\}$, while fixing source sample size $n=600$, and fixing the ER edge probability $p_0=\ldots=p_5=0.05$. Figure \ref{fig:simu1}(b) reveals that the   MSE  of Trans-$\our$ and Naive TL  increases gradually as the source-target domain shift grows, which is expected since a larger shift implies reduced transferability between source and target domains. When $h$ increases to 10, the advantage of transfer learning disappears. 

\textbf{Effect of source-target network density discrepancy.}
 To investigate the impact of the network density difference between the source and target networks,  we vary the ER edge probability in the source data $p_1=\ldots=p_5 \in \{0.01, \ldots, 0.1\}$ while fixing the ER edge probability in the target data as 0.05, and fixing $n=600$, $h=1$. Figure \ref{fig:simu1}(c) reveals that both Trans-$\our$ and Naive TL show a slight U-shape trend, with the best results achieved when the source and target network densities are similar. As the density difference increases in either direction, the performance degrades.
 While network density discrepancies can impact the performance of transfer learning approaches, Trans-$\our$ demonstrates strong robustness in handling these differences, consistently outperforming $\our$ and Naive TL across the range of density variations tested.\begin{wrapfigure}{r}{0.35\textwidth}
\vspace{-1pt} 
\centering
\includegraphics[width=0.8\linewidth]{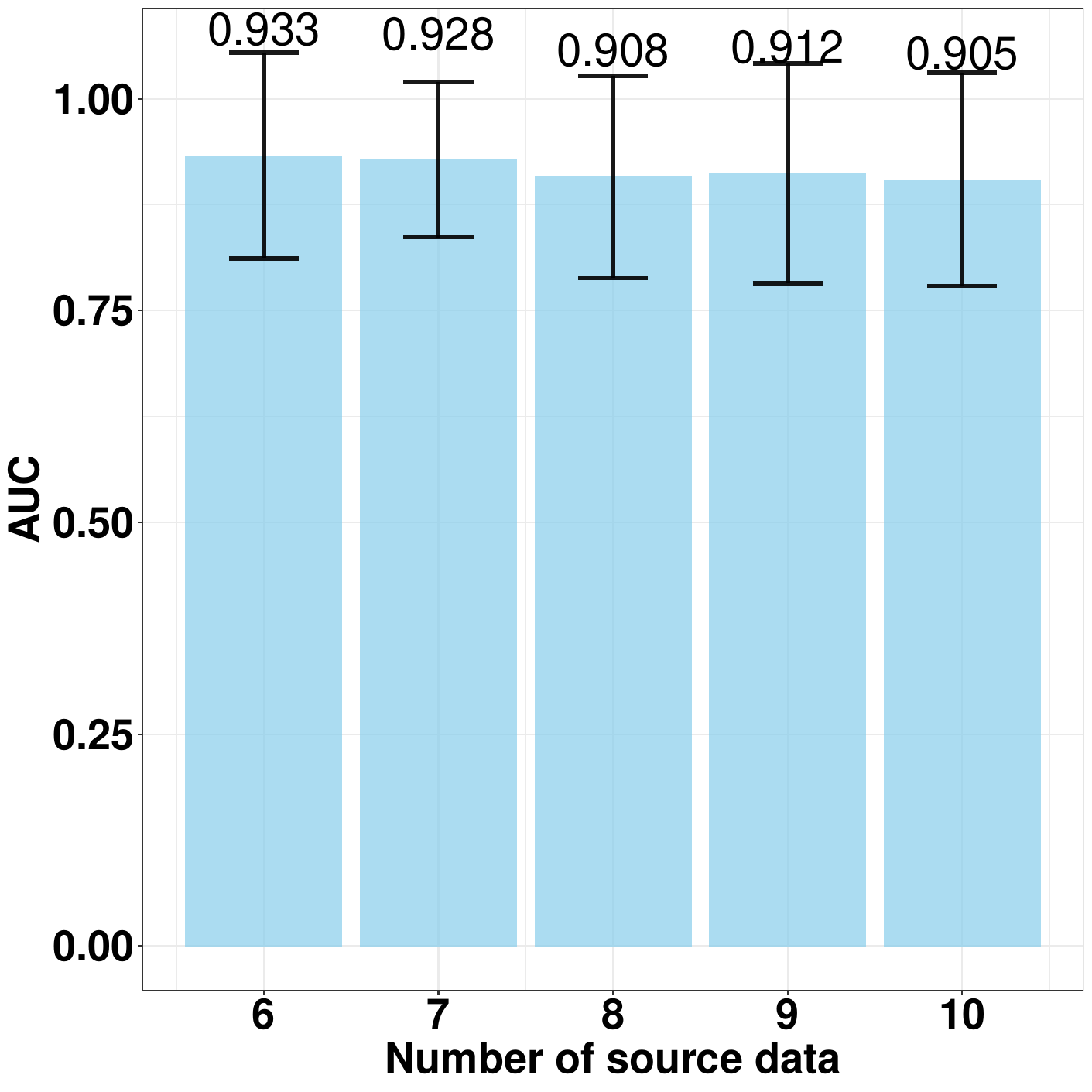}
\vspace{-3pt}
 \caption{\footnotesize{
AUC for detecting transferable sources.}}\label{fig:auc}
\vspace{-20pt} 
\end{wrapfigure} 
 

 \textbf{Effect of source-target 
 network distribution discrepancy.}
 To investigate the impact of distribution discrepancy between source and target networks,   we consider a scenario where the target network is generated from the ER model with parameter 0.05 while 
 the source networks are generated from a balanced two-block SBM.
 In the SBM,  we fix the between-community connection probability as 0.05 and the vary within-community probability from 0.05 to 0.14.  
When the within-community probability is 0.05, the SBM becomes identical to the ER(0.05) model used for the target network.
  We fix the source sample size $n=600$, and source-target domain shift $h=1$. 
  Figure \ref{fig:simu1}(d) shows that the  MSE of the Trans-$\our$  method slightly increases as the distribution discrepancy increases.  In contrast, the MSE of Naive TL shows a clear upward trend.


\subsection{Trasnferable Source  Detection Results}

Given $K$ source domains, we apply Algorithm \ref{al:source_selection_alg} to obtain the transferability score for each source domain. Here, we use a three-fold cross-validation. To assess the effectiveness of these scores in identifying transferable sources, we treat the task as a binary classification problem and compute the AUC (area under the ROC curve).  A higher AUC indicates better performance in distinguishing transferable and non-transferable sources based on the transferability scores.  Following the aforementioned setting, the first five source domains are defined as transferable.  We vary the number of total candidate source domains $K$ from 6 to 10, 
while fixing
ER edge probability  $p_0=\ldots=p_K=0.05$, and source data sample size  $n=600$. Figure \ref{fig:auc} shows that as $K$ increases, indicating a more challenging detection task, 
the AUC slightly decreases. Despite this, the AUC consistently remains above 0.9, showcasing the method's robustness and accuracy in identifying transferable source domains.

\section{Real Data Experiments}

\textbf{Data description.}
We conduct experiments on three widely-used real-world citation networks \citep{tang2008arnetminer, shen2020network}: (1)  DBLPv7 (abbreviated as D), containing 5484 nodes and  8130 edges,  extracted from the DBLP database; 
(2) Citationv1 (abbreviated as C), containing 8935 nodes and  15113 edges, obtained from the Microsoft Academic Graph database;
(3) ACMv9 (abbreviated as A), containing 9360 nodes and  15602 edges,  derived from the ACM digital library.
In these networks, each node represents a paper, and the adjacency matrix $\bA$ encodes the citation relationships between papers. 
The bag-of-words attribute vectors $\bX$ are derived from keywords extracted from paper titles, with a combined vocabulary of 6775 unique attributes across all networks. Thus, the feature dimension of $\bX$ is 6775. \begin{wrapfigure}{r}{0.4\textwidth}
\vspace{-1pt} 
\centering
\includegraphics[width=0.9\linewidth]{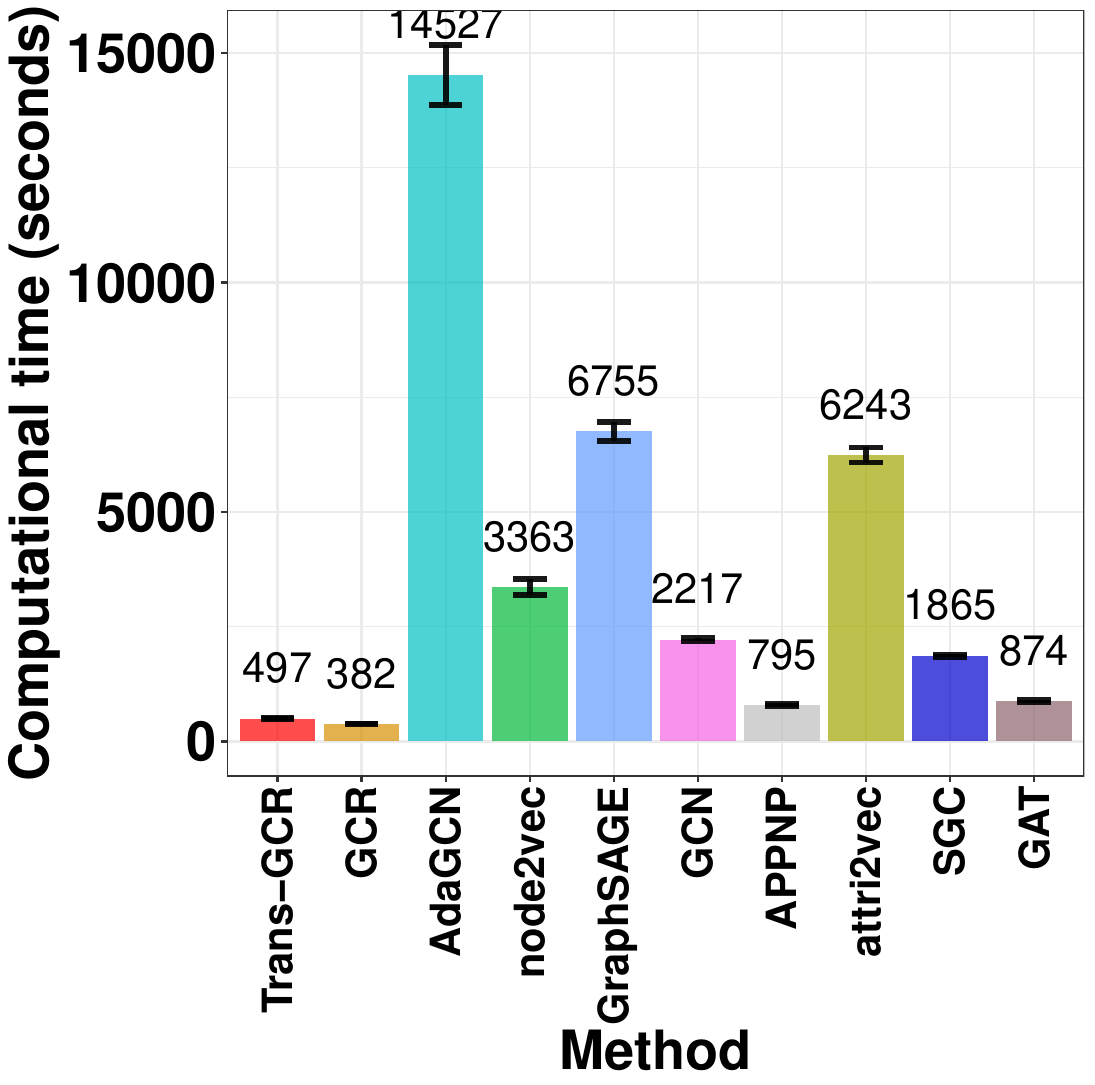}
\vspace{-1pt}
 \caption{\footnotesize{
 Averaged computational time (seconds) of different methods.
 }}\label{fig:real_time}
\vspace{-1pt} 
\end{wrapfigure}
Each paper is associated with a label belonging to one of these five classes: Databases, Artificial Intelligence, Computer Vision, Information Security, and Networking. 

\textbf{Transfer learning tasks.}
To evaluate the effectiveness of our method Trans-$\our$,  we conduct nine  transfer learning tasks between different source and target: (1) C $\rightarrow$ D, (2) A $\rightarrow$ D,  (3) C \& A $\rightarrow$ D, 
(4) D $\rightarrow$ C, (5) A $\rightarrow$ C, (6) D \& A $\rightarrow$ C,
(7)  D $\rightarrow$ A, (8) C $\rightarrow$ A, and (9) D \&  C $\rightarrow$ A.

\textbf{Baselines.} We compare our proposed method with several baselines: 
(1) AdaGCN \citep{dai2022graph}, which is a GCN-based  domain adaption method, and 
Naive TL methods based on (2) our proposed $\our$ model, and GCN-related methods, including (3)  Node2vec \citep{grover2016node2vec},  (4) GraphSAGE \citep{hamilton2017inductive}, 
(5) GCN \citep{kipf2017semi}, 
(6) APPNP \citep{gasteiger2018predict}, (7) attri2vec \citep{zhang2019attributed}, 
(8)
SGC \citep{wu2019simplifying},
and  (9) GAT \citep{velickovic2017graph}. 
Recall that Naive TL refers to pooling
the source and target data together in a brute-force way and training a single model on the combined dataset.
We used the default hyperparameter settings provided in the original implementations for the baseline methods.

\textbf{Evaluation.}
We evaluate the performance of our Trans-$\our$ method using two standard metrics: micro-F1 and macro-F1 scores \citep{dai2022graph}. 
These metrics assess the model's predictions on the testing subset of the target data. All experiments are replicated 10 times.


Our evaluation aims to answer three key questions:
(1) How does the performance of our Trans-$\our$ method compare with other baseline methods, given the fixed training rate? 
(2) How does the computational time of our Trans-$\our$ method compare with other baseline methods? (3) 
How does the training rate of the source and target networks affect the performance of Trans-$\our$? Here, the training rate of a network refers to the proportion of nodes whose labels were utilized for training the model.

\textbf{Results.} (1) \textbf{Performance comparison with fixed training rate.}  Table \ref{Tab:1} shows the averaged micro-F1 when a 75\% source training and a 3\% target training rate. Macro-F1 scores and standard deviation results are reported in Tables \ref{Tab:S1} and \ref{Tab:S2}.  The results demonstrate that Trans-$\our$ consistently outperforms the baseline methods across all tasks.   AdaGCN generally performs better than naive TL methods, indicating the advantages of domain adaptation. However, despite AdaGCN's good performance, it suffers from high computational costs, which we will discuss later.
Notably, Trans-$\our$ performs better with two source domains (e.g., C \& A $\rightarrow$ D)  than with one (e.g., C $\rightarrow$ D). This suggests that Trans-$\our$ effectively leverages complementary information from multiple source domains, enhancing transfer learning performance on the target task.
(2) \textbf{Computational time.}  Figure \ref{fig:real_time} shows that Trans-$\our$ has significantly lower computational time than AdaGCN, demonstrating its efficiency.

\begin{wrapfigure}{r}{0.45\textwidth}
\vspace{-1pt} 
\centering
\includegraphics[width=0.92\linewidth]{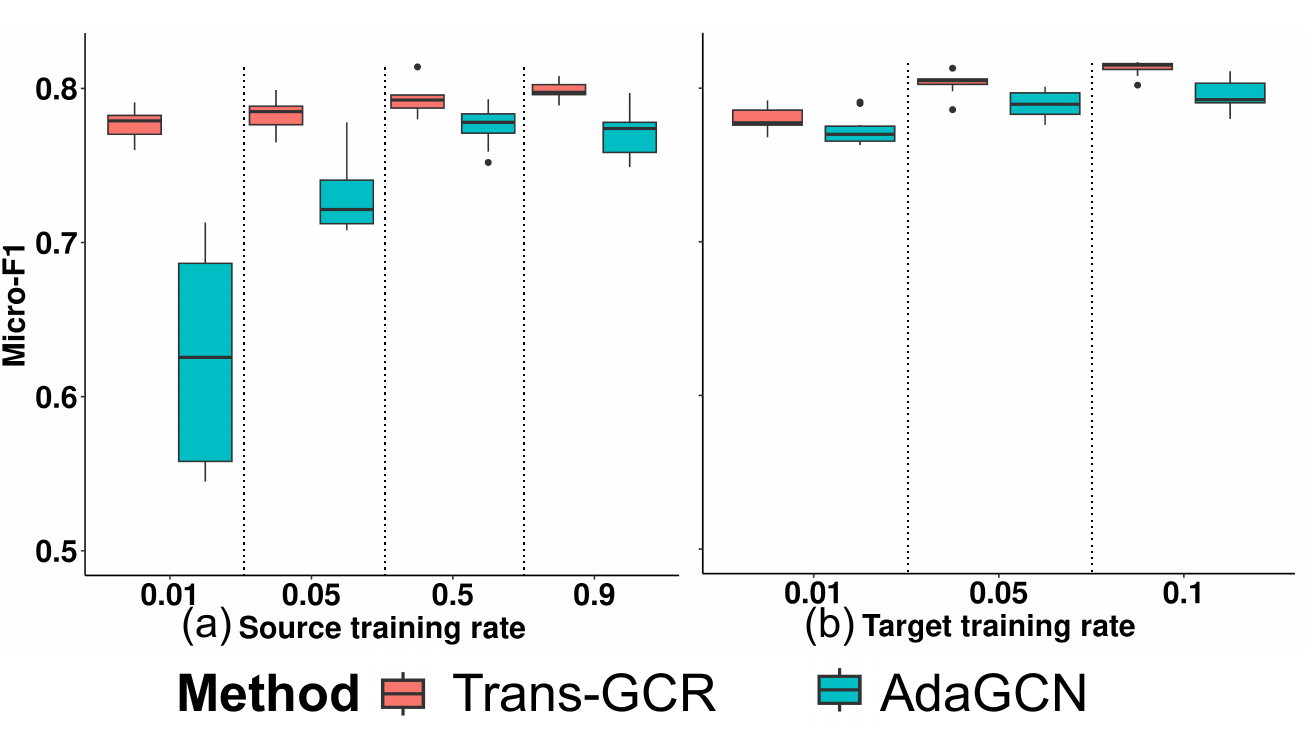}
\vspace{-1pt}
 \caption{\footnotesize{ (a) Varying source training rates (b)  Varying target training rates. 
} }\label{fig:rate}
\vspace{-1pt} 
\end{wrapfigure}
(3) \textbf{Effect of training rate.} Figure \ref{fig:rate}(a) shows the  micro-F1 for the transfer learning tasks D $\rightarrow$ C, as the source training rate increases from 1\% to 90\%, while 
the target rate is fixed at 3\%. 
Figure \ref{fig:rate}(b) depicts the micro-F1 results as the target training rate increases from 1\% to 10\%, while fixing the source training rate at 75\%.  As we can see,  our proposed method Trans-$\our$ consistently outperforms AdaGCN across all different source training rates and target training rates, while increasing rates leads to improved performance of our method. 
Results on other transfer learning tasks, leading to similar conclusions, are shown in Appendix \ref{sec:real_add}.  

\begin{table}[]
  \centering
  \caption{Averaged Micro F1 score (\%) of various methods, over 10 replicates,  with source training rate fixed at 0.75 and target training rate fixed at  0.03.}\label{Tab:1}
  \resizebox{1\linewidth}{!}{
\begin{tabular}{cccccccccccc} 
\toprule 
Target & Source & \textbf{Trans-\our}  & GCR  & AdaGCN & node2vec & GraphSAGE & GCN & APPNP & attri2vec & SGC & GAT  \\ 
\midrule
\multirow{3}{*}{D} & C      & \textbf{76.53}                    & 72.30  & 74.42                      & 66.55                        & 69.78                         & 71.59                   & 73.10                     & 67.64                         &    72.12                & 71.74                    \\
                   & A      & \textbf{75.16}                   & 69.75  & 74.52                     & 56.66                        & 65.82                         & 69.12                   & 68.23                     & 63.15                         &      68.64             & 67.34                    \\
                   & C\&A     & \textbf{76.61}                          & 70.09  & 73.93                      & 52.09                        & 56.22                         & 63.97                   & 70.31                     & 65.75                         & 71.31                 & 62.48                    \\ 
\midrule
\multirow{3}{*}{C} & D      & \textbf{78.99}  & 72.82   & 77.40                                 & 62.34                        & 69.63                         & 72.64                   & 75.26                     & 73.33                         & 75.06                 & 73.17                    \\
                   & A      & \textbf{80.37}    & 77.16   & 79.29                           & 63.17                        & 70.64                         & 73.85                   & 75.86                     & 69.91                         &  76.73            & 73.39                    \\
                   & D\&A     & \textbf{80.58} & 77.23 & 78.91                                          & 50.93                        & 60.77                         & 70.53                   & 74.81                     & 71.85                         &    77.31               & 68.83                    \\ 
\midrule
\multirow{3}{*}{A} & D      & \textbf{72.61} & 69.54  & 72.35                                       & 54.36                        & 62.59                         & 66.87                   & 66.56                     & 62.92                      &      65.55          & 66.67                    \\
                   & C      & \textbf{73.56} & 71.17    & 73.32                               & 61.53                        & 65.20                         & 66.10                   & 66.45                     & 64.23                         & 71.26                  & 67.79                    \\
                   & D\&C     & \textbf{73.78} & 71.32      & 73.26                                            & 49.27                        & 57.11                         & 63.19                   & 64.81                     & 63.42                         &    70.18                & 62.69                    \\ 
\bottomrule 
\end{tabular}}
\end{table}
\section{Discussion}
In this paper, we introduce the $\our$ model to capture the relationship between node classification labels ($\bY$), network structure ($\bA$), and node covariates (
$\bX$). We then propose Trans-$\our$, a transfer learning method that enhances estimation in the target domain using knowledge from the source domain. Despite its strong empirical performance and theoretical benefits, our method has limitations that need further research.  Firstly, our theoretical results are limited to the ER random graph model. While we show empirical success with other models like SBM and graphon, theoretical validation for these models is still needed. Secondly, the $\our$ model assumes a linear relationship between graph convolutional features and the log odds ratio; extending this to nonlinear models would be valuable. Thirdly, our source detection algorithm only provides a transferability score, and developing a hypothesis testing method for source domain transferability would be useful. Lastly, our algorithm currently transfers only point estimations of model parameters, and extending it to include confidence interval estimation would provide a measure of uncertainty.

\newpage

\bibliographystyle{unsrtnat}
\bibliography{00reference}

\newpage

\appendix


\newcommand{\beginsupplement}{%
        \setcounter{table}{0}
        \renewcommand{\thetable}{S\arabic{table}}%
        \setcounter{figure}{0}
        \renewcommand{\thefigure}{S\arabic{figure}}%
     }
     \beginsupplement

In this Appendix, we show (1) the notation table to briefly summarize important notations in this manuscript, (2) proofs of theoretical guarantees,
(3) additional experimental results, and (4) pseudo code of algorithms.

\section{Notation Table}\label{sec:notation_table}
We present the detailed expressions of notations widely used in the proposed model and algorithm in Table \ref{tab:notation}.

\begin{table*}[h]
\caption{Notations}\label{tab:notation}
\vskip 0.10in
\centering
\begin{tabular}{ll}
\toprule
Notations  & Description                        \\
\midrule
$n$                      & number of nodes in entire network               \\
$\bA\in \{0,1\}^{n\times n}$  & adjacency matrix of entire network without self-loops  \\
$\dims$ & number of covariates \\
$\bX \in \mathbb{R}^{n \times \dims}$                       &  entire covariate matrix                   \\
$C$ & number of classes \\
$\bY \in \{0,1\}^{n \times C}$                        & entire classification label matrix            \\
$\tilde{\bA}$                        & adjacency matrix of entire network with added self-connections                        \\
$\tilde{\bD}$                        & degree matrix of  $\tilde{\bA}$                        \\
$\bS$                     & normalized adjacency matrix of $\tilde{\bA}$                             \\
$n_0$ & target domain sample size \\
$\bA^{(0)} \in \{0,1\}^{n_0 \times n_0}$  & adjacency matrix of target network   \\
$\bX^{(0)} \in \mathbb{R}^{n_0 \times \dims}$                        &  covariate matrix of target domain     \\
$\bY^{(0)} \in \{0,1\}^{n_0 \times C}$                        & classification label matrix of target domain\\
$\bbeta^{(0)} \in \mathbb{R}^{\dims \times (C-1)}$         & true coefficient matrix associated with target domain under $\our$     \\
$n_k$ & the $k$th source domain sample size \\
 $\bA^{(k)} \in \{0,1\}^{n_k \times n_k}$       & adjacency matrix of the $k$th source network                       \\
$\bX^{(k)} \in \mathbb{R}^{n_k \times \dims}$                        &  covariate matrix of the $k$th source domain     \\
$\bY^{(k)} \in \{0,1\}^{n_k \times C}$               & classification label matrix of the $k$th source domain\\
$\bbeta^{(k)} \in \mathbb{R}^{\dims \times (C-1)}$     & true coefficient matrix associated with $k$th source domain under $\our$ \\
$\delta^{(k)}$                        &  difference between the  $k$th source and the target domain's coefficient       \\
$\mA_h$  & the set of $h-$level transferable source data, abbreviated as $\mA$ for brevity                                         \\
$p_0$        &  parameter of Erd\H{o}s--R\'enyi (ER) random graph model for target domain   \\
$p_k$        &  parameter of ER random graph model for the $k$th source domain   \\
$\bA^\mA$        &  pooled adjacency matrix  \\
$\bX^\mA$        &  pooled covariate matrix  \\
$\bY^\mA$        &  pooled label matrix  \\
$n^\mA$        &  sample size in the pooled sample  \\
$\bbeta^\mA$        & true underlying parameter related to the pooled sample $(\bA^\mA, \bX^\mA, \bY^\mA )$  \\
$\hat{\bbeta}^{(0)}$        &  estimate for $\bbeta^{(0)}$ obtained using our algorithms \\
\bottomrule
\end{tabular}
\end{table*}

\section{Proof of Theorem \ref{thm:main}}
 For notational simplicity, define $\psi(x) = \log{(1 + e^x)}$. Recall that our loss function is negative log-likelihood: 
    $$
    L_n({\bbeta}) = \frac1n \sum_i \left\{-{\bY}_i {\bZ}_i^\top {\bbeta} + \phi({\bZ}_i^\top {\bbeta})\right\} \,,
    $$
    and our estimator is: 
    $$
    \hat {\bbeta} = \argmin_{{\bbeta}} \left[L_n({\bbeta}) + \lambda \|{\bbeta}\|_1\right] \,.
    $$
    Furthermore, define the Bregman divergence $\delta L_n({\bbeta})$ as: 
    \begin{equation}
        \label{eq:bregman_div}
        \delta L_n({\bbeta}) = L_n({\bbeta}) - L_n({\bbeta}^{(0)}) - \langle {\bbeta} - {\bbeta}^{(0)}, \nabla L_n({\bbeta}^{(0)}) \rangle \,.
    \end{equation}
As $L_n(\cdot)$ is a convex loss function, it is immediate that $\delta L_n({\bbeta}) \ge 0$ for all ${\bbeta} \in \reals$. Furthermore, define $\hat v = \hat {\bbeta} - {\bbeta}^{(0)}$ and the penalized excess risk function $R_n$ as: 
$$
R_n(v) = L_n({\bbeta}^{(0)} + v) + \lambda \|{\bbeta}^{(0)} + v\|_1 - L_n({\bbeta}^{(0)}) - \lambda \|{\bbeta}^{(0)}\|_1 \,.
$$
As both the loss and penalty functions are convex, $R_n(\cdot)$ is a convex function. Furthermore, as $\hat {\bbeta}$ minimizes the penalized loss function, we have: 
\begin{equation}
\label{eq:mother_ineq}
R_n(\hat v) \le 0, \text{   or equivalently  
 } L_n(\hat {\bbeta}) + \lambda \|\hat {\bbeta}\|_1 \le L_n({\bbeta}^{(0)}) + \lambda \|{\bbeta}^{(0)}\|_1 \,.
\end{equation}
Recall that we need to show $\|\hat v\|_2^2 = \|\hat {\bbeta} - {\bbeta}^{(0)}\|_2^2 \le C(s\log{d})/n$. 
We prove this by reductio ad absurdum. 
As $s = o(\log{d}/n)$, we know $C(s\log{d})/n < 1$. Suppose that the claim of the theorem is not true. Then there exists $t \in (0, 1)$ such that $C(s\log{d})/n < \|t\hat v\|_2^2 \le 1$. Call $t \hat v = \tilde v$. By the convexity of $R_n(\cdot)$ we have: 
$$
R_n(\tilde v) = R_n(t\hat v) = R_n(t \hat v + (1 - t) 0) \le t\underbrace{R_n(\hat v)}_{\le 0} + (1-t) \underbrace{R_n(0)}_{=0} \le 0 \,. 
$$
The above chain of inequalities provides an immediate upper bound on $R_n(\tilde v)$. For the lower bound, we use Bregman divergence: 
\begin{align*}
    0 \ge R_n(\tilde v) & =  L_n({\bbeta}^{(0)} + \tilde v) + \lambda \|{\bbeta}^{(0)} + \tilde v\|_1 - L_n({\bbeta}^{(0)}) - \lambda \|{\bbeta}^{(0)}\|_1 \\
    & = \langle \tilde v, \nabla L_n({\bbeta}^{(0)}) \rangle + \delta L_n(\tilde v) + \lambda \|{\bbeta}^{(0)} + \tilde v\|_1 - \lambda \|{\bbeta}^{(0)}\|_1 \\
    & \ge  \delta L_n(\tilde v)  - \|\nabla L_n({\bbeta}^{(0)})\|_\infty \|\tilde v\|_1 + \lambda \|{\bbeta}^{(0)} + \tilde v\|_1 - \lambda \|{\bbeta}^{(0)}\|_1 
\end{align*}
Therefore, we have: 
\begin{align*}
\delta L_n(\tilde v) + \lambda \|{\bbeta}^{(0)} + \tilde v\|_1 \le \|\nabla L_n({\bbeta}^{(0)})\|_\infty \|\tilde v\|_1 + \lambda \|{\bbeta}^{(0)}\|_1 \,.
\end{align*}
Now suppose we choose $\lambda$ such that $\|\nabla L_n({\bbeta}^{(0)})\|_\infty \le \lambda/2$ with high probability. We will specify the choice of $\lambda$ later in the proof. Under this choice we have: 
\begin{align}
\label{eq:ineq_1}
& \delta L_n(\tilde v) + \lambda \|{\bbeta}^{(0)} + \tilde v\|_1 \le \frac{\lambda}{2}\|\tilde v\|_1 + \lambda \|{\bbeta}^{(0)}\|_1 \notag \\
\implies & \delta L_n(\tilde v) + \lambda \|\tilde v_{S^c}\|_1 \le \frac{3\lambda}{2}\|\tilde v_S\|_1 \,.
\end{align}
As $\delta L_n(\tilde v) \ge 0$, we further have $\|\tilde v_{S^c}\|_1 \le 3 \|\tilde v_S\|_1$, i.e. $\tilde v \in \cC(s, 3)$, 
where $\cC(s, \alpha)$ is defined as: 
$$
\cC(s, \alpha) = \left\{v \in \reals^p: \|v_{S^c}\|_1 \le \alpha \|v_S\|_1, \text{ for some } S \text{ with } |S| = s\right\} \,.
$$

We next present a lemma, which establishes a lower bound on $\delta L_n(\cdot)$ with high probability: 

\begin{lemma}
\label{lem:RSE_GLM}
Under Assumptions 1 and 2, there exists some positive constants $\kappa_{l}$ and $C_{4}$ such that,
$$
\delta L_n(u) \ge L_\psi(T) \|u\|_2^2 \left\{\kappa_l -  \left(C_4 \log{d} \sqrt{\frac{\Psi(p)}{n}}\right)\frac{\|u\|_1}{\|u\|_2}\right\}, \ \ \forall \ \|u\|_2 \le 1 \,.
$$
with probability at least $1- (\exp{\left(\frac12 \log{d} + 1-c_4 \log^2{d}\Psi(p)\right)} + 2\exp{\left(1-(c_1 -3/2)\log{d}\right)} + 2\exp{\left(\frac{3}{2}\log{d} + 1 - c_2 n\right)})$, where
$$
\Psi(p) = \frac{1-2p}{4p\log{((1-p)/p)}} \,,
$$
and $L_\psi(T)$ is a constant, same as in the proof of Proposition 2 of \cite{negahban2009unified}. 
\end{lemma}
We defer the proof the lemme to the end to maintain the flow. Consider the event when the upper bound of Lemma \ref{lem:RSE_GLM} holds. As by definition $\|\tilde v\|_2 \le 1$, applying Lemma \ref{lem:RSE_GLM} on $\tilde v$ yields: 
$$
\delta L_n(\tilde v) \ge L_\psi(T) \kappa_l \|\tilde v\|_2^2 - L_\psi(T)\left(C_4 \log{d} \sqrt{\frac{\Psi(p)}{n}}\right)\|\tilde v\|_1\|\tilde v\|_2 \,.
$$
Using this lower bound on the Bregman divergence in equation \eqref{eq:ineq_1}, we have: 
\begin{align*}
    L_\psi(T)\kappa_l \|\tilde v\|_2^2  + \lambda \|\tilde v_{S^c}\|_1 & \le \frac{3\lambda}{2}\|\tilde v_S\|_1 + L_\psi(T)\left(C_4 \log{d} \sqrt{\frac{\Psi(p)}{n}}\right)\|\tilde v\|_1\|\tilde v\|_2 \\
    & \le \frac{3\lambda\sqrt{s}}{2}\|\tilde v_S\|_2 + L_\psi(T)\left(C_4 \log{d} \sqrt{\frac{\Psi(p)}{n}}\right)(\|\tilde v_S\|_1 + \|\tilde v_{S^c}\|_1)\|\tilde v\|_2 \\
    & \le \frac{9}{8L_\psi(T) \kappa_l}s\lambda^2 + \frac{L_\psi(T) \kappa_l}{2}\|\tilde v_S\|_2 + 4L_\psi(T)\left(C_4 \log{d} \sqrt{\frac{\Psi(p)}{n}}\right)(\|\tilde v_S\|_1)\|\tilde v\|_2 \,.
\end{align*}
Changing sides we have: 
\begin{align*}
    \frac{L_\psi(T) \kappa_l}{2}\|\tilde v\|_2^2  + \lambda \|\tilde v_{S^c}\|_1 & \le \frac{9}{8L_\psi(T) \kappa_l}s\lambda^2 + 4L_\psi(T)\left(C_4 \log{d} \sqrt{\frac{\Psi(p)}{n}}\right)(\|\tilde v_S\|_1)\|\tilde v\|_2 \\
    & \le \frac{9}{8L_\psi(T) \kappa_l}s\lambda^2 + 4L_\psi(T)\sqrt{s}\left(C_4 \log{d} \sqrt{\frac{\Psi(p)}{n}}\right)\|\tilde v\|^2_2
\end{align*}
As under Assumption \ref{assm:sparsity}, we have $s\log{d} = o_p(\sqrt{n/\Psi(p)})$, eventually it will be smaller than $L_\psi(T)\kappa_l/4$. Therefore, we have: 
\begin{equation}
    \label{eq:ineq_2}
    \frac14 L_\psi(T)\kappa_l \|\tilde v\|_2^2+ \lambda \|\tilde v_{S^c}\|_1\le \frac{9}{8L_\psi(T) \kappa_l}s\lambda^2 \implies \|\tilde v\|_2^2 \le  \frac{9}{2L^2_\psi(T) \kappa^2_l}s\lambda^2\,.
\end{equation}
If we can show that a proper choice of $\lambda$ is $C_2 \sqrt{\log{d}/n}$, then we are done as in that we have: 
$$
\|\tilde v\|_2^2 \le \frac{9C_2}{2L^2_\psi(T) \kappa^2_l} \frac{s\log{d}}{n} \triangleq C \frac{s\log{d}}{n}
$$
which contradicts the assumption that $\|\tilde v\|_2 > C\sqrt{s\log{d}/n}$. This will complete the proof. In the following lemma, we show that, indeed $C_2 \sqrt{\log{d}/n}$ is a valid choice for $\lambda$: 
\begin{lemma}
\label{lem:lambda}
Under Assumption 4.1, there are universal positive constants $(c_1, c_2, c_3)$
such that
$$
\|\nabla L_n({\bbeta}^{(0)})\|_\infty = 
\left\|\frac{1}{n} \sum_{i = 1}^n {\bZ}_i \left\{{\bY}_i - \psi'({\bZ}_i^\top {\bbeta}^{(0)})\right\}\right\|_\infty \le C\sqrt{\frac{\log d}{n}} \,,
$$
with probability $1 - c_1 \left( d^{-c_2} + n^{-1} + e^{\log{n} - n p/c_3}\right) $.
\end{lemma}
Therefore, under the above lemma, we replace $\lambda$ by $C\sqrt{\log{d}/n}$ in equation \eqref{eq:ineq_2} we complete the proof. Proof of Lemma \ref{lem:RSE_GLM} and Lemma \ref{lem:lambda} can be found in Appendix \ref{app:additional_lem}

\section{Proof of Additional Lemmas}
\label{app:additional_lem}
\subsection{Proof of Lemma \ref{lem:RSE_GLM}}
The proof of Lemma \ref{lem:RSE_GLM} follows the basic structure of the proof of Proposition 2 of \cite{negahban2009unified}. 
However, suitable modifications are necessary to incorporate network dependency. 
We first state Proposition 2 of \cite{negahban2009unified} here for the convenience of the readers: 
\begin{proposition}[Proposition 2 of \cite{negahban2009unified}]
    Consider the logistic loss (negative log-likelihood) function: 
    $$
    L_n({\bbeta}) = \frac1n \sum_i \left\{-{\bY}_i {\bZ}_i^\top {\bbeta} + \log{(1 + e^{{\bZ}_i^\top {\bbeta}})}\right\} 
    $$
    Define the Bregman divergence of $L_n$ as follows: 
    $$
    \delta L_n(u) = L_n({\bbeta}^* + u) - L_n({\bbeta}^*) - \langle u, \nabla L_n({\bbeta}^*) \rangle = \frac{1}{n}\sum_i \psi''({\bZ}_i^\top {\bbeta}^* + v{\bZ}_i^\top u) ({\bZ}_i^\top u)^2 \,.
    $$
    where $\psi(x) = \log{(1 + e^x)}$. 
    Then we have: 
    $$
    \delta L_n(u) \ge \kappa_1 \|u\|_2 \left\{\|u\|_2 - \kappa_2 \sqrt{\frac{\log{d}}{n}}\|u\|_1\right\} \ \ \forall \ \ \|u\|_2 \le 1
    $$
    with probability at least $1 - c_1e^{-c_2 n}$ for some constant $\kappa_1, \kappa_2$. 
\end{proposition}
However, in their proof, the authors heavily used that ${\bZ}_i$'s are i.i.d., which is not longer true in our situation, as ${\bZ}_i$'s the rows of $\bA\bX/\sqrt{np}$. Below, we highlight the modification of the steps that are needed for the proof: 
\\\\
\noindent 
{\bf Modification 1: } We first need to show that there exists some constant $\kappa_l$ (not depending on $(n, p, d)$) such that 
$$
v^\top \bbE\left[\frac1n \sum_i {\bZ}_i^\top {\bZ}_i\right]v = v^\top \left(\frac{1}{n^2 p}\bbE[{\bX}^\top {\bA}^\top {\bA} {\bX}]\right) v \ge  \kappa_l \|v\|_2^2 
$$
for all $v \in \reals^d$. 
It is easy to see that  $\bbE[\bX^\top \bA^\top \bA \bX]/(n^2 p) = \Sigma_{X}$. Therefore, as long as we assume that $\lambda_{\min}(\Sigma_{X}) \ge \kappa_l$, we are done. The lower bound on the minimum eigenvalue of $\Sigma_{X}$ follows from Assumption \ref{assm1:dist_X}.
\begin{remark}
\label{rem:unknown_p}
     Suppose we do not know $p$. Then, we can estimate it from separate data (by data splitting) independent of current data. An application of Hoeffding's inequality yields: 
        $$
    \bbP\left(|\hat p - p| \ge t/n\right) \le Ce^{-ct^2} \,.
    $$
     Taking $t = np$ and $t= np/2$ (which goes to infinity as per Assumption \ref{assm:network_connection}), we have: 
    $$
    \bbP(p/2 \le \hat p \le 2p) \ge 1 - C_1e^{-cn^2p^2} \uparrow 1 \,.
    $$
    Therefore, we can perform the entire analysis conditioning on this event. 
\end{remark}
\begin{remark}
As per Remark \ref{rem:unknown_p}, if we condition on the event $\hat p \le 2p$, then we have on that event: 
$$
v^\top \bbE\left[\left(\frac{1}{n^2\hat p}\bbE[\bX^\top \bA^\top \bA \bX]\right)\right] v  \ge \frac{\kappa_l}{2}\|v\|_2^2 \,.
$$
Furthermore, if we add self-loop, it is easy to see that 
    $$
    \bbE[\bX^\top \bA^\top \bA \bX] = \left(1 - \frac{1-p}{np}\right)\Sigma_X 
    $$
    As the constant goes to $1$ as $n \uparrow \infty$, we have: 
    $$
      v^\top \bbE\left[\left(\frac{1}{n^2\hat p}\bbE[\bX^\top \bA^\top \bA \bX]\right)\right] v \ge \frac{\kappa_l}{4}\|v\|^2 \ \ \forall \ \ \text{large } n \,. 
    $$
\end{remark}
\noindent 
{\bf Modification 2: } To conclude equation (72) of \cite{negahban2009unified}, we need 
i) an upper bound on $\bbE[(u^\top {\bZ}_i)^4]$ and ii) a tail bound $\bbP(|u^\top {\bZ}_i| \ge t)$. Getting the tail bound is easy, as we can apply the Cauchy-Schwarz inequality: 
$$
\bbP(|u^\top {\bZ}_i| \ge t) \le \frac{\bbE[(u^\top {\bZ}_i)^2]}{t^2} \le \frac{\lambda_{\max}(\Sigma_X)}{t^2} \,.
$$
Now we need to bound the fourth moment (in fact, any $2 + \delta$ moment is sufficient): 
\allowdisplaybreaks
\begin{align*}
    & \bbE[(u^\top {\bZ}_i)^4] \\
    &= \bbE\left[\left(\frac{1}{\sqrt{np}}\sum_{j = 1}^n a_{ij}({\bX}_j^\top u)\right)^4\right] \\
    & = \bbE\left[\left(\frac{1}{\sqrt{np}}\sum_{j = 1}^n (a_{ij} - p + p)({\bX}_j^\top u)\right)^4\right] \\
    & \le 8\left\{\bbE\left[\left(\frac{1}{\sqrt{np}}\sum_{j = 1}^n (a_{ij} - p)({\bX}_j^\top u)\right)^4\right] + \bbE\left[\left(\frac{\sqrt{p}}{\sqrt{n}}\sum_{j = 1}^n({\bX}_j^\top u)\right)^4\right]\right\} \\
    & = 8\left\{\frac{1}{n^2}\sum_j \bbE\left[\left(\frac{a_{ij} - p}{\sqrt{p}}\right)^4 ({\bX}_j^\top u)^4\right] + \frac{1}{n^2} \sum_{j \neq j'}\bbE\left[\left(\frac{a_{ij} - p}{\sqrt{p}}\right)^2 ({\bX}_j^\top u)^2\right]\bbE\left[\left(\frac{a_{ij} - p}{\sqrt{p}}\right)^2 ({\bX}_j^\top u)^2\right] \right. \\
    & \hspace{10em} \left. + p^2 \frac{1}{n^2}\sum_j \bbE[({\bX}_j^\top u)^4] 
    + p^2 \frac{1}{n^2}\sum_{j \neq j'}\bbE[({\bX}_j^\top u)^2]\bbE[({\bX}_{j'}^\top u)^2]\right\} \\
    & \le \left\{\frac{1}{np}\bbE[({\bX}^\top u)^4] + \left(\bbE[({\bX}^\top u)^2\right)^2 + \frac{p^2}{n} \bbE[({\bX}^\top u)^4] + p^2\left(\bbE[({\bX}^\top u)^2\right)^2 \right\}
\end{align*}
which is finite as $\bbE[({\bX}^\top u)^4]$ is finite ($\bX$ is sub-gaussian) and $np \uparrow \infty$.
\begin{remark}
    If $p$ is unknown, then we can modify the first step by conditioning on the event $\hat p \ge p/2$ and have: 
    $$
    \bbE\left[\left(\frac{1}{\sqrt{n\hat p}}\sum_{j = 1}^n a_{ij}({\bX}_j^\top u)\right)^4\right]  \le 4 \bbE\left[\left(\frac{1}{\sqrt{np}}\sum_{j = 1}^n a_{ij}({\bX}_j^\top u)\right)^4\right] 
    $$
    The rest of the proof will remain the same. Furthermore, if we add self-loop, then we have to single the term $a_{ii} ({\bX}_i^\top u)^2$ out as now $a_{ii} = 1$. However, it will add another term of other $1/(n^2p^2)$ to the above bound, which is asymptotically negligible.  
\end{remark}
\noindent 
{\bf Modification 3: }We next show an analog of equation (76) of \cite{negahban2009unified}. 
Following the notations of \cite{negahban2009unified}, define a random process $Z(t)$ as: 
$$
Z(t) = \sup_{u \in \bbS_2(1) \cap \bbB_1(t)} \left|
\frac{1}{n}\sum_{i = 1}^n \left\{g_u({\bZ}_i) - \bbE[g_u({\bZ}_i)]\right\}\right| \triangleq F_t(Z_1, \dots, Z_n) \,.
$$
We will apply bounded difference inequality (Theorem 6.2 of \cite{boucheron2003concentration}). Note that conditional of $\bX$, ${\bZ}_i$'s are independent random vectors. Furthermore, for any $1 \le i \le n$ and for any $Z'_i \neq {\bZ}_i$: 
\begin{align*}
    & F_t(Z_1, \dots, Z_{i-1}, {\bZ}_i, \dots, Z_n) - F_t(Z_1, \dots, Z_{i-1}, Z'_i, \dots, Z_n) \\
    & \le \frac{1}{n}\sup_{u \in \bbS_2(1) \cap \bbS_1(t)} \left|g_u(Z'_i) - \bbE[g_u(Z'_i)]\right| \le \frac{\tau^2}{2n} \hspace{.1in} [\because g_u(\cdot) \le \tau^2/4]. 
\end{align*}
Therefore, by bounded difference inequality: 
$$
\bbP\left(Z(t) \ge \bbE[Z(t) \mid \bX] + t \mid {\bX}\right) \le \exp{\left(-\frac{8nt^2}{\tau^4}\right)}
$$
As the right-hand side does not depend on the value of $\bX$, we can further conclude the following by taking expectations with respect to $\bX$ on both sides: 
\begin{equation}
\label{eq:Zt_conc_bound_1}
\bbP\left(Z(t) \ge \bbE[Z(t) \mid \bX] + t \right) \le \exp{\left(-\frac{8nt^2}{\tau^4}\right)} \,.
\end{equation}
Next, using symmetrization and Rademacher complexity bounds, we bound $\bbE[Z(t) \mid \bX]$. For notational simplicity let us define: 
\begin{align*}
V_n & = \max_{1 \le j \le p} \left|\frac{1}{\sqrt{n}}\sum_{k = 1}^n {\bX}_{kj}\right| \\
\Gamma_n & = \max_{1 \le j \le d} \frac{1}{n}\sum_{k = 1}^n {\bX}_{kj}^2 \,.
\end{align*}
Now, as we have already pointed out, conditional on $\bX$, ${\bZ}_i$'s are i.i.d. random vectors. Therefore, the symmetrization argument holds, and following the same line of argument as of \cite{negahban2009unified}, we can conclude an analog of their equation (78): 
\begin{align*}
\bbE[Z(t) \mid \bX] & \le 8K_3t \bbE_{\eps, Z}\left[\max_{1 \le j \le d}\left|\frac{1}{n}\sum_{i=1}^n \eps_i {\bZ}_{ij}\mathds{1}_{|{\bZ}_i^\top {\bbeta}_*| \le T}\right| \mid \bX\right] \\
& = \frac{8K_3t}{\sqrt{n}} \bbE_{\bZ \mid \bX}\left[\bbE_{\eps\mid \bZ, \bX}\left[\max_{1 \le j \le d}\left|\frac{1}{\sqrt{n}}\sum_{i=1}^n \eps_i {\bZ}_{ij}\mathds{1}_{|{\bZ}_i^\top {\bbeta}_*| \le T}\right|\right]\right]
\end{align*}
First, observe that $\{\eps_1, \dots, \eps_n\}$ are Rademacher random variables (which are also subgaussian with sub-gaussian constant being 1), and therefore, conditionally on $\bZ$, 
$$
\frac{1}{\sqrt{n}}\sum_{i=1}^n \eps_i {\bZ}_{ij}\mathds{1}_{|{\bZ}_i^\top {\bbeta}_*| \le T} \text{ is subgaussian with norm } \sqrt{\frac{1}{n}\sum_{i=1}^n {\bZ}^2_{ij}\mathds{1}_{|{\bZ}_i^\top {\bbeta}_*| \le T}} \le \sqrt{\frac1n \sum_{i=1}^n {\bZ}^2_{ij}} \,.
$$
Therefore, from standard probability tail bound calculation, we have: 
$$
\bbE_{\eps\mid \bZ, \bX}\left[\max_{1 \le j \le d}\left|\frac{1}{\sqrt{n}}\sum_{i=1}^n \eps_i {\bZ}_{ij}\mathds{1}_{|{\bZ}_i^\top {\bbeta}_*| \le T}\right|\right] \le 
\sqrt{2\log{d}}\max_{1 \le j  \le d} \sqrt{\frac1n \sum_{i=1}^n {\bZ}^2_{ij}} \,.
$$
Therefore, we have: 
\begin{equation}
\label{eq:Zt_bound_1}
\bbE[Z(t) \mid \bX] \le 8\sqrt{2}K_3t \sqrt{\frac{\log{d}}{n}} \bbE\left[\max_{1 \le j  \le d} \sqrt{\frac1n \sum_{i=1}^n {\bZ}^2_{ij}} \mid \bX\right]
\end{equation}
Recall that by define ${\bZ}_{ij} = (\sum_k A_{ik}{\bX}_{kj})/\sqrt{np}$, which is not centered conditional on $\bX$. Therefore we first center it: 
$$
{Z}_{ij} = \frac{1}{\sqrt{np}}\sum_k A_{ik}{\bX}_{kj} = \frac{1}{\sqrt{np}}\sum_k (A_{ik} - p){\bX}_{kj} + \sqrt{\frac{p}{n}}\sum_k {\bX}_{kj} \triangleq \bar {\bZ}_{ij} + \sqrt{\frac{p}{n}}\sum_k {\bX}_{kj} \,.
$$
Using this we have: 
\begin{align}
\bbE\left[\max_{1 \le j  \le d} \sqrt{\frac1n \sum_{i=1}^n {\bZ}^2_{ij}} \mid \bX\right] & \le \bbE\left[\max_{1 \le j  \le d} \sqrt{\frac1n \sum_{i=1}^n \bar {\bZ}^2_{ij}} \mid \bX\right] + \sqrt{p} \max_{1 \le j  \le d} \left|\frac{1}{\sqrt{n}}\sum_k {\bX}_{kj}\right|  \notag \\
&= \bbE\left[\max_{1 \le j  \le d} \sqrt{\frac1n \sum_{i=1}^n \bar {\bZ}^2_{ij}} \mid \bX\right] + \sqrt{p}V_n \notag \\
\label{eq:Z_bound_1} & \le \sqrt{\bbE\left[\max_{1 \le j \le d} \frac1n \sum_{i=1}^n \bar {\bZ}^2_{ij} \mid \bX\right]}+ \sqrt{p}V_n \,.
\end{align}

\begin{remark}
\label{rem:p_hat_Z}
    If we have $\hat p$, then again, at the bound in equation \eqref{eq:Z_bound_1}, we have an additional factor of $\sqrt{2}$ for replacing $\hat p$ by $p$. 
\end{remark}
We next establish an upper bound on the conditional expectation of the maximum of the mean of $\bar {\bZ}_{ij}^2$. We first claim that $\bar {\bZ}_{ij}$ is a SG($\sigma_j)$ random variable with the value of $\sigma_j$ defined in equation \eqref{eq:Z_tilde_sg_norm} below. To see this, first note that, from Theorem 2.1 of \cite{ostrovsky2014exact}, we know $(A_{ik} - p)$ is SG($\sqrt{2}Q(p)$). Therefore, we have: 
\begin{equation}
\label{eq:Z_tilde_sg_norm}
\bar {Z}_{ij} = \frac{1}{\sqrt{np}}\sum_k (A_{ik} - p){\bX}_{kj} \in \mathrm{SG}\left(\sqrt{\frac{2Q^2(p)}{p} \frac1n \sum_k {\bX}_{kj}^2}\right) \triangleq \mathrm{SG}(\sigma_j) \,.
\end{equation}
\begin{remark}
    The same sub-gaussian bound continues to hold even under self-loop as $(1-p)$ is sub-gaussian with constant $\le 2Q^2(p)$. 
\end{remark}
Let $\mu_j = \bbE[\bar {\bZ}^2_{ij} | \bX]$. Then, by equation (37) of \cite{honorio2014tight}, we know $\bar {\bZ}^2_{ij} - \mu_j$ is a sub-exponential random variable, in particular: 
$$
\bar {\bZ}^2_{ij} - \mu_j \in \mathrm{SE}\left(\sqrt{32}\sigma_j, 4\sigma_j^2\right) \,.
$$
Hence we have, by equation (2.18) of \cite{wainwright2019high} (we use the version for the two-sided bound here): 
\begin{equation}
\label{eq:Z_bound_concentration_1}
\bbP\left(\left|\frac1n \sum_{i = 1}^n\left(\bar {\bZ}_{ij}^2 - \mu_j\right)\right| \ge t\right) \le \exp{\left(-\frac{1}{8\sigma_j^2}\min\left\{\frac{nt^2}{8}, nt\right\}\right)} \,.
\end{equation}
Going back to \eqref{eq:Z_bound_1}, we have: 
\begin{align*}
    \bbE\left[\max_{1 \le j \le d} \frac1n \sum_{i=1}^n \bar {\bZ}^2_{ij} \mid \bX\right] & = \bbE\left[\max_{1 \le j \le d} \left\{\left(\frac1n \sum_{i=1}^n (\bar {\bZ}^2_{ij} - \mu_j)\right) + \mu_j\right\} \mid \bX\right] \\
    & \le \bbE\left[\max_{1 \le j \le d} \left|\frac1n \sum_{i=1}^n (\bar {\bZ}^2_{ij} - \mu_j)\right| \mid \bX\right] + \max_{1 \le j \le d} \mu_j
\end{align*}
Now, bound the first term using the concentration inequality \eqref{eq:Z_bound_concentration_1}. Towards that end, define $\sigma_* = \max_j \sigma_j$ and observe that $\sigma_* = \sqrt{2Q^2(p)/p}\sqrt{\Gamma_n}$. 
\begin{align*}
    & \bbE\left[\max_{1 \le j \le d} \left|\frac1n \sum_{i=1}^n (\bar {\bZ}^2_{ij} - \mu_j)\right| \mid \bX\right] \le 8 \max\{\sigma_* \sqrt{\log{d}}, \sigma_*^2 \log{d}\} \,.
\end{align*}
Furthermore, observe that: 
\begin{align*}
    \mu_j = \bbE[\bar {\bZ}_{ij}^2 \mid \bX] & = \bbE\left[\left(\frac{1}{\sqrt{np}}\sum_k (A_{ik} - p)X_{kj}\right)^2 \mid \bX\right] = (1-p) \frac1n \sum_k X^2_{kj} \,,
\end{align*} 
which implies, $\max_{1 \le j \le d} \mu_j = (1-p) \Gamma_n$. Using these bounds in equation \eqref{eq:Z_bound_1}, we have: 
\begin{equation}
    \label{eq:Z_bound_2}
    \bbE\left[\max_{1 \le j  \le d} \sqrt{\frac1n \sum_{i=1}^n {\bZ}^2_{ij}} \mid \bX\right] \le \sqrt{\max\{\sigma_* \sqrt{\log{d}}, \sigma_*^2 \log{d}\} + (1-p) \Gamma_n} + \sqrt{p}V_n
\end{equation}
\begin{remark}
    Here also the constant will change by an additional factor of $\sqrt{2}$, see Remark \ref{rem:p_hat_Z}. 
\end{remark}
This, along, with equation \eqref{eq:Zt_bound_1},yields: 
\begin{align}
\label{eq:Zt_bound_2}
\bbE[Z(t) \mid \bX] & \le Ct\sqrt{\frac{\log{d}}{n}}\left(\sqrt{\max\{\sigma_* \sqrt{\log{d}}, \sigma_*^2 \log{d}\} + (1-p) \Gamma_n} + \sqrt{p}V_n\right) \notag \\
& \triangleq Ct\sqrt{\frac{\log{d}}{n}} g(\bX, p, d) \,.
\end{align}
Using this in the inequality \eqref{eq:Zt_conc_bound_1} yields: 
\begin{equation}
    \label{eq:Zt_conc_bound_2}
    \bbP\left(Z(t) \ge Ct\sqrt{\frac{\log{d}}{n}} g(\bX, p, d) + y \right) \le \exp{\left(-\frac{8ny^2}{\tau^4}\right)} \,.
\end{equation}
We next provide an upper bound for $g(\bX, p, d)$ term. Note that in the expression of $g(\bX, p, d)$, there are two key terms: $\Gamma_n, V_n$. Therefore, if we can obtain an upper bound on them individually, we can obtain an upper bound on $g(\bX, p, d)$. We start with $V_n$; for any fixed $j$, ${\bX}_{kj}$'s are i.i.d sub-gaussian random variable with constant $\sigma^2_X$. Therefore, we have: 
$$
\bbP\left(\left|\frac{1}{\sqrt{n}}\sum_{k = 1}^n {\bX}_{kj}\right| \ge t\right) \le 2e^{-\frac{t^2}{2\sigma^2_X}} 
$$
As a consequence, by union bound: 
$$
\bbP(V_n \ge t) = \bbP\left(\max_j \left|\frac{1}{\sqrt{n}}\sum_{k = 1}^n {\bX}_{kj}\right| \ge t\right) \le 2e^{\log{d} -\frac{t^2}{2\sigma^2_X}}
$$
Therefore, choosing $t = \sigma_X \sqrt{2c_1 \log{d}}$ (where $c_1 \ge 2$), we have: 
\begin{equation}
\label{eq:bound_Vn}
V_n \le  \sigma_X \sqrt{2c_1 \log{d}}) \ \ \text{with probability } \ge 1 - 2\exp{\left(-(c_1 -1)\log{d}\right)} \,.
\end{equation}
Call this event $\Omega_{n, {\bX}, 1}$. Our next target is $\Gamma_n$ which can be further upper bounded by: 
$$
\Gamma_n = \max_j \frac1n \sum_{k = 1}^n {\bX}_{kj}^2 \le \max_j \frac1n \sum_{k = 1}^n ({\bX}_{kj}^2 - \Sigma_{X, jj}) + \max_{j} \Sigma_{X, jj} \triangleq \bar \Gamma_n + \max_{j} \Sigma_{X, jj} \,.
$$
As we have assumed $\max_j \Sigma_{X, jj} \le C_1$ for some constant $C_1$, we need to bound $\bar \Gamma_n$. Here, we also use the fact that ${\bX}_{jk}^2 - \Sigma_{X, jj} \in \mathrm{SE}(\sqrt{32}\sigma_X, 4\sigma_X^2)$. Therefore, by equation (2.18) of \cite{wainwright2019high} we have: 
$$
\bbP\left(\left|\frac1n \sum_{k = 1}^n ({\bX}_{kj}^2 - \Sigma_{X, jj})\right| \ge t\right) \le 2\exp{\left(-\frac{n}{8\sigma^2_X}\min\left\{\frac{t^2}{8}, t\right\}\right)}
$$
Therefore, by union bound: 
$$
\bbP\left(\max_{1 \le j\le d}\left|\frac1n \sum_{k = 1}^n ({\bX}_{kj}^2 - \Sigma_{X, jj})\right| \ge t\right) \le 2\exp{\left(\log{d} -\frac{n}{8\sigma^2_X}\min\left\{\frac{t^2}{8}, t\right\}\right)}
$$
Choosing $t = \max_j \Sigma_{X, jj}$, we have: 
\begin{equation}
\label{eq:bound_Gamma_n}
\Gamma_n \le 2\max_j \Sigma_{X, jj} \le 2C_1 \ \ \text{with probability } \ge 1 -  2\exp{\left(\log{d} - c_2 n\right)} \,.
\end{equation}
Call this event $\Omega_{n, {\bX}, 2}$. 
Now, going back to the definition of $g(\bX, p, d)$ in equation \eqref{eq:Zt_bound_2}, we first note that, on the event $\Omega_{n,{\bX},1} \cap \Omega_{n, {\bX}, 2}$:  
$$
\sigma_* = \sqrt{\frac{2Q^2(p)}{p}\Gamma_n} \le 2\sqrt{\frac{C_1 Q^2(p)}{p}} \triangleq 2\sqrt{C_1 \Psi(p)}\,.
$$
It is immediate from the definition of $Q(p)$ that $\Psi(p) \sim 1/(-4p\log{p})$ for $p$ close to $0$. Therefore for all small $p$ and large $d$, $\sigma_*^2 \log{d} \ge 1$ and consequently $\max\{\sigma_* \sqrt{\log{d}}, \sigma_*^2 \log{d}\} = \sigma_*^2 \log{d}$. Hence, we have on the event $\Omega_{n,{\bX},1} \cap \Omega_{n, {\bX}, 2}$:
$$
g(\bX, p, d) \le C_2 \sqrt{\Psi(p) \log{d}} + 2C_1 + \sigma_X \sqrt{2c_1p \log{d}} \,.
$$
It is immediate that the dominating term is the first term, which implies: 
$$
g(\bX, p, d) \le 3C_2 \sqrt{\Psi(p) \log{d}} \,.
$$
We now use this bound in equation \eqref{eq:Zt_conc_bound_1}. Note that: 
\begin{align*}
    & \bbP\left(Z(t) \ge \bbE[Z(t) \mid \bX] + y \right) \\
    & \ge  \bbP\left(Z(t) \ge \bbE[Z(t) \mid \bX] + y, \Omega_{n,{\bX}, 1} \cap \Omega_{n, {\bX}, 2} \right)  \\
    & \ge \bbP\left(Z(t) \ge  3CC_2 t\log{d}\sqrt{\frac{\Psi(p)}{n}}  + y, \Omega_{n,{\bX}, 1} \cap \Omega_{n, {\bX}, 2} \right) \\
    & \ge \bbP\left(Z(t) \ge  3CC_2 t\log{d}\sqrt{\frac{\Psi(p)}{n}}  + y\right) + \bbP(\Omega_{n,{\bX}, 1} \cap \Omega_{n, {\bX}, 2} ) - 1
\end{align*}
Therefore, 
\begin{align}
    \label{eq:Zt_conc_bound_4}
     \bbP\left(Z(t) \ge  3CC_2 t\log{d}\sqrt{\frac{\Psi(p)}{n}}  + y\right)  & \le \exp{\left(-\frac{8ny^2}{\tau^4}\right)} + \bbP((\Omega_{n,{\bX}, 1} \cap \Omega_{n, {\bX}, 2})^c) \notag \\
     & \le \exp{\left(-\frac{8ny^2}{\tau^4}\right)} + 2\exp{\left(-(c_1 -1)\log{d}\right)} + 2\exp{\left(\log{d} - c_2 n\right)} \,.
\end{align}
Choosing $y = C_3 t \log{d}\sqrt{\Psi(p)/n}$, we have: 
\begin{align}
    \label{eq:Zt_conc_bound_3}
    & \bbP\left(Z(t) \ge  3CC_2 t\log{d}\sqrt{\frac{\Psi(p)}{n}}  + C_3t \log{d} \sqrt{\frac{\Psi(p)}{n}}\right) \notag \\
    & \le \exp{\left(-\frac{8C_3^2 t^2 \log^2{d}\Psi(p)}{\tau^4}\right)} + 2\exp{\left(-(c_1 -1)\log{d}\right)} + 2\exp{\left(\log{d} - c_2 n\right)} \,.
\end{align}

{\bf Modification 4: }Our last modification, not modification per se, but an application of peeling argument. Infact we want an upper bound on the event $\cE$ defined as: 
$$
\cE = \left\{Z(t) \ge 3eCC_2 t\log{d}\sqrt{\frac{\Psi(p)}{n}}  + C_3et \log{d} \sqrt{\frac{\Psi(p)}{n}}\ \text{ for some }t \in [1, \sqrt{d}] \right\} \,.
$$
Note that $t$ denotes the $\ell_1$ norm of a a vector $u$ such that $\|u\|_2 = 1$. Therefore, $t \in [1, \sqrt{d}]$. Also recall that $Z(t)$ is the suprema of the empirical process over all vectors $u$ such that $\|u\|_2 = 1$ and $\|u\|_1 \le t$. In peeling, we write $\cE$ as union of disjoint events. Define $\cE_j$ as: 
$$
\cE_j = \left\{Z(t) \ge 3eCC_2 t\log{d}\sqrt{\frac{\Psi(p)}{n}}  + C_3et \log{d} \sqrt{\frac{\Psi(p)}{n}}\ \text{ for some }t \in [\sqrt{d}/e^{j}, \sqrt{d}/e^{j-1}] \right\} \,.
$$
Therefore, 
$$
\cE \subseteq \cup_{j = 1}^{\lceil \frac12\log{d}\rceil} \cE_j \implies \bbP(\cE) \le \sum_{j = 1}^{\lceil \frac12\log{d}\rceil} \bbP(\cE_j) \,.
$$
Now observe that, for any $t \in [\sqrt{d}/e^{j}, \sqrt{d}/e^{j-1}]$, we have $Z(t) \le Z(\sqrt{d}/e^{j-1})$ and also 
\begin{align*}
3eCC_2 t\log{d}\sqrt{\frac{\Psi(p)}{n}}  + C_3et \log{d} \sqrt{\frac{\Psi(p)}{n}} & \ge 3eCC_2 \frac{\sqrt{d}}{e^{j}}\log{d}\sqrt{\frac{\Psi(p)}{n}}  + C_3e \frac{\sqrt{d}}{e^{j}} \log{d} \sqrt{\frac{\Psi(p)}{n}} \\
 & \ge 3CC_2 \frac{\sqrt{d}}{e^{j-1}}\log{d}\sqrt{\frac{\Psi(p)}{n}}  + C_3\frac{\sqrt{d}}{e^{j-1}} \log{d} \sqrt{\frac{\Psi(p)}{n}} \,.
\end{align*}
Therefore: 
\begin{align*}
    \bbP(\cE_j) & \le \bbP\left(Z\left(\frac{\sqrt{d}}{e^{j-1}}\right) \ge 3CC_2 \frac{\sqrt{d}}{e^{j-1}}\log{d}\sqrt{\frac{\Psi(p)}{n}}  + C_3\frac{\sqrt{d}}{e^{j-1}} \log{d} \sqrt{\frac{\Psi(p)}{n}}\right) \\
    & \le \exp{\left(-\frac{8C_3^2 d \log^2{d}\Psi(p)}{e^{2j - 2}\tau^4}\right)} + 2\exp{\left(-(c_1 -1)\log{d}\right)} + 2\exp{\left(\log{d} - c_2 n\right)} \\
    & \le  \exp{\left(-c_4 \log^2{d}\Psi(p)\right)} + 2\exp{\left(-(c_1 -1)\log{d}\right)} + 2\exp{\left(\log{d} - c_2 n\right)} 
\end{align*}
Hence: 
$$
\bbP(\cE) \le  \exp{\left(\frac12 \log{d} + 1-c_4 \log^2{d}\Psi(p)\right)} + 2\exp{\left(1-(c_1 -3/2)\log{d}\right)} + 2\exp{\left(\frac{3}{2}\log{d} + 1 - c_2 n\right)}  \,.
$$
On the event $\cE^c$ (which is a high probability event):  
$$
Z(t) \le 3eCC_2 t\log{d}\sqrt{\frac{\Psi(p)}{n}}  + C_3et \log{d} \sqrt{\frac{\Psi(p)}{n}}\ \text{ for all }t \in [1, \sqrt{d}] \,.
$$
Now let us conclude with the entire roadmap of the proof. First, following the same line of argument as of \cite{negahban2009unified} we show that 
\begin{align*}
\delta L_n(u) & \ge L_\psi(T) \frac1n \sum_i \phi_\tau\left((u^\top {\bZ}_i)^2\mathds{1}_{|{\bZ}_i^\top{\bbeta}_*| \le T}\right) \\
& = L_\psi(T) \|u\|^2 \frac1n \sum_i \phi_\tau\left(\left(u^\top {\bZ}_i/\|u\|\right)^2\mathds{1}_{|{\bZ}_i^\top{\bbeta}_*| \le T}\right) \\
& = L_\psi(T) \|u\|^2 \bbP_n(g_{u/\|u\|}(Z)) \\
& = L_\psi(T) \|u\|_2^2 \left\{P(g_{u/\|u\|}(Z)) + (\bbP_n - P)g_{u/\|u\|}(Z)\right\} 
\end{align*}
We have proved in Modification 1 that $P(g_{u/\|u\|}(Z)) \ge \kappa_l$. Therefore, 
$$
\delta L_n(u) \ge  L_\psi(T) \|u\|_2^2 \left\{\kappa_l + (\bbP_n - P)g_{u/\|u\|}(Z)\right\} 
$$
Now for any $u$, 
\begin{align*}
(\bbP_n - P)g_{u/\|u\|}(Z) & \le Z\left(\left\|\frac{u}{\|u\|_2}\right\|_1\right)  \\
& \le \left(3eCC_2 \log{d}\sqrt{\frac{\Psi(p)}{n}}  + C_3e \log{d} \sqrt{\frac{\Psi(p)}{n}}\right)\frac{\|u\|_1}{\|u\|_2} \,.
\end{align*}
Hence, we conclude that: 
$$
\delta L_n(u) \ge L_\psi(T) \|u\|_2^2 \left\{\kappa_l -  \left(C_4 \log{d} \sqrt{\frac{\Psi(p)}{n}}\right)\frac{\|u\|_1}{\|u\|_2}\right\} \,.
$$

\subsection{Proof of Lemma \ref{lem:lambda}}
Recall that we have: 
$$
\nabla L_n({\bbeta}^{(0)}) = \frac1n \sum_i {\bZ}_i\left\{{\bY}_i - \psi'({\bZ}_i^\top {\bbeta}^{(0)})\right\} \,.
$$
Now consider the $j^{th}$ element of $\nabla L_n({\bbeta}^{(0)})$, i.e., 
$$
\nabla L_n({\bbeta}^{(0)})_j - \frac1n \sum_i {\bZ}_{ij}\left\{{\bY}_i - \psi'({\bZ}_i^\top {\bbeta}^{(0)})\right\} \,.
$$
First, we show that conditional on $Z_{1j}, \dots, Z_{nj}$, the terms are mean $0$ (which is true from the definition of ${\bY}_i$), independent subgaussian random variable. The subgaussianity follows from the fact that: 
$$
\left\|\sum_i {\bZ}_{ij}\left\{{\bY}_i - \psi'({\bZ}_i^\top {\bbeta}^{(0)})\right\}\right\|_{\psi_2}^2 \le C \sum_i {\bZ}_{ij}^2 \| {\bY}_i - \psi'({\bZ}_i^\top {\bbeta}^{(0)})\|_{\psi_2}^2 \le C  \sum_i {\bZ}_{ij}^2 \,.
$$
Here $C$ is some absolute constant. Therefore, we have: 
\begin{equation}
\textstyle
    \label{eq:conditional_sg}
    \bbP\left(\max_{1 \le j \le d} \left|\frac{1}{\sqrt{n}} \sum_i {\bZ}_{ij}\left\{{\bY}_i - \psi'({\bZ}_i^\top {\bbeta}^{(0)})\right\}\right| \ge t \mid \bZ\right) \le c_1\exp{\left(\log{d} -c_2 \frac{t^2}{\max_{1 \le j \le d}\frac1n \sum_{i=1}^n{\bZ}_{ij}^2}\right)} 
\end{equation}
We next bound the term in the tail bound $\max_{1 \le j \le d} (\sum_{i=1}^n{\bZ}_{ij}^2)/n$. Towards that end, first observe that: 
\begin{align*}
\max_{1 \le j \le d} \frac1n \sum_{i=1}^{n} {\bZ}_{ij}^2 =  \max_{1 \le j \le d} \frac1n \bZ_{*j}^\top \bZ_{*j} & = \max_{1 \le j \le d}\frac{1}{n^2p} e_j^\top \bX^\top \bA^\top \bA \bX e_j \\
& = \max_{1 \le j \le d} \frac1n e_j^\top \bX\left(\frac{\bA^\top \bA}{np}\right) \bX^\top e_j \,.
\end{align*}
\begin{remark}
    If $p$ is unknown, i.e., we have $\hat p$, then, conditional on the event $p/2 \le \hat p \le 2p$, the above equality will be replaced by an inequality with an additional factor of $2$. 
\end{remark}
As we know $\bbE[\bX^\top \bA^\top \bA \bX]/(n^2p) = \Sigma_X$ and if we define $\sigma_+ = \max_j \Sigma_{X, jj}$, we have: 
$$
\max_{1 \le j \le d} \frac1n \sum_{i=1}^{n} {\bZ}_{ij}^2 \le \max_{1 \le j \le d} \left[\frac1n e_j^\top \bX\left(\frac{\bA^\top \bA}{np}\right) \bX^\top e_j - \Sigma_{X, jj} \right] + \sigma_+ \,.
$$

From Hanson-Wright inequality, we have for any matrix $\bQ$ (independent of ${\bX}$): 
$$
\bbP\left(\max_{1 \le j \le d } \left|\bX_{*j}^\top \bQ \bX_{*j} - \bbE[\bX_{*j}^\top \bQ \bX_{*j}]\right| \ge t\right) \le \exp\left(-c\min\Big(\frac{t^2}{\kappa_u^4\|\mathbf{Q}\|_{\mathrm{F}}^2},\frac{t}{\kappa_u^2 \|\mathbf{Q}\|_2}\Big)\right) \,.
$$
Here $\bQ = (\bA^\top \bA)/n^2p$. 
We use some concentration results on $\bQ$ in the rest of the proof. For notational convenience, set $\tilde \bA = \bA/\sqrt{np}$. We have the following concentration bound: 
\begin{lemma}
    \label{lem:conc}
For the Frobenous norm, we have with probability $\ge 1 - n^{-1}$: 
$$
\|\tilde \bA^\top \tilde \bA\|_F^2 \le \bbE[\|\tilde \bA^\top \tilde \bA\|_F^2] + n + n^2p^2 \le 2(n + n^2p^2) \,.
$$
For the operator norm, we have with probability $\ge 1 - e^{\log{n} -\frac{np}{c}}$: 
$$
\|\tilde \bA -\bbE[\tilde \bA]\|_{\rm op} \le 1 + 3\sqrt{2} \implies \|\tilde \bA\|_{\rm op} \le 2\sqrt{np} \,.
$$
\end{lemma}
\begin{remark}
    This lemma remains the same for the self-loop. 
\end{remark}
First, assume the above lemma is true, and consider the event so that the upper bound holds. Call that event $\cE$. On this event we have: 
$$
\bbP\left(\max_{1 \le j \le d } \frac1n\left|\bX_{*j}^\top \bQ \bX_{*j} - \bbE[\bX_{*j}^\top \bQ \bX_{*j}]\right| \ge t \mid \cE\right) \le \exp\left(\log{d}   -c\min\Big(\frac{n^2t^2}{2\kappa_u^4(n + n^2p^2)},\frac{t}{2\kappa_u^2 p}\Big)\right) \,.
$$
Choosing 
$$
t = K\max\left\{\sqrt{\frac{\log{d}}{n} + p^2 \log{d}}, p\log{d}\right\}
$$
we conclude: 
$$
\max_{1 \le j \le d } \frac1n\left|\bX_{*j}^\top \bQ \bX_{*j} - \bbE[\bX_{*j}^\top \bQ \bX_{*j}]\right| \le K\max\left\{\sqrt{\frac{\log{d}}{n} + p^2 \log{d}}, p\log{d}\right\} \le K\max\left\{\sqrt{\frac{\log{d}}{n}}, p\log{d}\right\} \,.
$$
Therefore, we have 
$$
\max_{1 \le j \le d} \frac1n \sum_{i=1}^{n} {\bZ}_{ij}^2 \le \sigma_+ + K\max\left\{\sqrt{\frac{\log{d}}{n}}, p\log{d}\right\}  \le 2\sigma_+
$$
with probability $\ge 1 - n^{-1} - e^{\log{n} -\frac{np}{c}}$. Call this event $\cE_1$. Therefore, we have: 
\begin{align*}
     & \bbP\left(\max_{1 \le j \le d} \left|\frac{1}{\sqrt{n}} \sum_i {\bZ}_{ij}\left\{{\bY}_i - \psi'({\bZ}_i^\top {\bbeta}^{(0)})\right\}\right| \ge t \right) \\
     & \le \bbP\left(\max_{1 \le j \le d} \left|\frac{1}{\sqrt{n}} \sum_i {\bZ}_{ij}\left\{{\bY}_i - \psi'({\bZ}_i^\top {\bbeta}^{(0)})\right\}\right| \ge t \mid \bZ \in \cE_1\right) + \bbP(\cE_1^c) \\
     & \le c_1 \exp{\left(\log{d} - \frac{c_2 t^2}{2\sigma_+}\right)} + \frac{1}{n} + \exp{\left(\log{n} - \frac{np}{c}\right)} \,.
\end{align*}
Choosing $t = K\sqrt{\log{d}}$ we complete the proof. 

\subsection{Proof of Lemma \ref{lem:conc}}

\def \bbE{\mathbb{E}}
\def \var{\operatorname{Var}}
\def \cov{\operatorname{Cov}}
\def \bin{\operatorname{Ber}}
\def \bbP{\mathbb{P}}
\def \op{\operatorname{op}}
\def \eps{\epsilon}

{\bf Upper bound on $\|\tilde \bA\|_{\op}$: }
To establish a bound on $\|\tilde \bA\|_{\op}$, we first center it: 
$$
\|\tilde \bA\|_{\op} = \|\tilde \bA - \bbE[\tilde \bA]\|_{\op} + \|\bbE[\tilde \bA]\|_{\op} \,.
$$
A bound on $\|\bbE[\tilde \bA]\|_{\op}$ directly follows from the definition: 
\begin{equation}
	\label{eq:A_op_exp_bound}
	\|\bbE[\tilde \bA]\|_{\op} = \frac{1}{\sqrt{np}}\|\bbE[\bA]\|_{\op} = \frac{1}{\sqrt{np}}\|p(\mathbf{1}\mathbf{1}^\top - I)\|_{\op} \le \sqrt{np} \,.
\end{equation}
Now we bound $ \|\tilde \bA - \bbE[\tilde \bA]\|_{\op}$.  
As $\tilde \bA = \bA/\sqrt{np}$, it is enough to bound $\|\bA - \bbE[\bA]\|_{\op}$. Using Corollary 3.12 and Remark 3.13 of \cite{bandeira2016sharp} (with $\eps = 1/2$), which implies: 
$$
\bbP\left(\|\bA - \bbE[\bA]\|_{\op} \ge 3\sqrt{2}\tilde \sigma + t\right) \le e^{\log{n} -\frac{t^2}{c\sigma_*^2}}
$$
where
$$
\tilde \sigma = \max_i \sqrt{\sum_j \var(\tilde \bA_{ij})} = \sqrt{np(1-p)} \le \sqrt{np}, \ \ \sigma_* = \max_{i, j} |\bA_{ij}| \le 1 \,.
$$
Therefore, we obtain: 
$$
\bbP\left(\|\bA - \bbE[\bA]\|_{\op} \ge 3\sqrt{2}\sqrt{np} + t\right) \le e^{\log{n} -\frac{t^2}{c}}
$$
Taking $t = \sqrt{np}$, we get: 
$$
\bbP\left(\|\bA - \bbE[\bA]\|_{\op} \ge (1 + 3\sqrt{2})\sqrt{np} \right) \le e^{\log{n} -\frac{np}{c}}
$$
As $\tilde \bA = \bA/\sqrt{np}$, we have: 
\begin{equation}
	\label{eq:A_op_dev_bound}
	\bbP\left(\|\tilde \bA - \bbE[\tilde \bA]\|_{\op} \ge (1 + 3\sqrt{2})\right) \le e^{\log{n} -\frac{np}{c}}
\end{equation}
Combining the bound on equation \eqref{eq:A_op_exp_bound} and \eqref{eq:A_op_dev_bound} we have with probability $\ge 1 - e^{\log{n} -\frac{np}{c}}$: 
\begin{equation}
	\label{eq:A_op_bound_final}
	\|\tilde \bA\|_{\op} \le \sqrt{np} + (1 + 3\sqrt{2}) \le 2 \sqrt{np}\,.
\end{equation}
\noindent
{\bf Finding a bound on $\|\tilde \bA^\top \tilde \bA\|_F^2$: }As before, we first find the expected value of $\|\tilde \bA^\top \tilde \bA\|_F^2$. For any $1 \le i \neq j \le n$: 
\begin{align*}
	\bbE[(\tilde \bA^\top \tilde \bA)_{ij}^2] = \frac{1}{n^2p^2}\bbE[(\bA^\top \bA)_{ij}^2] & = \frac{1}{n^2p^2}\bbE\left[\left(\sum_{k = 1}^n \bA_{ki}\bA_{kj}\right)^2\right] \\
	& = \frac{1}{n^2p^2}\left(\sum_{k = 1}^n \bbE[(\bA_{ki}\bA_{kj})^2] + \sum_{k \neq l} \bbE[(\bA_{ki}\bA_{kj})(\bA_{li}\bA_{lj})]\right) \\
	& \le \frac{1}{n^2p^2}( np^2 + n^2p^4) = \frac{1}{n} + p^2 \,.
\end{align*}
Now for $1 \le i = j \le n$: 
\begin{align*}
	\bbE[(\tilde \bA^\top \tilde \bA)_{ii}^2] & = \frac{1}{n^2p^2}\bbE\left[\left(\sum_{k = 1}^n \bA^2_{ki}\right)^2\right] \\
	& = \frac{1}{n^2p^2}\left(\sum_k \bbE[\bA^4_{ki}] + \sum_{k \neq l}\bbE[\bA_{ki}^2 \bA_{li}^2] \right)\\
	& \le \frac{1}{n^2p^2}(np + n^2p^2 ) = \frac{1}{np} + 1 \,.
\end{align*}
Therefore, we have: 
\begin{align*}
	\bbE[\|\tilde \bA^\top \tilde \bA\|_F^2] & = \sum_i \bbE[(\tilde \bA^\top \tilde \bA)_{ii}^2] + \sum_{i \neq j}\bbE[(\tilde \bA^\top \tilde \bA)_{ij}^2] \\
	& \le n\left(\frac{1}{np} + 1\right) + n^2 \left(\frac{1}{n} + p^2 \right) \\
	& \le n + \frac{1}{p} + n^2p^2 \le n + n^2p^2 \,. 
\end{align*}
The last inequality follows from $p \ge n^{-1}$. Next, we establish a bound on the variance: 
\begin{align*}
	\label{eq:var_break_1}
	\var\left(\|\tilde \bA^\top \tilde \bA\|_F^2\right) & = \frac{1}{n^4p^4} \var(\|\bA^\top \bA\|_F^2) \notag \\
	&  = \frac{1}{n^4p^4}\var\left(\sum_{i, j} (\bA^\top \bA)^2_{i,j}\right) \notag \\
	&  = \frac{1}{n^4p^4}\left[\sum_{i, j}\var\left((\bA^\top \bA)^2_{i,j}\right) + \sum_{(i, j) \neq (k, l)} \cov((\bA^\top \bA)^2_{i,j}, (\bA^\top \bA)^2_{k,l})\right] \notag \\
	& \triangleq \frac{1}{n^4p^4}(T_1 + T_2) \,.
\end{align*}
We bound $T_1$ and $T_2$ separately.
For that, we use some basic bounds on the raw moments of a binomial random variable; if $X \sim \bin(n, p)$, then $\bbE[X^k] \le Cn^kp^k$ for all $k \in \{1, 2, 3, 4\}$, for some universal constant $C$ as long as $np \uparrow \infty$. Observe that $({\bA}^\top {\bA})_{ii} \sim \bin(n-1, p)$ and $({\bA}^\top {\bA})_{ij} \sim \bin(n-2, p^2)$ for $i \neq j$. 
For $T_1$ we have: 
\begin{align*}
	\sum_{i, j}\var\left((\bA^\top \bA)^2_{i,j}\right) & = \sum_{i = 1}^n \var\left((\bA^\top \bA)^2_{ii}\right) + \sum_{i \neq j }\var\left((\bA^\top \bA)^2_{ij}\right) \\
	& \le \sum_i \bbE[(\bA^\top \bA)^4_{ii}] + \sum_{i \neq j }\bbE[(\bA^\top \bA)^4_{ij}] \\
	& \le C(n^5 p^4 + n^6p^8) \,.
\end{align*}
Next, we bound $T_2$, i.e., the covariance term. Note that if $(i, j, k, l)$ all are distinct, then covariance is $0$ as the terms are independent. Therefore, we only consider the cases when there are three or two distinct indices. We first deal with the terms of two distinct indices, i.e., $\cov(({\bA}^\top {\bA})^2_{ii}, ({\bA}^\top {\bA})^2_{ij})$ where $ i \neq j$. There are almost $n^2$ many terms of this form. For each of these types of terms: 
\begin{align*}
	\cov((\bA^\top \bA)^2_{ii}, (\bA^\top \bA)^2_{ij}) & = \bbE\left[(\bA^\top \bA)^2_{ii} (\bA^\top \bA)^2_{ij}\right] - \bbE\left[(\bA^\top \bA)^2_{ii}\right]\bbE\left[(\bA^\top \bA)^2_{ij}\right] \\
	& \le  \bbE\left[(\bA^\top \bA)^2_{ii} (\bA^\top \bA)^2_{ij}\right] \\
	& = \bbE\left[\left(\sum_{k, k' = 1}^n \bA^2_{ki}\bA_{k'i}\bA_{k'j}\right)^2\right] \\
	& = \bbE\left[\left(\sum_k \bA^3_{ki}\bA_{kj} + \sum_{k \neq k'} \bA^2_{ki} \bA_{k'i}\bA_{k'j}\right)^2\right] \\
	& = \bbE\left[\left(\sum_k \bA_{ki}\bA_{kj} + \sum_{k \neq k'} \bA_{ki} \bA_{k'i}\bA_{k'j}\right)^2\right] \\
	& \le 2\left(\bbE\left[\left(\sum_k \bA_{ki}\bA_{kj}\right)^2\right] + \bbE\left[\left(\sum_{k \neq k'} \bA_{ki} \bA_{k'i}\bA_{k'j}\right)^2\right]\right) \\
	& \le 2C(n^2 p^4 + n^4p^6) \,.
\end{align*}
Therefore, we have: 
\begin{equation}
	\label{eq:cov_bound_1}
	\sum_{i \neq j}  \cov(({\bA}^\top {\bA})^2_{ii}, ({\bA}^\top {\bA})^2_{ij}) \le 2C(n^4p^4 + n^6p^6) \,.
\end{equation}
Next, we bound the covariance terms of the form $\cov(({\bA}^\top {\bA})^2_{ij}, ({\bA}^\top {\bA})^2_{jk})$, i.e. two terms share an index with $i \neq j \neq k$. There are almost $n^3$ such terms. For each term: 
\begin{align*}
	\cov((\bA^\top \bA)^2_{ij}, (\bA^\top \bA)^2_{jk}) & \le \bbE\left[(\bA^\top \bA)^2_{ij} (\bA^\top \bA)^2_{jk}\right] \\
	& = \bbE\left[\left(\sum_{l, l'} \bA_{li}\bA_{lj}\bA_{l'i}\bA_{l'k}\right)^2\right] \\
	& = \bbE\left[\left(\sum_l \bA^2_{li}\bA_{lj}\bA_{lk} + \sum_{l \neq l'} \bA_{li}\bA_{lj}\bA_{l'i}\bA_{l'k}\right)^2\right] \\
	& = \bbE\left[\left(\sum_l \bA_{li}\bA_{lj}\bA_{lk} + \sum_{l \neq l'} \bA_{li}\bA_{lj}\bA_{l'i}\bA_{l'k}\right)^2\right] \\
	& = 2\left( \bbE\left[\left(\sum_l \bA_{li}\bA_{lj}\bA_{lk}\right)^2\right] + \bbE\left[\left(\sum_{l \neq l'} 
	\bA_{li}\bA_{lj}\bA_{l'i}\bA_{l'k}\right)^2\right]\right) \\
	& \le 2C(n^2p^6 + n^4p^8) \,.
\end{align*}
As there are almost $n^3$ such terms, we have: 
\begin{equation}
	\label{eq:cov_bound_2}
	\sum_{i \neq j \neq k} \cov((\bA^\top \bA)^2_{ij}, (\bA^\top \bA)^2_{jk}) \le 2C(n^5p^6 + n^7p^8)
\end{equation}
Therefore, combining \eqref{eq:cov_bound_1} and \eqref{eq:cov_bound_2}, we have: 
$$
T_2 \le C_1 \left(n^4p^4 + n^6p^6 + n^5p^6 + n^7p^8\right) \,.
$$
Combining the bounds on the variance and the covariance term, we conclude: 
$$
\var(\|\bA^\top \bA\|_F^2) \le C_2 (n^5 p^4 + n^6p^8 + n^4p^4 + n^6p^6 + n^5p^6 + n^7p^8) \le C_3(n^5 p^4 + n^6p^6 + n^7p^8) \,.
$$
Here the last equality follows from the fact that $n^5p^4 \ge n^4 p^4$, $n^6p^6 \ge n^6p^8$ and $n^6p^6 \ge n^4p^4$ (as $np \ge 1$). As a consequence, we have: 
$$
\var(\|\tilde \bA^\top \tilde \bA\|_F^2) = \frac{1}{n^4p^4}\var(\|\bA^\top \bA\|_F^2) \le C_3 (n + n^2p^2 + n^3p^4) \,.
$$
The last step involves an application of Chebychev's inequality: 
$$
\bbP\left(\|\tilde \bA^\top \tilde \bA\|_F^2 - \bbE[\|\tilde \bA^\top \tilde \bA\|_F^2] \ge t\right) \le \frac{\var(\|\tilde \bA^\top \tilde \bA\|_F^2)}{t^2} \,.
$$
Taking $t = n + n^2p^2$, we have: 
\begin{align*}
	\bbP\left(\|\tilde \bA^\top \tilde \bA\|_F^2 - \bbE[\|\tilde \bA^\top \tilde \bA\|_F^2] \ge n + n^2p^2\right) \le \frac{\var(\|\tilde \bA^\top \tilde \bA\|_F^2)}{(n + n^2p^2)^2} \le C_3 \frac{n(1 + np^2 + n^2p^4)}{n^2(1 + 2np^2 + n^2p^4)} \le \frac{C_3}{n} \,.
\end{align*}
Therefore, with probability $\ge 1 - n^{-1}$: 
$$
\|\tilde \bA^\top \tilde \bA\|_F^2 \le \bbE[\|\tilde \bA^\top \tilde \bA\|_F^2] + n + n^2p^2 \le 2(n + n^2p^2) \,.
$$

\section{Extension of Theorem \ref{thm:main} under multiple source}
\label{sec:thm_mult_source}
In Theorem \ref{thm:main}, we have established the convergence guarantee of $\hat \beta$ on a domain under network dependency. This section presents some ideas for extending our analysis when we have data from multiple related source domains. We have conjectured a theorem (Theorem \ref{thm:trans_gcr}) and lay down the steps needed to prove it. There is one conjecture (Conjecture \ref{D1_conjecture}), which, if true, will lead to a complete proof of the theorem. 

We start a simple setting with one source domain $\mA$ and one target domain. Consider the transfer learning setup, in which we have $n_1$ observations from the source domain and $n_0$ observations from the target domain. We assume that $p_0=p_1=p$. Define $\bZ = (\bA\bX)/\sqrt{np} \in \reals^{n \times d}$, and ${\bZ}_i$ is the $i$th row of $\bZ$. Given the logistic regression, the inverse link function \citep{mccullagh2019generalized} for logit link is $\psi'(u)=\logit(u)$, where $\psi(u)=\log(1+e^u)$.
 The method is similar to the proposed method, i.e., we have a two-step estimator: 
\begin{enumerate}
    \item  Step 1: First estimate $\hat \bbeta^{\mA}$ as: 
    $$
    \hat \bbeta^{\mA} = \argmin_{\bbeta} \ -\frac{1}{n_\mA+ n_0} \sum_{k \in \{0,\mA\} } \left\{ (\bY^{(k)})^\top  \bZ^{(k)}\bbeta - \log{(1 + e^{\bZ_i^\top \bbeta})}\right\} + \lambda_{{\bbeta}} \left\| \bbeta\right\|_1
    $$
    \item  Step 2: Then estimate the correction $\hat \delta^{\mA}$ only based on the target observations: 
    $$
    \hat \bdelta^{\mA} = \argmin_{\bdelta} \ -\frac{1}{n_0}  \left\{(\bY^{(0)})^\top \bZ^{(0)} (\hat \bbeta^{\mA} + \bdelta) - \log{(1 + e^{\bZ^\top (\hat \bbeta^{\mA} + \bdelta)})}\right\} + \lambda_{\delta} \left\| \bdelta\right\|_1 \,.
    $$
\end{enumerate}
Our final estimator for the target coefficient $\bbeta^{(0)}$ is $\hat{\bbeta}^{(0)} = \hat \bbeta^{\mA} + \hat \bdelta^{\mA}$. We must extend Theorem 1 of \cite{tian2023transfer} to handle the network dependency.
There are two key steps in the proof: i) to establish the rate of convergence of $\hat \bbeta^{\mA}$ (which is the estimator of $\bbeta^{\mA}$ obtained by combining all the observations from both the set of transferable source $\mA$ and the target domain, and ii) then establish the rate of convergence of the $\bbeta^{(0)}$ is $\hat{\bbeta}^{(0)} = \hat \bbeta^{\mA} + \hat \bdelta^{\mA}$, where $\hat \bdelta^{\mA}$ is obtained using only the target observations. 

We will use bolded $\boldsymbol{\psi}^{\prime}$ hereafter to denote the vector with each component from the scalar function $\psi^{\prime}$ with corresponding variables. Define $\hat{\bu}^{\mathcal{A}}=\hat \bbeta^{\mA} -  \bbeta^{\mA}$, $\mathcal{D}=\left\{\left(\boldsymbol{\bZ}^{(k)}, \bY^{(k)}\right)\right\}_{k \in \{0,\mA\} }$, and $L(\bbeta, \mathcal{D})$ is the negative log likelihood on the combined sample $\mathcal{D}$: 
$$
\begin{aligned}
L(\bbeta, \mathcal{D}) & =-\frac{1}{n_{\mathcal{A}}+n_{0}} \sum_{k \in \{0,\mA\} }\left(\bY^{(k)}\right)^{T} \bZ^{(k)} \bbeta + \frac{1}{n_{\mathcal{A}}+n_{0}} \sum_{k \in \{0,\mA\} } \sum_{i=1}^{n_{k}} \psi\left(\bbeta^{T} \bZ_{i}^{(k)}\right) \\
\nabla L(\bbeta, \mathcal{D}) & =-\frac{1}{n_{\mathcal{A}}+n_{0}} \sum_{k \in \{0,\mA\} }\left(\bZ^{(k)}\right)^{T} \bY^{(k)}+\frac{1}{n_{\mathcal{A}}+n_{0}} \sum_{k \in \{0,\mA\} }\left(\bZ^{(k)}\right)^{T} \boldsymbol{\psi}^{\prime}\left(\bbeta^{T} \bZ_{i}^{(k)}\right) \\
\delta L(\bu, \mathcal{D}) & ={L}\left(\bbeta^{\mathcal{A}}+\bu, \mathcal{D}\right)-L\left(\bbeta^{\mathcal{A}}\right)-\nabla L\left(\bbeta^{\mathcal{A}}\right)^{T} \bu.
\end{aligned}
$$

We present useful assumptions and lemmas first.


\begin{assumption} \label{assump2}
\textbf{(SubGaussian Assumption.)}
For any $\boldsymbol{a} \in \mathbb{R}^{p}, \boldsymbol{a}^{T} \bX^{(k)}$ are $\kappa_{u}\|\boldsymbol{a}\|_{2}^{2}$-subGaussian variables with zero mean for all $k \in \{0, \mA \}$, where $\kappa_{u}$ is a positive constant. 
\end{assumption}

\begin{assumption} \label{assump3}
\textbf{(Positive Definite Covariance Assumption.)}
Denote the covariance matrix of $\bX^{(k)}$ as $\boldsymbol{\Sigma}^{(k)}_{\bX}$, $k \in \{0,\mA \}$, we require that  $\lambda_{\min }\left(\boldsymbol{\Sigma}^{(k)}_{\bX} \right) \geq \kappa_{l}>0$, where $\kappa_{l}$ is a positive constant.
\end{assumption}

\begin{assumption}\label{assump4} 
\textbf{(Connectivity Bound of Network.)}
$p_k > \frac{\log n_k}{n_k}$, $k\in \{0, \mA \}$.
\end{assumption}

\begin{assumption}\label{assump5}
Denote 
$$
\widetilde{\boldsymbol{\Sigma}}_{h}=\sum_{k \in \{0,\mA \} } \alpha_{k} \mathbb{E}\left[
\boldsymbol{\bS}^{(k)}\left(\boldsymbol{\bS}^{(k)}\right)^{T}
\int_{0}^{1} \psi^{\prime \prime}\left(\left(\boldsymbol{\bS}^{(k)}\right)^{T} \bbeta^{(0)}
+t\left(\boldsymbol{\bS}^{(k)}\right)^{T}\left(\bbeta^{\mathcal{A}}- \bbeta^{(0)} \right)\right) dt \right]
$$
and $\widetilde{\boldsymbol{\Sigma}}_{h}^{(k)}=\mathbb{E}\left[\int_{0}^{1} \psi^{\prime \prime}\left(\left(\boldsymbol{\bS}^{(k)}\right)^{T} \boldsymbol{\beta^{(0)}}+t\left(\boldsymbol{\bZ}^{(k)}\right)^{T}\left({\bbeta}^{(k)}-\boldsymbol{\beta^{(0)}}\right)\right) d t \cdot \boldsymbol{\bS}^{(k)}\left(\boldsymbol{\bS}^{(k)}\right)^{T}\right]$. It holds that $\sup _{k \in \{0,\mA \} }\left\|\widetilde{\boldsymbol{\Sigma}}_{h}^{-1} \widetilde{\boldsymbol{\Sigma}}_{h}^{(k)}\right\|_{1}<\infty$.

\end{assumption}

\begin{lemma} \label{D1_lemma1}
Under Assumptions \ref{assump2} and \ref{assump5},

$$
\left\|{\bdelta}^{\mathcal{A}}\right\|_{1}=\left\|\bbeta^{\mathcal{A}}-\bbeta^{(0)}\right\|_{1} \leq C_{l} h
$$

where ${\bbeta}^{\mA}$ is the true coefficient of Step 1, $\bdelta^{\mathcal{A}}$ is the true coefficient of Step 2, and $\bbeta^{(0)}$ is the true coefficient of target domain. And $C_{l}:=\sup _{k \in\cT \cup \mathcal{A}}\left\|\widetilde{\boldsymbol{\Sigma}}_{h}^{-1} \widetilde{\boldsymbol{\Sigma}}_{h}^{(k)}\right\|_{1}<\infty$.

\end{lemma}
\begin{lemma} \label{D1_lemma2}
Under Assumptions \ref{assump2} and \ref{assump3}, there exists some positive constants $\kappa_{l}$ and $C_{4}$ such that,


$$
\delta L(\bu, \mathcal{D}) \ge L_\psi(T) \|\bu\|_2^2 \left\{\kappa_l -  \left(C_4 \log{d} \sqrt{\frac{\Psi(p)}{n}}\right)\frac{\|\bu\|_1}{\|\bu\|_2}\right\}
$$
with probability at least $1- (\exp{\left(\frac12 \log{d} + 1-c_4 \log^2{d}\Psi(p)\right)} + 2\exp{\left(1-(c_1 -3/2)\log{d}\right)} + 2\exp{\left(\frac{3}{2}\log{d} + 1 - c_2 n\right)})$, where $T$ is some constant, $L_\psi(T)= \min\limits_{u \leq |2T|}{ \psi^{{\prime}{\prime}}(u) }$, and $\Psi(p) \sim 1/(-4p\log{p})$.
\end{lemma}

\begin{lemma} \label{D1_lemma3}
Under Assumption \ref{assump2}, there are universal positive constants $(c_6, c_7, c_8)$
such that
$$\frac{1}{n_{\mA}+n_0}\left\|\sum_{k \in \{0,\mA\}} \left(\mathbf{Z}^{(k)}\right)^T\left[\bY^{(k)}-\boldsymbol{\psi}^{\prime}\left(\mathbf{Z}^{(k)}\boldsymbol{\beta}^{(k)}\right)\right]\right\|_\infty\lesssim\sqrt{\frac{\log d}{n_{\mA}+n_0}}$$
with probability $1 - (c_6 \left( d^{-c_7} + \sum_k n_k^{-1} + \sum_k e^{\log{n_k} - n_k p_k/c_8}\right)) $.
\end{lemma}

\begin{conjecture} \label{D1_conjecture}
Under Assumption \ref{assump2}, there are universal positive constants $(c_9, c_{10}, c_{11})$
such that
$$\frac{1}{n_{\mA}+n_0}\left\|\sum_{k \in \{0,\mA\}} \left(\mathbf{Z}^{(k)}\right)^T\left[\boldsymbol{\psi}^{\prime}\left(\mathbf{Z}^{(k)}\boldsymbol{\beta}^{(k)}\right) - \boldsymbol{\psi}^{\prime}\left(\mathbf{Z}^{(k)}\boldsymbol{\beta}^{\mA}\right)\right]\right\|_\infty\lesssim\sqrt{\frac{\log d}{n_{\mA}+n_0}}$$
with probability $1 - c_9 d^{-c_{10}} + \exp \left[  - c_{11} \left( n_{\mA} + n_0 \right)\right] $.
\end{conjecture}

\begin{lemma} \label{D1_lemma5}
 With high probability of at least $ 1 - d^{-\tilde K} - n^{-1} - e^{\log{n} -\frac{np}{c}}$, where $\tilde K$ and $c$ are some constants, there exists some constant $C$ such that:

$$
\frac{\|\bA\bX \bv\|_2^2}{n^2 p \|\bv\|_2^2} = \frac{\bv^\top \bX^\top \bA^\top \bA \bX \bv}{n^2 p\|\bv\|_2^2} \le C  \qquad \forall \ \bv: \|\bv_{S^c}\|_1 \le \kappa \|\bv_S\|_1 \,.
$$
\end{lemma}

\begin{theorem}\label{thm:trans_gcr} \textbf{(Convergence rate of Trans-$\our$).}
Under Assumptions  \ref{assump2}, \ref{assump3}, \ref{assump4}, \ref{assump5}, suppose $h \ll \sqrt{\frac{n_{0}}{\log d}}, h \leq c \sqrt{s}, n_{0} \geq c \log d$ and $n_{\mathcal{A} } \geq$ $c s \log d$, where $c>0$ is a constant, 
we have 
\[
\begin{array}{r}
\sup _{\xi \in \Xi(s, h)} \mathbb{P}\left(\|\hat{\boldsymbol{\beta}}^{(0)}-\boldsymbol{\beta}^{(0)} \|_{2} \lesssim h \log{d} \sqrt{\frac{\Psi(p)}{n_{\mA}+n_{0}}}+\sqrt{\frac{s \log d}{n_{\mA}+n_{0}}}+\left(\frac{\log d}{n_{\mA}+n_{0}}\right)^{1 / 4} \sqrt{h}  \right) \geq 1-n_{0}^{-1} 

\end{array}
\]
\end{theorem}


\subsection{Proof of Lemma \ref{D1_lemma1}} \label{lemma4}
By definition in (\ref{2.12}),

\begin{equation*}
\sum_{k \in \{0,\mA\}} \alpha_{k} \mathbb{E}\left\{ \left[\psi^{\prime}\left(\left({\bbeta}^{\mathcal{A}}\right)^{T} \bZ^{(k)}\right)-\psi^{\prime}\left(\left({\bbeta}^{(k)}\right)^{T} {\bZ}^{(k)}\right)\right] {\bZ}^{(k)}\right\}=\mathbf{0}_{p} 
\end{equation*} 

which implies

$$
\begin{aligned}
& \sum_{k \in \{0,\mA\}} \alpha_{k} \mathbb{E} 
\left\{\left[\psi^{\prime}\left(\left(\bbeta^{\mathcal{A}}\right)^{T} \bZ^{(k)}\right)-\psi^{\prime}\left(\left({\bbeta}^{(0)}\right)^{T} \bZ^{(k)}\right)\right] \bZ^{(k)}\right\}  \\
= & \sum_{k \in \{0,\mA\}} \alpha_{k} \mathbb{E}\left\{\left[\psi^{\prime}\left(\left(\bbeta^{(k)}\right)^{T} \bZ^{(k)}\right)-\psi^{\prime}\left(\left(\bbeta^{(0)}\right)^{T} \bZ^{(k)}\right)\right] \bZ^{(k)}\right\}
\end{aligned}
$$

By Taylor expansion,

$$
\begin{aligned}
& \sum_{k \in \{0,\mA\}} \alpha_{k} \mathbb{E}\left[\int_{0}^{1} \psi^{\prime \prime}\left(\left(\bbeta^{\mathcal{A}}\right)^{T} {\bZ}^{(k)}+t\left(\bbeta^{\mathcal{A}} -\bbeta^{(0)}\right)^{T} \bZ^{(k)}\right) \bZ^{(k)}\left(\bZ^{(k)}\right)^{T}\right]\left(\bbeta^{\mathcal{A}}-\bbeta^{(0)}\right) \\
= & \sum_{k \in \{0,\mA\}} \alpha_{k}  \mathbb{E}\left[\int_{0}^{1} \psi^{\prime \prime}\left(\left(\bbeta^{(k)}\right)^{T} {\bZ}^{(k)}+t\left(\bbeta^{(k)} -\bbeta^{(0)}\right)^{T} \bZ^{(k)}\right) \bZ^{(k)}\left(\bZ^{(k)}\right)^{T}\right] \left(\bbeta^{(k)}-\bbeta^{(0)}\right)
\end{aligned}
$$

Therefore, by Assumption \ref{assump5}, $\left\|\bbeta^{\mathcal{A}} -\bbeta^{(0)}\right\|_{1} \leq \sum_{k \in \mathcal{A}} \alpha_{k}\left\|\widetilde{\boldsymbol{\Sigma}}_{h}^{-1} \widetilde{\boldsymbol{\Sigma}}_{h}^{(k)}\right\|_{1} \cdot\left\|\bbeta^{(k)}-\bbeta^{(0)}\right\|_{1} \leq C_{l} h$.

\subsection{Proof of Lemma \ref{D1_lemma2}} 
\label{lemma:5}
See proof of Lemma \ref{lem:RSE_GLM}. 

\subsection{Proof of Lemma \ref{D1_lemma3}}
\label{sec:lemma6}

Here, we define $\mathbf{B}_i$ as the $i$-th row of matrix $\mathbf{B}$, and $\mathbf{B}_{(j)}$ as the $j$-th column of matrix $\mathbf{B}$. For a fixed index $j \in\{1,2, \ldots, p\}$, we denote $R_{i j}^{(k)}:=\bZ_{i j}^{(k)} \left(Y_{i}^{(k)}-\psi^{\prime}\left(\left\langle {\bbeta}^{(k)}, \mathbf{Z}_i^{(k)}\right\rangle\right)\right)$, and the $j$-th element of $ \left(\mathbf{Z}^{(k)}\right)^T \left[\bY^{(k)}-\boldsymbol{\psi}^{\prime}\left(\mathbf{Z}^{(k)}\bbeta^{(k)}\right)\right] $ can be written as $\sum_{i=1}^{n_k} R_{i j}^{(k)}$. Given the condition $\left\{\mathbf{Z}_i^{(k)}\right\}_{i=1}^{n_k}$ , $y_{i}^{(k)}$ follow a Bernoulli distribution with parameter $\frac{\exp \left(\left\langle \boldsymbol{\beta}^{(k)}, \mathbf{Z}_i^{(k)} \right\rangle \right) }{1+ \exp \left(\left\langle \boldsymbol{\beta}^{(k)}, \mathbf{Z}_i^{(k)} \right\rangle \right)}$. For any $t \in \mathbb{R}$, we compute 

$$
\begin{aligned}
\log \mathbb{E}\left[\exp \left(t R_{i j}^{(k)}\right) \mid \mathbf{Z}_i^{(k)}\right] & =\log \left\{\mathbb{E}\left[\exp \left(t \bZ_{i j}^{(k)} Y_{i}^{(k)}\right)\mid \mathbf{Z}_i^{(k)}\right] \exp \left(-t \bZ_{i j}^{(k)} \psi^{\prime}\left(\left\langle \boldsymbol{\beta}^{(k)}, \mathbf{Z}_i^{(k)} \right\rangle\right)\right)\right\} \\
& =\psi\left(t \bZ_{i j}^{(k)} + \left\langle \boldsymbol{\beta}^{(k)}, \mathbf{Z}_i^{(k)} \right\rangle\right)-\psi\left(\left\langle \boldsymbol{\beta}^{(k)}, \mathbf{Z}_i^{(k)} \right\rangle\right)- t {\bZ}_{ij}^{(k)} \psi^{\prime}\left(\left\langle \boldsymbol{\beta}^{(k)}, \mathbf{Z}_i^{(k)} \right\rangle\right)
\end{aligned}
$$

By second-order Taylor series expansion, we have

$$
\log \mathbb{E}\left[\exp \left(t R_{i j}^{(k)} \right) \mid \mathbf{Z}_i^{(k)}\right]=\frac{t^{2}}{2} \bZ_{i j}^{2} \psi^{\prime \prime}\left(\left\langle \boldsymbol{\beta}^{(k)}, \mathbf{Z}_i^{(k)} \right\rangle + v_{i} t \bZ_{i j}^{(k)} \right) \quad \text { for some } v_{i} \in[0,1]
$$

Since this upper bound holds for each $i=1,2, \ldots, n_k$, we have shown that

\begin{equation*}
\sum_{i=1}^{n_k} \log \mathbb{E}\left[\exp \left(t R_{i j}^{(k)} \right) \mid \mathbf{Z}_i^{(k)} \right] \leq \frac{t^{2}}{2}\left\{\sum_{i=1}^{n_k} \left( \bZ_{i j}^{(k)}\right) ^{2} \psi^{\prime \prime}\left(\left\langle \boldsymbol{\beta}^{(k)}, \mathbf{Z}_i^{(k)} \right\rangle + v_{i} t \bZ_{i j}^{(k)} \right)\right\}
\end{equation*}

For the link function $\psi(x) = \log\left\lbrace 1+ \exp\left(x \right) \right\rbrace $, it is easy to know that its second derivative $\psi^{\prime \prime}(x) = \exp\left(x \right)/\left(1+ \exp\left(x \right) \right)^2 $ takes values between 0 and 1, therefore the aforementioned equation can be simply bounded by an upper bound:

$$
\sum_{i=1}^{n_k} \log \mathbb{E}\left[\exp \left(t R_{i j}^{(k)} \right) \mid \mathbf{Z}_i^{(k)} \right] \leq \frac{t^{2}}{2}\sum_{i=1}^{n_k} \left( \bZ_{i j}^{(k)} \right)^{2}
$$

and

$$
\sum_{k\in\{0,\mA\}} \sum_{i=1}^{n_k} \log \mathbb{E}\left[\exp \left(t R_{i j}^{(k)} \right) \mid \mathbf{Z}_i^{(k)} \right] \leq \frac{t^{2}}{2} \sum_{k\in\{0,\mA\}} \sum_{i=1}^{n_k} \left( \bZ_{i j}^{(k)} \right)^{2}
$$

To control $\sum_{k\in\{0,\mA\}} \sum_{i=1}^{n_k} \left( \bZ_{i j}^{(k)} \right)^{2}$, it is easy to observe that it is a kind of quadratic forms respect to $\mathbf{X}^{(k)}_{(j)}$ so we will use Hanson-Wright inequality:

$$\begin{aligned}
	&\sum_{k\in\{0,\mA\}} \sum_{i=1}^{n_k} \left( \bZ_{i j}^{(k)}\right)^{2} \\
	= & \sum_{k\in\{0,\mA\}} {\mathbf{Z}_{(j)}^{(k)}}^{T}  \mathbf{Z}_{(j)}^{(k)}\\
	= & \sum_{k\in\{0,\mA\}} \left( \widetilde{\mathbf{A}}^{(k)} \mathbf{X}_{(j)}^{(k)}\right)^T \widetilde{\mathbf{A}}^{(k)} \mathbf{X}_{(j)}^{(k)}\\ \stackrel{\triangle}{=} & \sum_{k\in\{0,\mA\}} {\mathbf{X}_{(j)}^{(k)}}^T \mathbf{Q}^{(k)} \mathbf{X}_{(j)}^{(k)}\\
	= & {\mathbf{X}_{(j)}}^T \mathbf{Q} \mathbf{X}_{(j)}
\end{aligned}$$

where $\mathbf{X}_{(j)} \in \mathbb{R}^{(n_{\mathcal{A}}+n_0)}$ represents the vector obtained by vertically concatenating $\mathbf{X}_{(j)}^{(k)}$, and $\mathbf{Q} \in \mathbb{R}^{(n_{\mathcal{A}}+n_0) \times (n_{\mathcal{A}}+n_0)}$ represents the block diagonal matrix with diagonal elements $\mathbf{Q}^{(k)}$.
Hanson-Wright inequality tell us:

$$\begin{aligned}
	&\mathbb{P}\left\{\left| {\mathbf{X}_{(j)}}^T \mathbf{Q} \mathbf{X}_{(j)} -\mathrm{E}\left[{\mathbf{X}_{(j)}}^T \mathbf{Q} \mathbf{X}_{(j)} \right] \right| >t\right\}\\
	\leq & 2 \exp\Big[-c\min\Big(\frac{t^2}{\kappa_u^4\|\mathbf{Q}\|_{\mathrm{F}}^2},\frac{t}{\kappa_u^2 \|\mathbf{Q}\|_2}\Big)\Big]
\end{aligned}$$

In our article on linear regression, we have already proven that

$$
\begin{aligned}
	& \|\mathbf{Q} \|_F^2 = \sum_{k\in\{0,\mA\}} \|\mathbf{Q}^{(k)} \|_F^2 \leq \sum_{k\in\{0,\mA\}} 2(n_k + (n_k p_k)^2) \\
	& \|\mathbf{Q} \|_2 = \max_{k\in\{0,\mA\}} \|\mathbf{Q}^{(k)} \|_2 \leq  4 \max_{k\in\{0,\mA\}} \left\lbrace n_k p_k \right\rbrace 
\end{aligned}
$$

with high probability $1 - \sum_k n_k^{-1} - \sum_k e^{\log{n_k} -\frac{n_k p_k}{c}}$ converge to 1, and \\$\mathrm{E}\left[{\mathbf{X}_{(j)}}^T \mathbf{Q} \mathbf{X}_{(j)} \right] = \sigma_{jj}^2 \left( n_{\mathcal{A}}+n_0\right)  $

So we have the tail bound

$$
\begin{aligned}
&\mathbb{P}\left[\frac{1}{n_{\mathcal{A}}+n_0} \sum_{k\in\{0,\mA\}} \sum_{i=1}^{n_k} \left( \bZ_{i j}^{(k)} \right)^{2} \ge  C \right] \\
\leq & 2 \exp (- \frac{n_{\mathcal{A}}+n_0}{2} ) + \sum_k n_k^{-1} + \sum_k e^{\log{n_k} -\frac{n_k p_k}{c}}
\end{aligned}
$$

Define the event $\mathcal{E}=\left\{\max _{j=1, \ldots, p} \frac{1}{n_{\mathcal{A}}+n_0} \sum_{k\in\{0,\mA\}} \sum_{i=1}^{n_k} \left( \bZ_{i j}^{(k)} \right)^{2} \leq C \right\}$, we have

$$
\begin{aligned}
\mathbb{P}\left[\mathcal{E}^{c}\right] & \leq 2 \exp \left( -  \frac{n_{\mathcal{A}} + n_0}{2} +\log d\right) + \sum_k n_k^{-1} + \sum_k e^{\log{n_k} -\frac{n_k p_k}{c}} \\
\leq &2 \exp (-c \left( n_{\mathcal{A}}+n_0)\right)  + \sum_k n_k^{-1} + \sum_k e^{\log{n_k} -\frac{n_k p_k}{c}}
\end{aligned}
$$

where we have used the fact that $n_{\mathcal{A}} \gg \log d$. 

Given that $\left\lbrace \mathbf{Z}_i^{(k)}\right\rbrace  \in \mathcal{E}$, using the independence between $R_{i j}^{(k)}$ given $\mathbf{Z}_i^{(k)}$, we have

$$
\begin{aligned}
	&\frac{1}{n_{\mathcal{A}}+n_0} \sum_{k\in\{0,\mA\}} \sum_{i=1}^{n_k} \log \mathbb{E}\left[\exp \left(t R_{i j}^{(k)} \right) \mid \mathbf{Z}_i^{(k)} \right] \\
	= & \frac{1}{n_{\mathcal{A}}+n_0} \sum_{k\in\{0,\mA\}} \sum_{i=1}^{n_k} \mathbb{E}\left[ t R_{i j}^{(k)}  \mid \mathbf{Z}_i^{(k)} \right] \\
	\leq &c t^{2}  \quad \text { for each } j=1,2, \ldots, d
\end{aligned}
$$

By the Chernoff bound, we obtain 
$$\mathbb{P}\left[ \left|\frac{1}{n_{\mathcal{A}}+n_0} \sum_{k\in\{0,\mA\}} \sum_{i=1}^{n_k} R_{i j}^{(k)}\right| \geq \delta \mid \mathbf{Z}_i \right] \leq 2 \exp \left(-c(n_{\mathcal{A}}+n_0) \delta^{2}\right) $$

Combining this bound with the union bound yields
\begin{equation*}
\mathbb{P}\left[\max_{j=1, \ldots, d}\left|\frac{1}{n_{\mathcal{A}}+n_0} \sum_{k\in\{0,\mA\}} \sum_{i=1}^{n_k} R_{i j}^{(k)}\right| \geq t \mid \mathcal{E}\right] \leq 2 \exp \left(-c(n_{\mathcal{A}}+n_0) t^{2} + \log d\right)   
\end{equation*}

Setting $t = c \sqrt{\frac{\log d}{n_{\mathcal{A}}+n_0}}$, and putting together the pieces yields

$$
\begin{aligned}
&\mathbb{P}\left[\max_{j=1, \ldots, d}\left|\frac{1}{n_{\mathcal{A}}+n_0} \sum_{k\in\{0,\mA\}} \sum_{i=1}^{n_k} R_{i j}^{(k)}\right| \geq c \sqrt{\frac{\log d}{n_{\mathcal{A}}+n_0}} \right] \\
\leq & \mathbb{P}\left[\mathcal{E}^{c}\right]+\mathbb{P}\left[\max_{j=1, \ldots, d}\left|\frac{1}{n_{\mathcal{A}}+n_0} \sum_{k\in\{0,\mA\}} \sum_{i=1}^{n_k} R_{i j}^{(k)}\right| \geq t \mid \mathcal{E}\right] \\
\leq &  c_6 d^{-c_7} + \sum_k n_k^{-1} + \sum_k e^{\log{n_k} -\frac{n_k p_k}{c_8}}
\end{aligned}
$$

\subsection{Proof of Lemma \ref{D1_lemma5}}\label{sec:lemma5}
In this section, we prove that with high probability, there exists some constant $c_{12}$ such that:

$$
\frac{\|\bZ \bv\|_2^2}{n \|\bv\|_2^2}=\frac{\|\bA\bX \bv\|_2^2}{n^2 p \|\bv\|_2^2} = \frac{\bv^\top \bX^\top \bA^\top \bA \bX \bv}{n^2 p\|\bv\|_2^2} \le c_{12}  \ \forall \ \bv: \|\bv_{S^c}\|_1 \le \kappa \|\bv_S\|_1 \,.
$$

Using our previous notation, we define $\bZ = (\bA \bX)/\sqrt{np}$.  Define events $\Omega_{n, 1} = \left\{ \| \bA^\top  \bA\|_{\op} /np \le 4 np \right\}$ and $\Omega_{n, 2} = \left\{ \| \bA^\top \bA\|_F^2 / np \le  2(n + n^2p^2) \right\}$

Therefore, we have: 
\begin{align*}
    & \bbP\left(\left\|\frac{(\bX^\top \bA^\top \bA \bX)}{n^2 p} - \Sigma_X \left(1 - \frac{1}{n}\right)\right\|_\infty \ge t \right)  \\
    & \le \bbP\left(\left\|\frac{(\bX^\top \bA^\top \bA \bX)}{n^2 p} - \Sigma_X \left(1 - \frac{1}{n}\right)\right\|_\infty \ge t \mid \bA \in \Omega_{n, 1} \cap \Omega_{n, 2}\right) + \bbP(\bA \in (\Omega_{n, 1} \cap \Omega_{n, 2})^c) \\
    & \le 2\exp{\left(2\log{d} -c'\min\left(\frac{n^2 t^2}{n + n^2p^2}, \frac{t}{p}\right)\right)}  + e^{\log{n} -\frac{np}{c}} + \frac{1}{n} \,.
\end{align*}
The last step comes from Hanson Wright inequality.

Therefore, choosing 
$$
t = K\max\left\{\sqrt{\frac{\log{d}}{n} + p^2 \log{d}}, p\log{d}\right\}
$$
we conclude: 

\scalebox{0.85}[1]{
$$
\boxed{
\left\|\frac{(\bX^\top \bA^\top \bA \bX)}{n^2 p} - \Sigma_X \left(1 - \frac{1}{n}\right)\right\|_\infty \le K\max\left\{\sqrt{\frac{\log{d}}{n} + p^2 \log{d}}, p\log{d}\right\} \le K\max\left\{\sqrt{\frac{\log{d}}{n}}, p\log{d}\right\} \,.
}
$$
}

which means 

$$
\left\|\frac{\bX^\top \bA^\top \bA \bX}{n^2 p} - \Sigma_X\right\|_\infty \le K\max\left\{\sqrt{\frac{\log{d}}{n}}, p\log{d}\right\} \triangleq \eps_n \,.
$$
with probability $\ge 1 - d^{-\tilde K} - n^{-1} - e^{\log{n} -\frac{np}{c}}$ for some constant $c$ and $\tilde K$. 

Using this, we have: 

\begin{align*}
    \frac{\bv^\top \bZ^\top \bZ \bv}{n \|\bv\|^2} & = \frac{\bv^\top \Sigma_X \bv}{\|\bv\|^2} + \frac{\bv^\top (\bZ^\top \bZ/n - \Sigma_Z) \bv}{\|\bv\|^2} \\
    & \le \lambda_{\max}(\Sigma_X) +  \left\|\frac{\bZ^\top \bZ}{n} - \Sigma_X\right\|_\infty\frac{\|\bv\|_1^2}{\|\bv\|^2_2} \\
    & \le \lambda_{\max}(\Sigma_X) + \eps_n \frac{(1+\kappa)^2 s\|\bv\|_2^2}{\|\bv\|_2^2} \\
    & \le \lambda_{\max}(\Sigma_X) +(1+\kappa)^2  s\eps_n \,. 
\end{align*}

Here the penultimate inequality follows from the fact: 

$$
\|\bv\|_1 = \|\bv_S\|_1 + \|\bv_{S^c}\|_1 \le (1+k)\|\bv_S\|_1 \le (1+k)\sqrt{s}\|\bv_S\|_2 \,.
$$

Hence as soon as we assume $s\eps_n$ is bounded or goes to $0$ we are good. 

\subsection{Proof of Theorem~\ref{thm:trans_gcr}}\label{subsec:consistency}
We follow the proof of Theorem 1 in \cite{tian2023transfer} and extend and modify the results to allow for network dependency. Notice that we assume both source and target domains share the same ER graph probability, denoted as $p$.

\textbf{Step 1:}

Step 1 aims to solve the following equation w.r.t. ${\bbeta} \in \mathbb{R}^{d}$ :

$$
\sum_{k\in\{0,\mA\}}\left[\left(\boldsymbol{\bZ}^{(k)}\right)^{T} \boldsymbol{\bY}^{(k)}-\sum_{i=1}^{n_{k}} \psi^{\prime}\left({\bbeta}^{T} \boldsymbol{\bZ}_{i}^{(k)}\right) \boldsymbol{\bZ}_{i}^{(k)}\right]=\mathbf{0}_{p}
$$

converging to its population version's solution under certain conditions with $\alpha_{k}=\frac{n_{k}}{n_{\mathcal{A}}+n_{0}}$:

\begin{equation}
\label{2.12}
\sum_{k\in\{0,\mA\}} \alpha_{k} \mathbb{E}\left\{\left[\psi^{\prime}\left(\left({\bbeta}^{\mA}\right)^{T} \bZ^{(k)}\right)-\psi^{\prime}\left(\left({\bbeta}^{(k)}\right)^{T} {\bZ}^{(k)}\right)\right] \boldsymbol{\bZ}^{(k)}\right\}=\mathbf{0}_{p} 
\end{equation}

As we define before, $\hat{\bu} ^{\mA}=\hat{\bbeta}^{\mA}-{\bbeta}^{\mA}$ and $\mathcal{D}=\left\{\left(\boldsymbol{\bZ}^{(k)}, \bY^{(k)}\right)\right\}_{k\in\{0,\mA\}}$. Firstly, we claim that when $\lambda_{{{\bbeta}}} \geq 2\left\|\nabla L\left({\bbeta}^{\mA}, \mathcal{D}\right)\right\|_{\infty}$, it holds that with probability of at least $1-(\exp{\left(\frac12 \log{d} + 1-c_4 \log^2{d}\Psi(p)\right)} + 2\exp{\left(1-(c_1 -3/2)\log{d}\right)} + 2\exp{\left(\frac{3}{2}\log{d} + 1 - c_2 n\right)})$ that

\begin{equation}
\label{2.14}
\left\|\hat{{\bu}}^{\mA}\right\|_{2} \leq 8 \frac{C_4}{\kappa_l} C_{l} h \log{d} \sqrt{\frac{\Psi(p)}{n_{\mA}+n_{0}}}+3 \frac{\sqrt{s}}{\kappa_{1}} \lambda_{{\omega}}+2 \sqrt{\frac{C_{l}}{\kappa_{1}} h \lambda_{{\omega}}}
\end{equation}

According to the definition of $\hat{{\omega}}^{\mA}$, Hölder inequality and Lemma 1, we will have

\begin{align}
 \delta \hat{L}\left(\hat{{\bu}}^{\mA}, \mathcal{D}\right) & \leq \lambda_{{{\bbeta}}}\left(\left\|{\bbeta}_{S}^{\mA}\right\|_{1}+\left\|{\bbeta}_{S^{c}}^{\mA}\right\|_{1}\right)-\lambda_{{{\bbeta}}}\left(\left\|\hat{{\bbeta}}_{S}^{\mA}\right\|_{1}+\left\|\hat{{\bbeta}}_{S^{c}}^{\mA}\right\|_{1}\right)+\nabla \hat{L}({\bbeta}^{\mA}, \mathcal{D})^{T} \hat{{\bu}}^{\mA} \notag \\
& \leq \lambda_{{{\bbeta}}}\left(\left\|{\bbeta}_{S}^{\mA}\right\|_{1}+\left\|{\bbeta}_{S^{c}}^{\mA}\right\|_{1}\right)-\lambda_{{{\bbeta}}}\left(\left\|\hat{{\bbeta}}_{S}^{\mA}\right\|_{1}+\left\|\hat{{\bbeta}}_{S^{c}}^{\mA}\right\|_{1}\right)+\frac{1}{2} \lambda_{{{\bbeta}}}\left\|\hat{{\bu}}^{\mA}\right\|_{1}
\notag \\
& \leq \frac{3}{2} \lambda_{{{\bbeta}}}\left\|\hat{{\bu}}_{S}^{\mA}\right\|_{1}-\frac{1}{2} \lambda_{{{\bbeta}}}\left\|\hat{{\bu}}_{S^{c}}^{\mA}\right\|_{1}+2 \lambda_{{{\bbeta}}}\left\|{\bbeta}_{S^{c}}^{\mA}\right\|_{1}\notag \\
& \leq \frac{3}{2} \lambda_{{{\bbeta}}}\left\|\hat{{\bu}}_{S}^{\mA}\right\|_{1}-\frac{1}{2} \lambda_{{{\bbeta}}}\left\|\hat{{\bu}}_{S^{c}}^{\mA}\right\|_{1}+2 \lambda_{{{\bbeta}}} C_{l} h \label{eq:2.15}
\end{align}

If we assume that the claim we stated does not hold, we consider $\mathbb{C}=\left\{{\bu}: \frac{3}{2}\left\|{\bu}_{S}\right\|_{1}-\frac{1}{2}\left\|{\bu}_{S^{c}}\right\|_{1}+2 C_{l} h \geq 0\right\}$. By (\ref{eq:2.15}) and the convexity of $\hat{L}$, we conclude $\hat{{\bu}}^{\mA} \in \mathbb{C}$. Then for any $t \in(0,1)$, we can see that 

$$
\frac{1}{2}\left\|t \hat{{\bu}}_{S^{c}}^{\mA}\right\|_{1}=t \cdot \frac{1}{2}\left\|\hat{{\bu}}_{S^{c}}^{\mA}\right\|_{1} \leq t \cdot\left(\frac{3}{2}\left\|\hat{{\bu}}_{S}^{\mA}\right\|_{1}+2 C_{{{\bbeta}}} h\right) \leq \frac{3}{2}\left\|t \hat{{\bu}}_{S}^{\mA}\right\|_{1}+2 C_{l} h
$$

which also implies that $t \hat{{\bu}}^{\mA} \in \mathbb{C}$. There exists certain $t$ satisfying that $\left\|t \hat{{\bu}}^{\mA}\right\|_{2}>8 \kappa_{2} C_{l} h \sqrt{\frac{\log d}{n_{\mA}+n_{0}}}+$ $3 \frac{\sqrt{s}}{\kappa_{1}} \lambda_{{\omega}}+2 \sqrt{\frac{C_{l}}{\kappa_{1}} h \lambda_{{\omega}}}$ and $\left\|t \hat{{\bu}}^{\mA}\right\|_{2} \leq 1$. We denote $\tilde{{\bu}}^{\mA}=t \hat{{\bu}}^{\mA}$ and $F({\bu})=\hat{L}\left({\bbeta}^{\mA}+{\bu}, \mathcal{D}\right)-$ $\hat{L}\left({\bbeta}^{\mA}\right)+\lambda_{{{\bbeta}}}\left(\left\|{\bbeta}^{\mA}+{\bu}\right\|_{1}-\left\|{\bbeta}^{\mA}\right\|_{1}\right)$. As $F(\mathbf{0})=0$ and $F\left(\hat{{\bu}}^{\mA}\right) \leq 0$, by convexity, we establish

\begin{equation}
\label{eq:B.15}
F\left(\tilde{{\bu}}^{\mA}\right)=F\left(t \hat{{\bu}}^{\mA}+(1-t) \mathbf{0}\right) \leq t F\left(\hat{{\bu}}^{\mA}\right) \leq 0 
\end{equation}

However, by Lemma \ref{D1_lemma2} and the same trick we use for (\ref{eq:2.15}),

$$
\begin{aligned}
F\left(\tilde{{\bu}}^{\mA}\right) & \geq \delta \hat{L}\left(\hat{{\bu}}^{\mA}, \mathcal{D}\right)+\nabla \hat{L}\left({\bbeta}^{\mA}\right)^{T} \tilde{{\bu}}^{\mA}-\lambda_{{{\bbeta}}}\left\|{\bbeta}^{\mA}\right\|_{1}+\lambda_{{{\bbeta}}}\left\|{\bbeta}^{\mA}+\tilde{{\bu}}^{\mA}\right\|_{1} \\
& \geq L_\psi(T) \kappa_{l}\left\|\tilde{{\bu}}^{\mA}\right\|_{2}^{2}-L_\psi(T) \left(C_4 \log{d} \sqrt{\frac{\Psi(p)}{{n_{\mA}+n_{0}}}}\right)\left\|\tilde{{\bu}}^{\mA}\right\|_{1}\left\|\tilde{{\bu}}^{\mA}\right\|_{2}\\
& -\frac{3}{2} \lambda_{{{\bbeta}}}\left\|\tilde{{\bu}}_{S}^{\mA}\right\|_{1}+\frac{1}{2} \lambda_{{{\bbeta}}}\left\|\tilde{{\bu}}_{S^{c}}^{\mA}\right\|_{1}
-2 \lambda_{{{\bbeta}}} C_{l} h \\
& \geq L_\psi(T) \kappa_{l}\left\|\tilde{{\bu}}^{\mA}\right\|_{2}^{2}-L_\psi(T) C_4 \log{d} \sqrt{\frac{\Psi(p)}{{n_{\mA}+n_{0}}}}\left\|\tilde{{\bu}}^{\mA}\right\|_{1}\left\|\tilde{{\bu}}^{\mA}\right\|_{2}\\
& -\frac{3}{2} \lambda_{{{\bbeta}}}\left\|\tilde{{\bu}}_{S}^{\mA}\right\|_{1}-2 \lambda_{{{\bbeta}}} C_{l} h
\end{aligned}
$$

Due to $\tilde{{\bu}}^{\mA} \in \mathbb{C}$, it holds that

$$
\frac{1}{2}\left\|\tilde{{\bu}}^{\mA}\right\|_{1} \leq 2\left\|\tilde{{\bu}}_{S}^{\mA}\right\|_{1}+2 C_{{{\bbeta}}} h \leq 2 \sqrt{s}\left\|\tilde{{\bu}}^{\mA}\right\|_{2}+2 C_{l} h
$$

Here, we denote $\kappa_1=L_\psi(T) \kappa_{l}$, $\kappa_2=C_4 \log{d}/\kappa_l$, when $n_{\mA}+n_{0}>16 \kappa_{2}^{2} s \Psi(p)$, we have $2 \frac{C_4}{\kappa_l} \log{d} \sqrt{\frac{s \Psi(p)}{n_{\mA}+n_{0}}} \leq \frac{1}{2}$.Then it follows

$$
\begin{aligned}
    F\left(\tilde{{\bu}}^{\mA}\right) \geq \frac{1}{2} \kappa_{1}\left\|\tilde{{\bu}}^{\mA}\right\|_{2}^{2}-\left[2 \kappa_{1} \kappa_{2} \sqrt{\frac{\Psi(p)}{n_{\mA}+n_{0}}} C_{{{\bbeta}}} h+\frac{3}{2} \lambda_{{{\bbeta}}} \sqrt{s}\right]\left\|\tilde{{\bu}}^{\mA}\right\|_{2}-2 \lambda_{{{\bbeta}}} C_{l}h = \\
\frac{1}{2} \kappa_{1}\left\|\tilde{{\bu}}^{\mA}\right\|_{2}^{2}-\left[2 \kappa_{1} \frac{C_4}{\kappa_l} \log{d} \sqrt{\frac{\Psi(p)}{n_{\mA}+n_{0}}} C_{{{\bbeta}}} h+\frac{3}{2} \lambda_{{{\bbeta}}} \sqrt{s}\right]\left\|\tilde{{\bu}}^{\mA}\right\|_{2}-2 \lambda_{{{\bbeta}}} C_{l}h> 0  
\end{aligned}
$$

that conflicts with (\ref{eq:B.15}). Thus our claim at the beginning holds.

Next, we will prove $\left\|\nabla \hat{L}\left({\bbeta}^{\mA}\right)\right\|_{\infty} \lesssim \sqrt{\frac{\log d}{n_{\mA}+n_{0}}}$ with probability at least $1- (c_{12} d^{-c_{13}} + \sum_k n_k^{-1} + \sum_k e^{\log{n_k} -\frac{n_k p_k}{c_{14}}})$. To see this, we notice that

\begin{align}
\nabla \hat{L}\left({\bbeta}^{\mA}\right)= & \frac{1}{n_{\mA}+n_{0}} \sum_{k\in\{0,\mA\}}\left(\boldsymbol{\bZ}^{(k)}\right)^{T}\left[-\bY^{(k)}+\boldsymbol{\psi}^{\prime}\left(\boldsymbol{\bZ}^{(k)} {\bbeta}^{\mA}\right)\right] \notag\\
= & \frac{1}{n_{\mA}+n_{0}} \sum_{k\in\{0,\mA\}}\left(\boldsymbol{\bZ}^{(k)}\right)^{T}\left[-\bY^{(k)}+\boldsymbol{\psi}^{\prime}\left(\boldsymbol{\bZ}^{(k)} \boldsymbol{\beta}^{(k)}\right)\right] \notag \\
& +\frac{1}{n_{\mA}+n_{0}} \sum_{k\in\{0,\mA\}}\left(\boldsymbol{\bZ}^{(k)}\right)^{T}\left[-\boldsymbol{\psi}^{\prime}\left(\boldsymbol{\bZ}^{(k)} \boldsymbol{\beta}^{(k)}\right)+\boldsymbol{\psi}^{\prime}\left(\boldsymbol{\bZ}^{(k)} {\bbeta}^{\mA}\right)\right] \label{B.16}
\end{align}

By extending Lemma 6 of \cite{negahban2009unified} for network dependency in our settings, under Assumptions \ref{assump2} and the fact $n_{\mA} \geq C s \log d$, we have shown in Lemma \ref{D1_lemma3} that

$$
\frac{1}{n_{\mA}+n_{0}}\left\|\sum_{k\in\{0,\mA\}}\left(\boldsymbol{\bZ}^{(k)}\right)^{T}\left[-\bY^{(k)}+\boldsymbol{\psi}^{\prime}\left(\boldsymbol{\bZ}^{(k)} {\bbeta}^{(k)}\right)\right]\right\|_{\infty} \lesssim \sqrt{\frac{\log d}{n_{\mA}+n_{0}}}
$$

with probability at least $1- (c_6 d^{-c_7} + \sum_k n_k^{-1} + \sum_k e^{\log{n_k} -\frac{n_k p_k}{c_8}})$.

The remaining work aims to bound the infinity norm of the second term in (\ref{B.16}). We denote $U_{i j}^{(k)}=\bZ_{i j}^{(k)}\left[-\psi^{\prime}\left(\left(\bZ_{i}^{(k)}\right)^{T} {\bbeta}^{(k)}\right)+\psi^{\prime}\left(\left(\bZ_{i}^{(k)}\right)^{T} {\bbeta}^{\mA}\right)\right]$. Under Assumption \ref{assump2}, we have shown in Conjecture \ref{D1_conjecture} that:

$$
\frac{1}{n_{\mA}+n_{0}} \sup _{j=1, \ldots, d}\left|\sum_{k\in\{0,\mA\}} \sum_{i=1}^{n_{k}} U_{i j}^{(k)}\right| \lesssim  \sqrt{\frac{\log d}{n_{\mA}+n_{0}}}
$$
with probability at least $1- (c_9 d^{-c_{10}} + \sum_k n_k^{-1} + \sum_k e^{\log{n_k} -\frac{n_k p_k}{c_{11}}})$.  Hence $\left\|\nabla \hat{L}\left({\bbeta}^{\mA}\right)\right\|_{\infty} \lesssim \sqrt{\frac{\log d}{n_{\mA}+n_{0}}}$ holds with probability at least $1- (c_{12} d^{-c_{13}} + \sum_k n_k^{-1} + \sum_k e^{\log{n_k} -\frac{n_k p_k}{c_{14}}})$. 
We plug this rate into (\ref{2.14}), and get

\begin{equation}\label{B.17}
\left\|\hat{{\bu}}^{\mA}\right\|_{2} \lesssim h \log{d} \sqrt{\frac{\Psi(p)}{n_{\mA}+n_{0}}}+\sqrt{\frac{s \log d}{n_{\mA}+n_{0}}}+\left(\frac{\log d}{n_{\mA}+n_{0}}\right)^{1 / 4} \sqrt{h} 
\end{equation}

with probability at least $1- (c_{12} d^{-c_{13}} + \sum_k n_k^{-1} + \sum_k e^{\log{n_k} -\frac{n_k p_k}{c_{14}}})$ when $\lambda_{{{\bbeta}}} \asymp C_{{{\bbeta}}} \sqrt{\frac{\log d}{n_{\mA}+n_{0}}}$ with $C_{{{\bbeta}}}>0$ sufficiently large. As $\hat{{\bu}}^{\mA} \in \mathbb{C}$, (\ref{B.17}) implies

\begin{equation}\label{B.18}
\left\|\hat{{\bu}}^{\mA}\right\|_{1} \lesssim s \sqrt{\frac{\log d}{n_{\mA}+n_{0}}}+\left(\frac{\log d}{n_{\mA}+n_{0}}\right)^{1 / 4} \sqrt{s h}+h\left(1+\log d\sqrt{\frac{s \Psi(p)}{n_{\mA}+n_{0}}}\right) 
\end{equation}

with probability at least $1- (c_{12} d^{-c_{13}} + \sum_k n_k^{-1} + \sum_k e^{\log{n_k} -\frac{n_k p_k}{c_{14}}})$.

\textbf{Step 2:}\\
For our convenience, we state the notation again here: 
$\mathcal{D}^{(0)}=\left(\boldsymbol{\bZ}^{(0)}, \bY^{(0)}\right)$, $\hat{L}^{(0)}\left(\bbeta, \mathcal{D}^{(0)}\right) = -\frac{1}{n_{0}}\left(\bY^{(0)}\right)^{T} \bZ^{(0)} \bbeta +\frac{1}{n_{0}} \sum_{i=1}^{n_{0}} \psi\left(\left(\bZ_{i}^{(0)}\right)^{T} \bbeta\right)$, $\nabla \hat{L}^{(0)}\left(\bbeta, \mathcal{D}^{(0)}\right)=-\frac{1}{n_{0}}\left(\bZ^{(0)}\right)^{T} \bY^{(0)}+\frac{1}{n_{0}}\left(\bZ^{(0)}\right)^{T} \boldsymbol{\psi}^{\prime}\left(\bZ^{(0)} \bbeta\right), {\delta}^{\mA}=$ ${\bbeta}^{(0)}-\bbeta^{\mA}, \hat{{\bbeta}}^{(0)}=\hat{\bbeta}^{\mA}+\hat{{\bdelta}}^{\mA}, \hat{{\bv}}^{\mA}=\hat{{\bdelta}}^{\mA}-{\bdelta}^{\mA}$, and $\delta \hat{L}^{(0)}({\bdelta}, \mathcal{D})=\hat{L}^{(0)}\left(\hat{\bbeta}^{\mA}+{\bdelta}, \mathcal{D}^{(0)}\right) - \hat{L}^{(0)}\left(\hat{\bbeta}^{\mA}+{\bdelta}^{\mA}, \mathcal{D}^{(0)}\right)-\nabla \hat{L}^{(0)}\left(\hat{\bbeta}^{\mA}+{\bdelta}^{\mA}, \mathcal{D}^{(0)}\right)^{T} \hat{{\bv}}^{\mA}$.

Following similar derivations for (\ref{eq:2.15}), when $\lambda_{\delta} \geq 2\left\|\nabla \hat{L}^{(0)}\left({\bbeta}^{(0)}, \mathcal{D}^{(0)}\right)\right\|_{\infty}$, we establish

\begin{align}
\delta \hat{L}^{(0)}\left(\hat{{\bdelta}}^{\mA}, \mathcal{D}\right) \leq & \lambda_{{\delta}}\left(\left\|{\bdelta}^{\mA}\right\|_{1}-\left\|\hat{{\bdelta}}^{\mA}\right\|_{1}\right)-\nabla \hat{L}^{(0)}\left(\hat{\bbeta}^{\mA}+{\bdelta}^{\mA}, \mathcal{D}^{(0)}\right)^{T} \hat{{\bv}}^{\mA} \notag \\
& \leq  \lambda_{{\bdelta}}\left(2\left\|{\bdelta}^{\mA}\right\|_{1}-\left\|\hat{{\bv}}^{\mA}\right\|_{1}\right)+\left\|\nabla \hat{L}^{(0)}\left({\bbeta}^{(0)}, \mathcal{D}^{(0)}\right)\right\|_{\infty}\left\|\hat{{\bv}}^{\mA}\right\|_{1} \notag \\
& -\left[\nabla \hat{L}^{(0)}\left(\hat{\bbeta}^{\mA}+{\bdelta}^{\mA}, \mathcal{D}^{(0)}\right)-\nabla \hat{L}^{(0)}\left({\bbeta}^{(0)}, \mathcal{D}^{(0)}\right)\right]^{T} \hat{{\bv}}^{\mA} \notag \\
& \leq  2  \lambda_{{\delta}}\left\|{\bdelta}^{\mA}\right\|_{1}-\frac{1}{2} \lambda_{{\delta}}\left\|\hat{{\bv}}^{\mA}\right\|_{1}\notag \\ 
& -\frac{1}{n_{0}}\left[\boldsymbol{\psi}^{\prime}\left(\left({\bZ}^{(0)}\right)^{T}\left(\hat{\bbeta}^{\mA}+{\bdelta}^{\mA}\right)\right)-\boldsymbol{\psi}^{\prime}\left(\left(\boldsymbol{\bZ}^{(0)}\right)^{T} {\bbeta}^{(0)}\right)\right]^{T} \hat{{\bv}}^{\mA} \notag\\
&\leq  2 \lambda_{{\delta}}\left\|{\bdelta}^{\mA}\right\|_{1}-\frac{1}{2} \lambda_{{\delta}}\left\|\hat{{\bv}}^{\mA}\right\|_{1}+\frac{1}{4 c_{0}} M_{\psi}^{2} \cdot \frac{1}{n_{0}}\left\|{\bZ}^{(0)} \hat{{\bu}}^{\mA}\right\|_{2}^{2}\notag\\
& +c_{0} \cdot \frac{1}{n_{0}}\left\|{\bZ}^{(0)} \hat{{\bv}}^{\mA}\right\|_{2}^{2} \label{eq:2.20}
\end{align}

with $c_{0}>0$ being a constant that is enough small. The last inequality holds according to:

\begin{align*}
& -\frac{1}{n_{0}}\left[\boldsymbol{\psi}^{\prime}\left(\left(\boldsymbol{\bZ}^{(0)}\right)^{T}\left(\hat{\bbeta}^{\mA}+{\bdelta}^{\mA}\right)\right)-\boldsymbol{\psi}^{\prime}\left(\left({\bZ}^{(0)}\right)^{T} {\bbeta}^{(0)}\right)\right]^{T} \hat{{\bv}}^{\mA} \\
& =\frac{1}{n_{0}}\left(\hat{\bu}^{\mA}\right)^{T}\left(\boldsymbol{\bZ}^{(0)}\right)^{T} \Lambda^{(0)} \boldsymbol{\bZ}^{(0)} \hat{{\bv}}^{\mA} \\
& \leq \frac{1}{4 c_{0}} M_{\psi}^{2} \cdot \frac{1}{n_{0}}\left\|\boldsymbol{\bZ}^{(0)} \hat{{\bu}}^{\mA}\right\|_{2}^{2}+c_{0} \cdot \frac{1}{n_{0}}\left\|\boldsymbol{\bZ}^{(0)} \hat{{\bv}}^{\mA}\right\|_{2}^{2}
\end{align*}

where $\Lambda^{(0)}=\operatorname{diag}\left(\left\{\psi^{\prime \prime}\left(\left(\boldsymbol{\bZ}_{i}^{(0)}\right)^{T} {\bbeta}^{(0)}+t_{i}\left(\boldsymbol{\bZ}_{i}^{(0)}\right)^{T} \hat{{\bu}}^{\mA}\right)\right\}_{i=1}^{n_{0}}\right)$ is a $n_{0} \times n_{0}$ diagonal matrix and $\left\|\Lambda^{(0)}\right\|_{\max } \leq M_{\psi}$.

We denote $\tilde{{\bv}}^{\mA}=t \hat{\bv}^{\mA}$ and similar to what we defined before, let $F^{(0)}({\bv})=\hat{L}^{(0)}\left(\hat{\bbeta}^{\mA}+{\bdelta}^{\mA}+\right.$ $\left.{\bv}, \mathcal{D}^{(0)}\right)-\hat{L}^{(0)}\left(\hat{\bbeta}^{\mA}+{\bdelta}^{\mA}, \mathcal{D}^{(0)}\right)+\lambda_{{\delta}}\left(\left\|{\bdelta}^{\mA}+{\bv}\right\|_{1}-\left\|{\bdelta}^{\mA}\right\|_{1}\right)$. As $F(\mathbf{0})=0$ and $F^{(0)}\left(\hat{{\bv}}^{\mA}\right) \leq$ 0 , by convexity, for any $t \in(0,1]$, we establish

\begin{equation}\label{2.21}
F^{(0)}\left(\tilde{{\bv}}_{\mA}\right)=F^{(0)}\left(t \hat{{\bv}}_{\mA}+(1-t) \mathbf{0}\right) \leq t F^{(0)}\left(\hat{{\bu}}^{\mA}\right) \leq 0 
\end{equation}

Setting $t \in(0,1]$ ensures that $\left\|\tilde{\boldsymbol{v}}^{\mA}\right\|_{2} \leq 1$. By noticing the fact that $ \left\|\tilde{\boldsymbol{v}}^{\mA}\right\|_{2} \le \left\|\tilde{\boldsymbol{v}}^{\mA}\right\|_{1} $, we can apply Lemma \ref{D1_lemma2} on $\tilde{\boldsymbol{v}}^{\mA}$ with minor modifications. Also by (\ref{2.21}) and (\ref{eq:2.20}), we establish: 

\begin{align} 
  \kappa_1\left\|\tilde{{\bv}}^{\mA}\right\|_{2}^{2}-\kappa_1\kappa_3 \left(\log{d} \sqrt{\frac{\Psi(p)}{{n_{0}}}}\right) \cdot \left\|\tilde{{\bv}}^{\mA}\right\|_{1}^{2} 
   \leq & \notag\\
   F^{(0)}\left(\tilde{{\bv}}^{\mA}\right)
   -\nabla \hat{L}^{(0)}\left(\hat{\bbeta}^{\mA}+{\bdelta}^{\mA}, \mathcal{D}^{(0)}\right)^{T} \tilde{{\bv}}^{\mA} 
 & \leq 2 \lambda_{{\delta}}\left\|{\bdelta}^{\mA}\right\|_{1}-\frac{1}{2} \lambda_{{\delta}}\left\|\tilde{{\bv}}^{\mA}\right\|_{1}+\notag\\
 \frac{1}{4 c_{0}} M_{\psi}^{2} \cdot \frac{1}{n_{0}}\left\|{\bZ}^{(0)} \hat{{\bu}}^{\mA}\right\|_{2}^{2} 
& +c_{0} \cdot \frac{1}{n_{0}}\left\|{\bZ}^{(0)} \tilde{{\bv}}^{\mA}\right\|_{2}^{2}\label{2:22}
\end{align}

with $\kappa_1=L_\psi(T) \kappa_{l}$, $\kappa_3=C_4/\kappa_l$.

We showed in the proof of Step 1 that $\left\|\hat{{\bu}}_{S^{c}}^{\mA}\right\|_{1} \leq 3\left\|\hat{{\bu}}_{S}^{\mA}\right\|_{1}+4 C_{l} h$. Next we discuss about bounding $\frac{1}{n_{0}}\left\|{\bZ}^{(0)} \hat{{\bu}}^{\mA}\right\|_{2}^{2}$ by $\left\| \hat{{\bu}}^{\mA}\right\|_{2}^{2}$ using this fact.

If $3\left\|\hat{{\bu}}_{S}^{\mA}\right\|_{1} \geq 4 C_{l} h$, then $\left\|\hat{{\bu}}_{S^{c}}^{\mA}\right\|_{1} \leq 6\left\|\hat{{\bu}}_{S}^{\mA}\right\|_{1}$. Then by Lemma \ref{D1_lemma5} (the extension on Theorem 1.6 of \cite{zhou2009restricted} for network dependency), we have

\begin{equation}\label{2:23}
\frac{1}{n_{0}}\left\|{\bZ}^{(0)} \hat{{\bu}}^{\mA}\right\|_{2}^{2} \lesssim\left\|\hat{{\bu}}^{\mA}\right\|_{2}^{2} \lesssim \frac{s \log d}{n_{\mathcal{A}}+n_{0}}+h \cdot \sqrt{\frac{\log d}{n_{\mathcal{A}}+n_{0}}}
\end{equation}

with probability at least $1 - d^{-\tilde K} - n_{0}^{-1} - e^{\log{n_{0}} -\frac{n_{0}p}{c}}$.

If $3\left\|\hat{{\bu}}_{S}^{\mA}\right\|_{1}<4 C_{l} h$, then $\left\|\hat{{\bu}}_{S^{c}}^{\mA}\right\|_{1} \leq 8 C_{l} h \leq \sqrt{s}$. Also $\left\|\hat{{\bu}}^{\mA}\right\|_{2} \leq 1$ with probability $1- (c_{12} d^{-c_{13}} + \sum_k n_k^{-1} + \sum_k e^{\log{n_k} -\frac{n_k p_k}{c_{14}}})$. We denote

$$
\begin{aligned}
& \Pi_{0}(s)=\left\{{\bu} \in \mathbb{R}^{p}:\|{\bu}\|_{2} \leq 1,\|{\bu}\|_{0} \leq s\right\} \\
& \Pi_{1}(s)=\left\{{\bu} \in \mathbb{R}^{p}:\|{\bu}\|_{2} \leq 1,\|{\bu}\|_{1} \leq \sqrt{s}\right\}
\end{aligned}
$$

Due to Lemma 3.1 of \cite{plan2013one}, $\Pi_{1}(s) \subseteq 2 \overline{\operatorname{conv}}\left(\Pi_{0}(s)\right)$, where $\overline{\operatorname{conv}}\left(\Pi_{0}(s)\right)$ is the closure of convex hull of $\Pi_{0}(s)$. Similarly, an extension for network dependency on the proof of Theorem 2.4 in \cite{mendelson2008uniform} will also conclude (\ref{2:23}).

Next we bound $\frac{1}{n_{0}}\left\|\boldsymbol{\bZ}^{(0)} \tilde{{\bv}}^{\mA}\right\|_{2}^{2}$ by $\left\|\tilde{{\bv}}^{\mA}\right\|_{2}^{2}$. From basic inequality, we establish

\begin{align}
0 & \leq \hat{L}^{(0)}\left(\hat{{\bbeta}}^{(0)}, \mathcal{D}^{(0)}\right)-\hat{L}^{(0)}\left({\bbeta}^{(0)}, \mathcal{D}^{(0)}\right)-\nabla \hat{L}^{(0)}\left({\bbeta}^{(0)}, \mathcal{D}^{(0)}\right)^{T}(\hat{{\bbeta}}^{(0)}-{\bbeta}^{(0)}) \notag \\
& \leq \lambda_{{\delta}}\left(\left\|{\bbeta}^{(0)}-\hat{\bbeta}^{\mA}\right\|_{1}-\left\|\hat{{\bdelta}}^{\mA}\right\|_{1}\right)+\left\|\nabla \hat{L}^{(0)}\left({\bbeta}^{(0)}, \mathcal{D}^{(0)}\right)\right\|_{\infty}\|\hat{{\bbeta}}^{(0)}-{\bbeta}^{(0)}\|_{1}  \notag \\
& \leq \lambda_{{\delta}}\left(\left\|{\bbeta}^{(0)}-\hat{\bbeta}^{\mA}\right\|_{1}-\left\|\hat{{\bdelta}}^{\mA}\right\|_{1}\right)+\frac{1}{2} \lambda_{{\delta}}\|\hat{{\bbeta}}^{(0)}-{\bbeta}^{(0)}\|_{1} \notag \\
& \leq \frac{3}{2} \lambda_{{\delta}}\left\|{\bbeta}^{(0)}-\hat{\bbeta}^{\mA}\right\|_{1}-\frac{1}{2} \lambda_{{\delta}}\left\|\hat{{\bdelta}}^{\mA}\right\|_{1} \notag \\
& \leq \frac{3}{2} \lambda_{{\delta}} C_{l} h+\frac{3}{2} \lambda_{{\delta}}\left\|\hat{{\bu}}^{\mathcal{A}}\right\|_{1}-\frac{1}{2} \lambda_{{\delta}}\left\|\hat{{\bdelta}}^{\mA}\right\|_{1}
\end{align}

implying
$$
\left\|\hat{{\bv}}^{\mA}\right\|_{1} \leq\left\|\hat{{\bdelta}}^{\mA}\right\|_{1}+C_{l} h \leq 3\left\|\hat{{\bu}}^{\mathcal{A}}\right\|_{1}+4 C_{l} h
$$

Combined with results by (\ref{B.18}), we have $\left\|\tilde{{\bv}}^{\mA}\right\|_{1} \leq\left\|\hat{{\bv}}^{\mA}\right\|_{1} \leq \sqrt{s}$ when $s \log d /\left(n_{\mA}+n_{0}\right)$ and $h$ are small enough. We can see $\delta \hat{L}^{(0)}\left(\hat{{\bdelta}}^{\mA}, \mathcal{D}\right)>0$ from the strict convexity, which leads to $\left\|\hat{{\bdelta}}^{\mA}\right\|_{1} \leq$ $3\left\|\hat{{\bu}}^{\mA}\right\|_{1}+3 h$. Then we get

\begin{equation}
\left\|\hat{{\bv}}^{\mA}\right\|_{1} \leq\left\|{\bbeta}^{(0)}-\hat{\bbeta}^{\mA}\right\|_{1}+\left\|\hat{{\bdelta}}^{\mA}\right\|_{1} \leq 4\left\|\hat{{\bu}}^{\mA}\right\|_{1}+4 h \leq \sqrt{s} \label{2.24}
\end{equation}

Similar to the analysis considering $3\left\|\hat{{\bu}}_{S}^{\mA}\right\|_{1}<4 C_{1} h$ above, we establish

$$
c_{0} \cdot \frac{1}{n_{0}}\left\|{\bZ}^{(0)} \tilde{{\bv}}^{\mA}\right\|_{2}^{2} \leq c_{0} \cdot C\left\|\tilde{{\bv}}^{\mA}\right\|_{2}^{2}
$$

holds with probability at least $1 - d^{-\tilde K} - n_{0}^{-1} - e^{\log{n_{0}} -\frac{n_{0}p}{c}}$. As long as $c_{0} C<c_{9} / 2$, by (\ref{2:22}), we have

\begin{align}
&\kappa_1\left\|\tilde{{\bv}}^{\mA}\right\|_{2}^{2}-\kappa_1\kappa_3 \left(\log{d} \sqrt{\frac{\Psi(p)}{{n_{0}}}}\right) \cdot \left\|\tilde{{\bv}}^{\mA}\right\|_{1}^{2} \notag\\
\leq & 2 \lambda_{{\delta}}\left\|{\mathcal { \bdelta}}^{\mA}\right\|_{1}-\frac{1}{2} \lambda_{{\delta}}\left\|\tilde{{\bv}}^{\mA}\right\|_{1}+C \frac{s \log d}{n_{\mA}+n_{0}}+C h \sqrt{\frac{\log d}{n_{\mA}+n_{0}}}+c_{3} / 2\left\|\tilde{{\bv}}^{\mA}\right\|_{2}^{2} \label{B.25}
\end{align}

with probability at least \textcolor{black}{$1-C^{\prime} n_{0}^{-1}$}, $\kappa_1=L_\psi(T) \kappa_{l}$, $\kappa_3=C_4/\kappa_l$.

If satisfying $\lambda_{{\delta}}\left\|{\bdelta}^{\mA}\right\|_{1}  \leq C \frac{s \log d}{n_{\mA}+n_{0}}+C h \sqrt{\frac{\log d}{n_{\mA}+n_{0}}}$, then

\begin{align*}
    \left\|\tilde{{\bv}}^{\mA}\right\|_{1} \lesssim\left[\frac{s \log d}{n_{\mA}+n_{0}}+h \sqrt{\frac{\log d}{n_{\mA}+n_{0}}}\right] \cdot \sqrt{\frac{1}{\log d}\sqrt{\frac{n_{0}}{\Psi(p)}}}+\left\|\tilde{{\bv}}^{\mA}\right\|_{2}^{2}
\end{align*}

Because $\left\|\tilde{{\bv}}^{\mA}\right\|_{2} \leq 1$, by (\ref{B.25}), the following inequality holds

$$
\left\|\tilde{{\bv}}^{\mA}\right\|_{2}^{2} \lesssim \frac{s \log d}{n_{\mA}+n_{0}}+h \sqrt{\frac{\log d}{n_{\mA}+n_{0}}} \lesssim \frac{s \log d}{n_{\mA}+n_{0}}+\left[h \sqrt{\frac{\log d}{n_{0}}}\right] \wedge h^{2}
$$

with probability at least $1-C^{\prime} n_{0}^{-1}$.

If $\lambda_{\delta}\left\|{\bdelta}^{\mA}\right\|_{1}>C \frac{s \log d}{n_{\mA}+n_{0}}+C h \sqrt{\frac{\log d}{n_{\mA}+n_{0}}}$, then $\left\|\tilde{{\bv}}^{\mA}\right\|_{1} \lesssim h+\left\|\tilde{{\bv}}^{\mA}\right\|_{2}^{2}$, leading to

$$
\left\|\tilde{{\bv}}^{\mA}\right\|_{2}^{2} \lesssim 2 \lambda_{{\delta}}\left\|{\bdelta}^{\mA}\right\|_{1}-\frac{1}{2} \lambda_{{\delta}}\left\|\tilde{{\bv}}^{\mA}\right\|_{1}
$$

 which implies $\left\|\tilde{{\bv}}^{\mA}\right\|_{1} \leq 4\left\|{\bdelta}^{\mA}\right\|_{1} \leq 4 C_{l} h$. By plugging this result into (\ref{B.25}), we obtain

 \begin{equation}
\left\|\tilde{{\bv}}^{\mA}\right\|_{2}^{2} \lesssim \frac{s \log d}{n_{\mA}+n_{0}}+\left[h \sqrt{\frac{\log d}{n_{0}}}\right] \wedge h^{2} \label{b27}
\end{equation}

\textcolor{black}{with probability at least $1-C^{\prime} n_{0}^{-1}$.}

When $s \log d /\left(n_{\mA}+n_{0}\right)$ and $h$ is small enough, due to $h \sqrt{\frac{\log d}{{n_{0}}}}=o(1)$, the right side of (\ref{b27}) can be very small, implying $\left\|\tilde{{\bv}}^{\mA}\right\|_{2} \leq c<1$ with probability at least $1-C^{\prime} n_{0}^{-1}$. We should notice that this result holds for any $t \in(0,1]$ such that $\left\|\tilde{{\bv}}^{\mA}\right\|_{2} \leq 1$. Finally let's consider the vector of interest: $\hat{{\bv}}^{\mA}$. Suppose $\left\|\tilde{{\bv}}^{\mA}\right\|_{2}^{2} \geq \frac{s \log d}{n_{\mA}+n_{0}}+\left[h {\left(\sqrt{\frac{\log d}{{n_{0}}}}\right)}\right] \wedge h^{2}$ for some constant $C>0$ with probability at least $C^{\prime} n_{0}^{-1}$, then there exists $t \in(0,1]$ such that $c<\left\|\tilde{{\bv}}^{\mA}\right\|_{2} \leq 1$. This contradicts with the fact $\left\|\tilde{{\bv}}^{\mA}\right\|_{2} \leq c$ with probability at least $1-C^{\prime} n_{0}^{-1}$. Hence we establish

$$
\left\|\hat{{\bv}}^{\mA}\right\|_{2}^{2} \lesssim \frac{s \log d}{n_{\mA}+n_{0}}+\left[h \sqrt{\left({\frac{\log d}{{n_{0}}}}\right)}\right] \wedge h^{2} \lesssim 1
$$

\textcolor{black}{with probability at least $1-C^{\prime} n_{0}^{-1}$. }

Similarly, the $\ell_{1}$-bound on $\hat{{\bv}}^{\mA}$ will be obtained by going over the analysis procedure of $\tilde{{\bv}}^{\mA}$ 

$$
\left\|\hat{{\bv}}^{\mA}\right\|_{1} \lesssim \left[s \sqrt{\frac{\log d}{n_{\mA}+n_{0}}}+h\right] \cdot \sqrt{\sqrt{\frac{1}{\Psi(p)}}}
$$

\textcolor{black}{with probability at least $1-C^{\prime} n_{0}^{-1}$.}

Lastly, we combine the conclusions in this Step 2 with the upper bounds on $\left\|\hat{{\bu}}^{\mA}\right\|_{2}$ and $\left\|\hat{{\bu}}^{\mA}\right\|_{1}$ in Step 1, to complete the proof.

Combining the above inequalities, we obtain:

$$
\begin{aligned}
\left\|\hat{\bbeta}^{(0)} - \bbeta ^{(0)}\right\|_{2} & \leq \left\|\hat{\bu} \right\|_2 + \left\|\hat{\bv} \right\|_2 \lesssim   h \log{d} \sqrt{\frac{\Psi(p)}{n_{\mA}+n_{0}}}+\sqrt{\frac{s \log d}{n_{\mA}+n_{0}}}+\left(\frac{\log d}{n_{\mA}+n_{0}}\right)^{1 / 4} \sqrt{h} .
\end{aligned}
$$

And

$$
\begin{aligned}
	\left\|\hat{\bbeta}^{(0)} - \bbeta^{(0)} \right\|_{1} & \leq \left\|\hat{\bu} \right\|_1 + \left\|\hat{\bv} \right\|_1 \lesssim  s \sqrt{\frac{\log d}{n_{\mA}+n_{0}}}+\left(\frac{\log d}{n_{\mA}+n_{0}}\right)^{1 / 4} \sqrt{s h}+ \\& h\left(1+\log d\sqrt{\frac{s \Psi(p)}{n_{\mA}+n_{0}}}\right)+\left[s \sqrt{\frac{\log d}{n_{\mA}+n_{0}}}+h\right] \cdot \sqrt{\sqrt{\frac{1}{\Psi(p)}}}.
\end{aligned}
$$

\section{Additional Experimental Results}\label{sec:results_add}
In this section, we first conduct additional simulation studies considering other network models such as SBM and graphon models, and also consider multiple convolution layers $M=2$ (see in Appendix \ref{sec:simu_add}). Subsequently, we present complete results of real data analyses including the summary of averaged Macro F1 and standard deviation of Micro and Macro F1 scores for transfer learning tasks, as well as the performance considering varying source and target data training rate in Appendix \ref{sec:real_add}.

\subsection{Additional Simulation  Results}\label{sec:simu_add}
Here, we conduct simulation studies considering similar settings in Section \ref{sec:5.1} but generating the adjacency matrices from SBM or graphon models.

\begin{figure}[!h]
    \centering
    \includegraphics[scale=0.4]{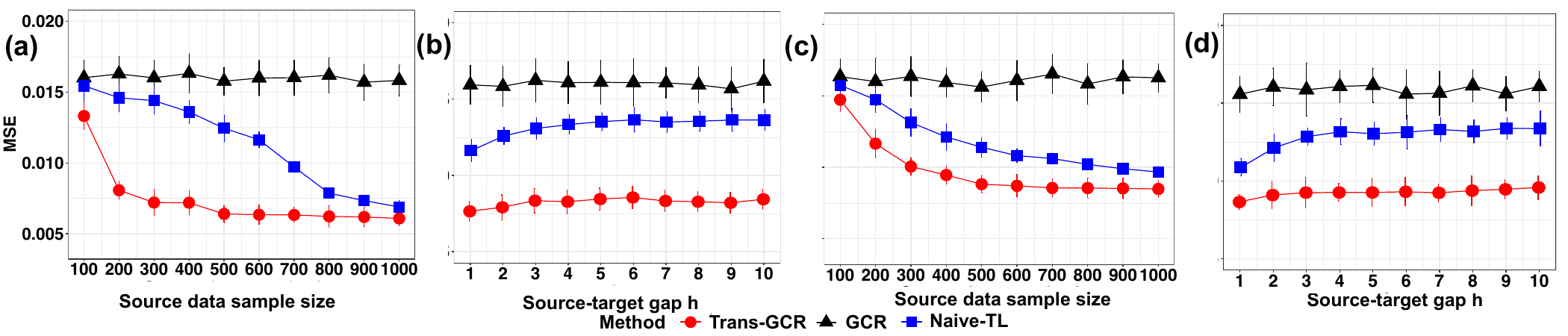}
     \caption{Performance comparison (MSE) of Trans-$\our$ (red), $\our$ (black), Naive TL (blue) across varying (a)(c) Source sample size, (b)(d) Source-target gap $h$, for two additional SBM models.}
    \label{fig:append-sbm}
\end{figure}

Figure \ref{fig:append-sbm} presents the results for SBM models. Figures \ref{fig:append-sbm} (a)(b) were performed when SBM generated the adjacency matrices of both target and source domains with between-community connection probability as 0.08 and the within-community probability as 0.1. Figures \ref{fig:append-sbm} (c)(d) were performed when the adjacency matrices of both target and source domains were generated by SBM with between-community connection probability as 0.08 and within-community probability as 0.04.

\begin{figure}[!h]
    \centering
    \includegraphics[scale=0.4]{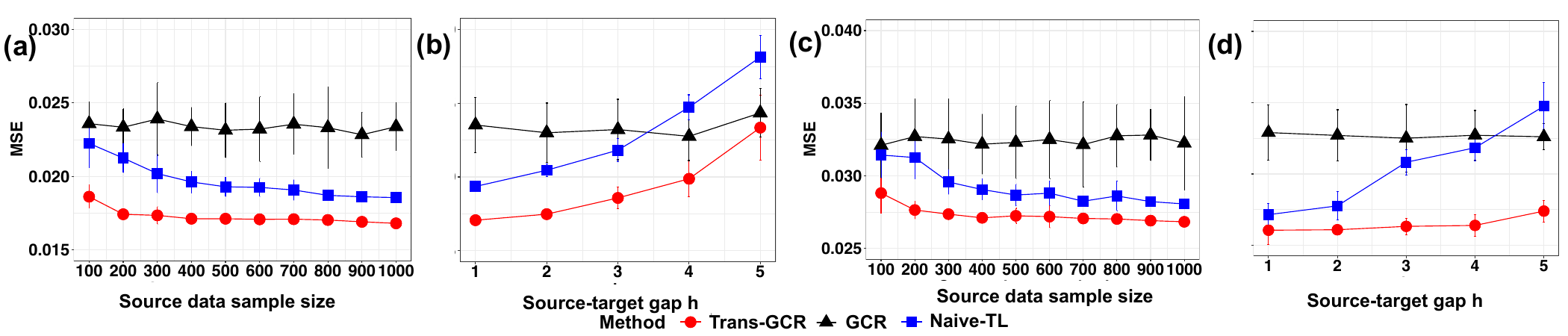}
       \caption{Performance comparison (MSE) of Trans-$\our$ (red), $\our$ (black), Naive TL (blue) across varying (a)(c) Source sample size, (b)(d) Source-target gap $h$, for two additional graphon models. }
    \label{fig:append-graphon}
\end{figure}

Figure \ref{fig:append-graphon} presents the results for graphon models. Figures \ref{fig:append-graphon} (a)(b), and (c)(d) were performed with the adjacency matrices of both target and source domains generated by two types of graphons, respectively.

\begin{figure}[!h]
    \centering
    \includegraphics[scale=0.37]{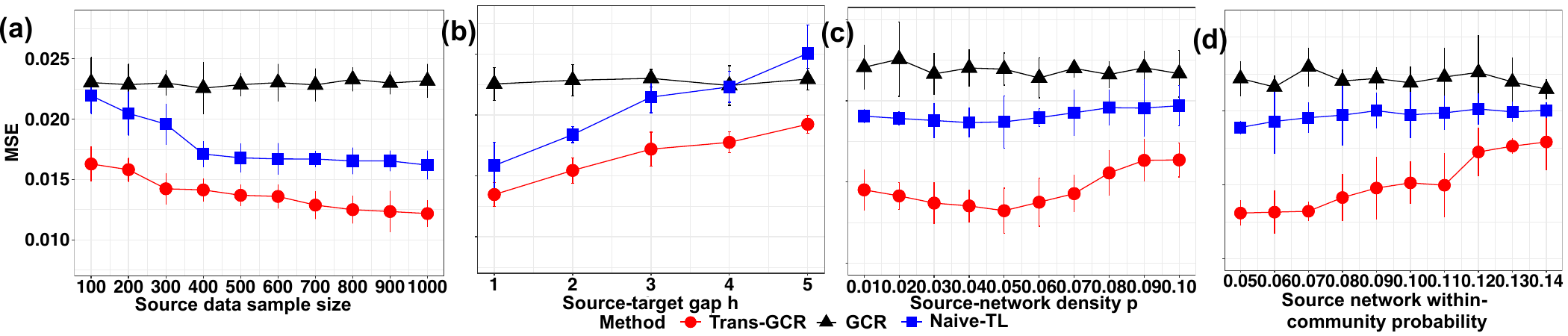}
     \caption{Performance comparison (MSE) of Trans-$\our$ (red), $\our$ (black), Naive TL (blue) across varying (a) Source sample size, (b) Source-target gap $h$, (c) Source network density (0.05 means identical densities) (d) Source network within-community probability (higher value means more discrepancy) when convolution layers  $M=2$.  }
    \label{fig:append-layer}
\end{figure}

We also show the performance comparisons when we consider multiple convolution layers such as $M=2$ in Figure \ref{fig:append-layer}.

\subsection{Additional Real Data Results}\label{sec:real_add}
\begin{figure}[!ht]
    \centering
    \includegraphics[scale=0.38]{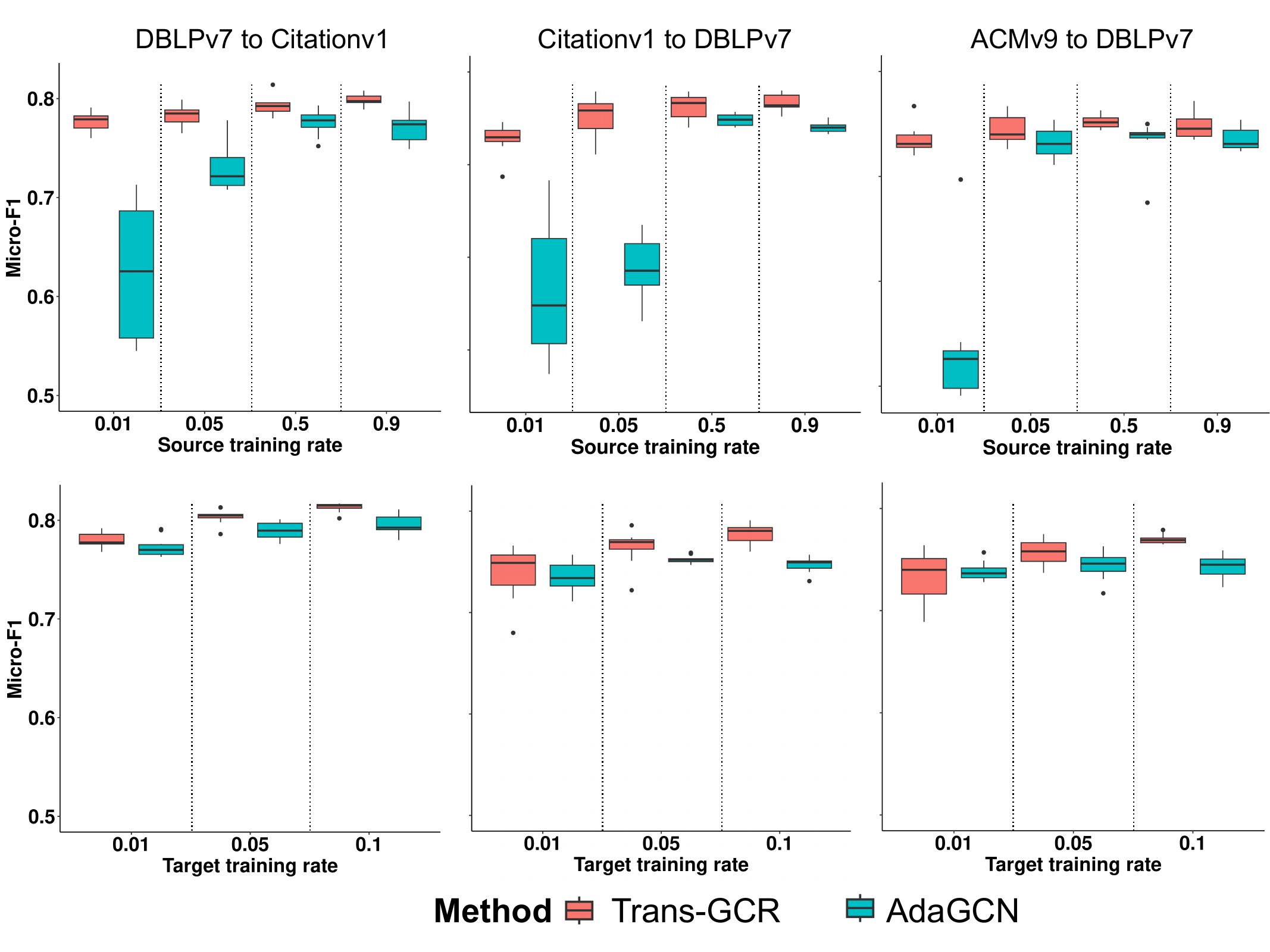}
\caption{Multi-label classification with varying source training rates (first column, with target training rate fixed to be 0.03), with varying target training rate (second column, with source training rate fixed to be 0.75).  } \label{fig:vary_rate} 
\end{figure}

\begin{figure}[!ht]
    \centering
    \includegraphics[scale=0.38]{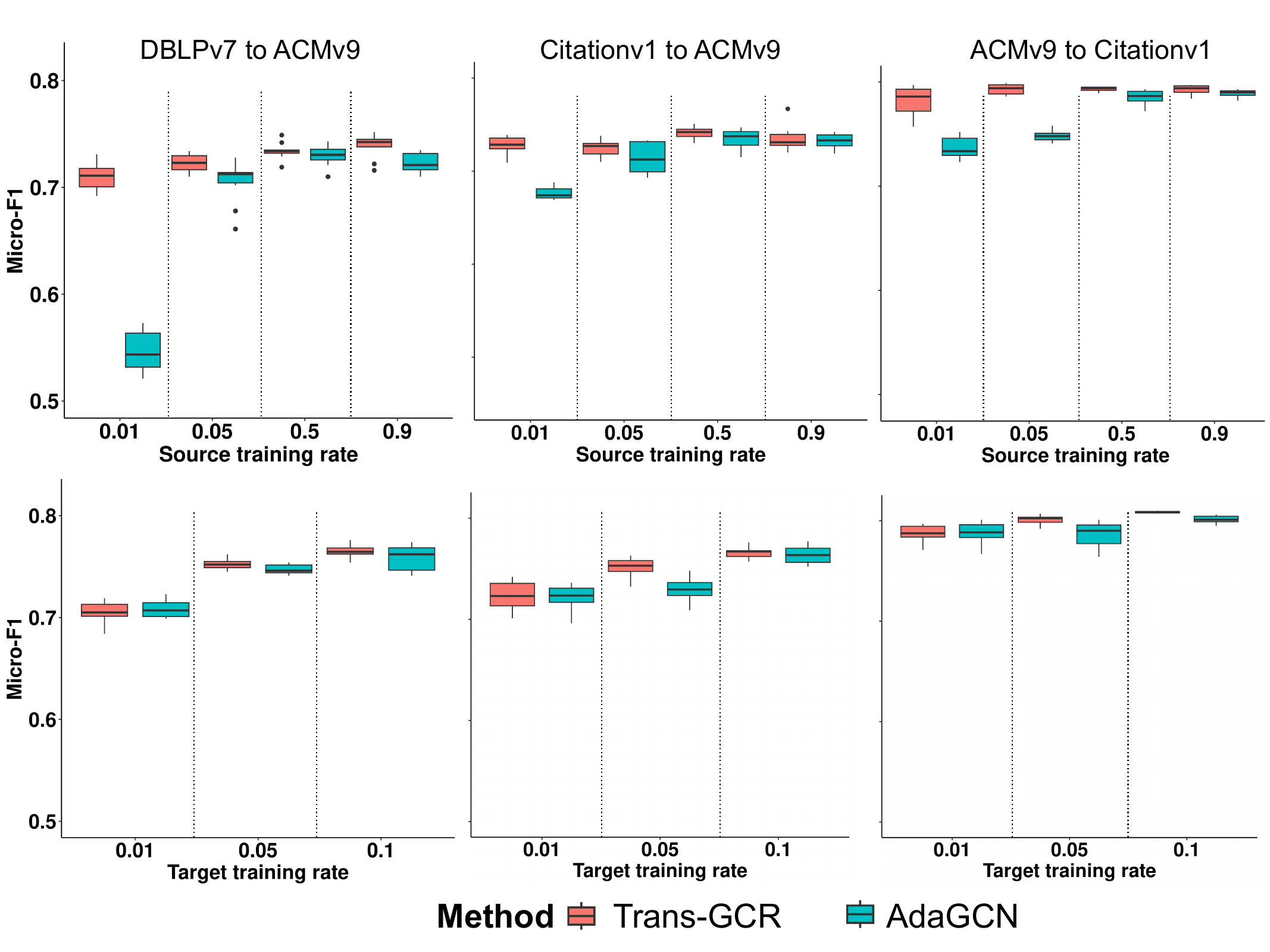}
\caption{Multi-label classification with varying source training rates (first column, with target training rate fixed to be 0.03), with varying target training rate (second column, with source training rate fixed to be 0.75).  } \label{fig:vary_rate2} 
\end{figure}

Here we present more detailed results from the real data analysis. Table \ref{Tab:S1} shows the averaged Macro-F1 scores. Table \ref{Tab:S2} presents the standard deviation (SD) across various methods and tasks, corresponding to the averaged Micro and Macro-F1 values we presented in Table \ref{Tab:1} and Table \ref{Tab:S1}. We also present the effect of source training rate and target training rate on Micro-F1 in Figure \ref{fig:vary_rate} for the transfer learning tasks D $\rightarrow$ C, C $\rightarrow$ D, and A $\rightarrow$ D, in Figure \ref{fig:vary_rate2} for the transfer learning tasks D $\rightarrow$ A, C $\rightarrow$ A, and A $\rightarrow$ C, respectively.

\begin{table}[!ht]
  \centering
  \caption{Averaged Macro F1 score of various methods, over 10 replicates,  with source training rate fixed at 0.75 and target training rate fixed at  0.03.}\label{Tab:S1}
  \resizebox{1\linewidth}{!}{
\begin{tabular}{cccccccccccc} 
\toprule 
Target & Source & \textbf{Trans-\our} & GCR & AdaGCN  & node2vec & GraphSAGE & GCN & APPNP & attri2vec & SGC & GAT  \\ 
\midrule
\multicolumn{12}{c}{Macro-F1 (\%)}                                                                                                                                                                                                                                                                                                                                                                                                                                                                                                                                                                                                                                 \\ \midrule 
\multirow{3}{*}{D} & C      & \textbf{73.69} & 69.55 & 73.53                                       & 60.77                        & 66.76                         & 66.92                   & 59.95                     & 63.48                         &        66.99             & 66.29                    \\
                   & A      & 72.94        & 66.95    & 73.37                                        & 57.97                        & 62.01                         & 62.76                   & 55.43                     & 58.08                         & 61.78                  & 61.97                    \\
                   & C\&A     & \textbf{73.74}& 68.89     & 73.61                                             & 49.13                        & 46.15                         & 56.73                   & 57.55                     & 57.54                         &67.13                     & 54.65                    \\ 
 \midrule
\multirow{3}{*}{C} & D      & \textbf{77.23} & 69.67   & 75.41                                     & 58.02                        & 66.02                         & 66.71                   & 62.61                     & 69.02                         & 66.61                  & 67.61                    \\
                   & A      & \textbf{78.49}   & 73.98    & 78.10                                    & 62.67                        & 67.69                         & 69.98                   & 63.13                     & 66.29                         &  75.56                  & 68.48                    \\
                   & D\&A     & \textbf{78.94} & 74.68   & 77.15                                               & 47.59                        & 54.30                         & 63.03                   & 62.09                     & 67.63                         & 75.53                   & 62.87                    \\ 
 \midrule
\multirow{3}{*}{A} & D      & \textbf{73.45} & 71.09  & 72.50                                       & 54.78                        & 62.95                         & 65.92                   & 55.92                     & 61.67                         &        63.49           & 66.30                    \\
                   & C      & \textbf{74.21}  & 72.03& 73.60                                       & 59.93                        & 65.30                         & 64.04                   & 56.18                     & 63.68                         &  72.01                & 67.47                    \\
                   & D\&C     & \textbf{75.01}& 72.23    & 74.03                                           & 44.56                        & 56.81                         & 60.21                   & 54.62                     & 62.93                         &        70.39            & 63.11                    \\

\bottomrule 
\end{tabular}}
\end{table}

\begin{table}[!ht]
  \centering
  \caption{Standard deviation for Micro and Macro F1 score of various methods, over 10 replicates,  with source training rate fixed at 0.75 and target training rate fixed at  0.03.}  \label{Tab:S2}
  \resizebox{1\linewidth}{!}{
\begin{tabular}{cccccccccccc} 
\toprule 
Target & Source & \textbf{Trans-\our} & GCR & AdaGCN  & node2vec & GraphSAGE & GCN & APPNP & attri2vec & SGC & GAT  \\ 
\midrule
\multicolumn{12}{c}{Micro-F1 (\%)}                                                                                                                                                                                                                                                                                                                                                                                                                                                                                                                                                                                                                                 \\  \midrule
\multirow{3}{*}{D} & C      & \textbf{<0.01}  & <0.01  & 0.01                                                     & 0.01                       & 0.01                 & 0.01                   & 0.02                         & 0.01                  & 0.02            & 0.01      \\
                   & A     & \textbf{<0.01} & <0.01  & 0.01                                                      & 0.01                       & 0.01                 & 0.02                  & 0.02                         & 0.01                  & 0.02             & 0.01      \\
                   & C\&A      & \textbf{0.01} & <0.01 & 0.02                                                    & 0.01                       & 0.01                 & 0.02                  & 0.01                       & 0.01                  & 0.02             & 0.03    \\
\midrule
\multirow{3}{*}{C} & D       & \textbf{<0.01} & <0.01    & 0.02                                                 & 0.01                       & 0.01                 & 0.01                 & 0.01                       & 0.01                  & 0.01             & 0.01  \\
                   & A      & \textbf{<0.01} & <0.01                              & 0.01                       & 0.01                       & 0.01                 & 0.01                 & 0.02                      & 0.01                  & 0.01             & 0.01  \\
                   & D\&A     & \textbf{<0.01}                               & <0.01  & 0.02                               & 0.02                      & 0.02                  & 0.03             & 0.01    & 0.01                       & 0.02                & 0.02  \\
\midrule
\multirow{3}{*}{A} & D      & \textbf{<0.01} & <0.01  & 0.01                                                      & 0.01                       & 0.01                 & 0.01                   & 0.01                         & 0.01                  & 0.01              & 0.01      \\
                   & C       & \textbf{<0.01}                                 & <0.01                & 0.01        & 0.01                       & 0.01                 & 0.01                   & 0.01                         & 0.01                  & 0.01              & 0.01      \\
                   & D\&C      & \textbf{<0.01}                                 & <0.01            & 0.01            & 0.01                       & 0.02                 & 0.04                   & 0.01                         & 0.01                  & 0.03              & 0.02     \\
 \midrule
\multicolumn{12}{c}{Macro-F1 (\%)}                                                                                                                                                                                                                                                                                                                                                                                                                                                                                                                                                                                                                                 \\ \midrule 
\multirow{3}{*}{D} & C      & \textbf{<0.01}                              & <0.01                & 0.01           & 0.01                       & 0.01                 & 0.01                   & 0.01                         & 0.01                  & 0.03             & 0.01      \\
                   & A      & \textbf{<0.01}                                   & <0.01              & 0.01        & 0.01                       & 0.01                 & 0.02                   & 0.02                         & 0.01                  & 0.04              & 0.02      \\
                   & C\&A     & \textbf{0.01}                                & <0.01            & 0.02             & 0.02                      & 0.01                 & 0.04                   & 0.01                         & 0.04                  & 0.02              & 0.03      \\
 \midrule
\multirow{3}{*}{C} & D     & \textbf{<0.01}                                & <0.01               & 0.02         & 0.01                       & 0.01                 & 0.01                   & 0.01                         & 0.01                  & 0.02              & 0.02      \\
                   & A      & \textbf{<0.01}                                  & <0.01     & 0.01                  & 0.01                       & 0.01                 & 0.02                   & 0.02                         & 0.01                  & 0.02              & 0.02      \\
                   & D\&A      & \textbf{<0.01}                                  & <0.01              & 0.02         & 0.04                       & 0.02                 & 0.06                   & 0.01                         & 0.01                  & 0.02              & 0.02      \\
 \midrule
\multirow{3}{*}{A} & D     & \textbf{<0.01}                               & <0.01                & 0.01         & 0.01                       & 0.01                 & 0.01                   & 0.01                         & 0.01                  & 0.01             & 0.01      \\
                   & C       & \textbf{<0.01}                                 & <0.01      & 0.02                & 0.01                       & 0.01                 & 0.01                   & 0.01                         & 0.01                  & 0.01              & 0.01      \\
                   & D\&C    & \textbf{<0.01}                                 & <0.01             & 0.01          & 0.03                      & 0.02                 & 0.07                   & 0.01                         & 0.01                  & 0.04              & 0.02      \\

\bottomrule 
\end{tabular}}
\end{table}

\clearpage

\section{Pseudo Code for Algorithms}\label{sec:algorihtm_add}
Here, we summarize the procedure of Trans-$\our$ in Algorithm \ref{al:trans_alg}, and the procedure of source domain selection in 
Algorithm \ref{al:source_selection_alg}.

\begin{algorithm}[h]
\caption{Trans-$\our$ Algorithm}
\begin{algorithmic}
\Require Target data $(\bA^{(0)}, \bX^{(0)},\bY^{(0)})$;  Source data $(\bA^{(k)}, \bX^{(k)},\bY^{(k)}), k\in \mA$; Hyperparameter $M$ and $\lambda$.
\State Step 1. \textbf{Preprocessing.}
  Calculate the normalized adjacency matrix $\bS^{(k)}$ based on $\bA^{(k)}$, $k \in \{0, \mA\}$.
\State Step 2. \textbf{Pooled source samples.}
     Get pooled source sample with $\bS^{\mathcal{A}} \in \mathbb{R}^{n_{\mA} \times n_{\mA}}$, 
$\bX^{\mathcal{A}} \in \mathbb{R}^{n_{\mA} \times \dims}$, 
$\bY^{\mathcal{A}} \in \{0,1\}^{n_{\mA} \times C}$.
\State {Step 3.}
\textbf{Source domain parameter estimation.} Get $\hat{\bbeta}^{\mathcal{A}}$ by  Eq. \ref{eq:mle}, using the  pooled source samples $(\bS^{\mathcal{A}},\bX^{\mathcal{A}},\bY^{\mathcal{A}})$.
\State Step 4. \textbf{Domain shift estimation}. Obtain domain shift estimate  $\hat\delta^\mathcal{A}$ using Eq. \ref{eq:delta_esti} and target data. 
\State Step 5. \textbf{Target domain parameter estimation.}
Obtain $\hat\bbeta^{(0)}=\hat\bbeta^\mathcal{A}+\hat\delta^\mathcal{A}$. \\
\State \textbf{Output}: $\hat\bbeta^{(0)}$.
\end{algorithmic}\label{al:trans_alg}
\end{algorithm}

\begin{algorithm}[!ht]
\caption{Source Domain Transferability Score Calculation Algorithm }
\begin{algorithmic}
\Require Data: Target data $(\bA^{(0)}, \bX^{(0)},\bY^{(0)})$;  Source data $(\bA^{(k)}, \bX^{(k)},\bY^{(k)}), k=1,\ldots,K$; \\
Hyperparameters: Number of layers $M$; Cross-validation folds $V$; Number of selected source data $L$. \
\State \textbf{Step 1.}  \textbf{Target data partition.}  Randomly partition data points in the target domain $\{1, \ldots, n_0\}$ into $V$ subsets of approximately equal size $s_1, \ldots,s_V$.
\State \textbf{Step 2.}  \textbf{Training and testing target data construction.} 
        Construct testing target data $(\bA^{(0)},\bX^{(0)}$, $\bY_{s_v}^{(0)})$, where we only use the label information of nodes in $s_v$. Similarly, construct training target data $(\bA^{(0)}$, $\bX^{(0)}$, $\bY_{-s_v}^{(0)})$ by excluding the label information of nodes in  $s_v$.      
\State \textbf{Step 3.} \textbf{Cross-validation based score.} For $k$th source data, $k=1,\ldots, K$, repeat the following procedure. 
      \State\hskip 15pt   \textbf{For $v=1,\ldots, V$,} 
    \begin{itemize}
        \item \textbf{Model estimation.} Apply the transfer learning Algorithm \ref{al:trans_alg} using the source data $\{\bA^{(k)}, \bX^{(k)},  \bY^{(k)}\}$  and training target data
        $\{ \bA^{(0)}, \bX^{(0)}, \bY_{-s_v}^{(0)}\}$ to obtain the estimate $\hat{{\bbeta}}_{vk}^{(0)}$ for the target data after transfer learning. 
        \item \textbf{Model evaluation}  Using the learned $\hat{{\bbeta}}_{vk}^{(0)}$ from the previous step, evaluate its prediction performance in the
        target domain testing data $\{ \bA^{(0)}, \bX^{(0)}, \bY_{s_v}^{(0)}\}$ by calculating the negative log-likelihood value $\mbox{NL}_v^{(k)}$ for nodes in $s_v$.
   \end{itemize}
    \State\hskip 15pt   \textbf{Averaged score over folds.}  Calculate averaged negative log-likelihood  over $V$ folds for each $k$ source data, $\mbox{NL}^{(k)} = \frac{1}{V}\sum_{v=1}^{V}\mbox{NL}_v^{(k)}$
\item \textbf{Step 4.} \textbf{Selection.} Rank the $K$ sources according to $\mbox{NL}^{(k)}$ and select among the top $L$ lowest  sources as $\hat{\mathcal{A}}$.
\State \textbf{Output}: $\hat{\mA}$ and  
Transferability score $\mbox{NL}^{(k)}$, $k=1,\ldots,K$.
\end{algorithmic}\label{al:source_selection_alg}
\end{algorithm}

\clearpage

\end{document}

%% file: Theory.tex
\section{Theoretical Properties}
\label{sec:theory}

In this section, we present our main theoretical results. 
Recall that in this paper, we proposed a $\our$ model to account for the network dependency between observations, and a transfer learning method Trans-$\our$ to promote the estimation of the target data utilizing knowledge from source data.  Establishing theoretical guarantees for the GCR model under network dependency alone is extremely challenging as our data i) is high-dimensional (i.e., the number of covariates can be much larger than the number of samples) and ii) admits a network dependency among the observations. 
This dependency prevents the application of standard concentration results, valid only for i.i.d. data. 
Consequently, a careful mathematical analysis is necessary. 
In this paper, we take a step towards that goal by rigorously analyzing the theoretical properties of $\hat{\bbeta}$ 
obtained by minimizing the loss function in Eq. \ref{eq:GCMLR}
 when  $\bA$ is generated from an Erd\H{o}s--R\'enyi (ER) random graph model with parameter $p$, $M=1$ (i.e., one convolutional layer),
 given observed data $(\bA, \bX, \bY)$.  For simplicity and ease of presentation, this section focuses on results for the case where $C=2$ corresponds to a two-class classification problem.
Nevertheless, our theoretical results can be easily generalized to the multiclass cases. Under ER model, the expected degree of each node is $np$. In what follows, we build theoretical guarantees under the normalized adjacency matrix $\bA\bX/\sqrt{np}$. 
To ease presentations, we let
$\bZ = (\bA \bX)/\sqrt{np}$ (replace $p$ by $\hat p$ when uknown). 
We would like to highlight that rows of $\bZ$ are \textit{not} independent. 
Let $\|{\bbeta}^{(0)}\|_0 = s$ for some $s \ll d$. The estimate 
$\hat{\bbeta}$ is obtained by minimizing the loss function in Eq. \ref{eq:GCMLR}, which can further rewritten using $\bZ$, as shown below, 
\begin{equation}
\textstyle
    \label{def:beta_hat}
    \hat \bbeta = \argmin_{\bbeta} \left\{ - \sum_{i = 1}^n \left[\bY_i \bZ_{i}^\top \bbeta - \log{(1 + \exp{(\bZ_{i}^\top \bbeta)})}\right] + \lambda \|\bbeta\|_1\right\},
\end{equation}
 where $\bZ_{i}$ is the $i$th row of $\bZ$. 
The key difference between our analysis of $\hat \bbeta$ and that of a standard high dimensional generalized linear model (e.g., \cite{van2014asymptotically}) is precisely the dependence among observations. 
Below, we present the assumptions required to establish the theoretical guarantees of our penalized estimator $\hat \bbeta$: 
\begin{assumption}
    \label{assm1:dist_X}
    We assume $\bX_i$'s are generated independently from a sub-gaussian distribution with parameter $\sigma_\bX$, i.e., $\bbE[\exp{(\lambda a^\top \bX_i)}] \le \exp{(\lambda^2\sigma_\bX^2/2)}$ for all $\lambda \in \reals$ and for all $a \in S^{p-1}$. Let $\Sigma_\bX$ be the covariance matrix of $\bX_i$. Assume that $\lambda_{\min}(\Sigma_\bX) \ge \kappa_l > 0$ and $\max_j \Sigma_{\bX, jj} \le \sigma_+$ for some constant $(\kappa_l, \sigma_+)$. 
\end{assumption}

\begin{assumption}[Network connectivity]
    \label{assm:network_connection}
    The network connectivity parameter $p$ of the Erd\H{o}s--R\'enyi random graph $\bA$ satisfies $(np)/\log{n} \to \infty$, and $p\log{d} \to 0$ as $n \rightarrow \infty$. 
\end{assumption}

\begin{assumption}[Sparsity]
    \label{assm:sparsity}
    The sparsity parameter $s$ of ${\bbeta}^{(0)}$ satisfies: 
        $\textstyle 
        s\frac{\log^2{d}}{n}\frac{(1-2p)}{4p\log{\left((1-p)/p\right)}} = o(1)$.
\end{assumption}
All three assumptions above are quite standard in the literature on the analysis of high-dimensional linear and generalized linear models. Assumption \ref{assm1:dist_X} requires the covariates to be sub-gaussian, a ubiquitous assumption, which is often needed to deal with ultra-dimensional setup (i.e., $d$ can be as large as $e^{n^\gamma}$ for some $\gamma < 1$) (see Chapter 6 of \cite{buhlmann2011statistics} for details). One may relax this assumption as the cost of the trade-off between $d$ and $n$; the thicker the tail of $\bX$, the more stringent condition is needed on $d$ for estimation, or one may use robust loss function (\cite{goldsmith2015lasso}). Assumption \ref{assm:network_connection} is also a standard assumption in the literature of network estimation, which merely assumes $p$ to be (slightly) larger than $n^{-1}$ in order \citep{lei2015consistency}. As for Assumption \ref{assm:sparsity}, in standard i.i.d. setup we require $s\log{d}/n \to 0$. However, here, our condition is slightly different; we have an additional factor involving a factor of $p$. Whether this dependence is optimal is out of the scope of this paper. Given these assumptions, we are now ready to state our main theorem: 
\begin{theorem}
\label{thm:main}
    Under Assumption \ref{assm1:dist_X}-\ref{assm:sparsity}, the $\ell_1$-penalized estimator $\hat \bbeta$ obtained in \eqref{def:beta_hat} satisfies; 
    \begin{equation*}
    \textstyle
    \|\hat \bbeta - {\bbeta}^{(0)}\|_2^2 \le c \frac{s\log{d}}{n}
    \end{equation*}
    for some constant $c > 0$ with probability $1 - g(n, p, d)$, where $g(n, p, d)$ goes to $0$ as $n \rightarrow \infty$, mentioned explicitly in the proof. 
\end{theorem}
The proof of the theorem is deferred to the Appendix. 
One of the main technical challenges lies in establishing a condition equivalent to the standard restricted strong convexity (RSC) or restricted eigenvalue (RE) condition in the presence of network dependency.
This task is difficult even in an i.i.d. setup; see \cite{raskutti2010restricted, zhou2009restricted, rudelson2012reconstruction, negahban2009unified} for related references. The modified RSC condition is presented in Lemma \ref{lem:RSE_GLM}, which is of independent interest.
\begin{remark}
    It may seem surprising that the convergence rate of $\hat \bbeta$, as established in Theorem \ref{thm:main} does not depend on $p$. This is precisely because we have appropriately scaled $\bZ = (\bA\bX)/\sqrt{np}$ by $p$. If we change the scale, i.e., say we use symmetric normalized Laplacian $\bD^{-1/2}\bA \bD^{-1/2}$, then the effective calling would be $\bZ = (\bA\bX)/np$. In that case, the estimation rate of $\hat \bbeta$ would depend on $p$, and this dependence can be easily quantified by carefully tracking the steps of our proof. 
    However, as this involves technical algebra without adding further insight, we refrain from pursuing it here.
\end{remark}

\begin{remark}
   Our theorem precisely quantifies the estimation rate of $\bbeta^{(0)}$ in a single domain, which can be viewed as the estimation rate using only data from the target domain. Recently, estimating $\bbeta^{(0)}$ using related source samples under the high-dimensional generalized linear model setup has been explored by \cite{tian2023transfer, li2023estimation}. As previously mentioned, extending these ideas to incorporate network dependency is quite challenging. However, we have conjectured an estimation rate for $\bbeta^{(0)}$ obtained using Trans-$\our$. See Appendix \ref{sec:thm_mult_source} for details. 
\end{remark}

\begin{remark}
    Although in our theory we have assumed all the entries of $\bA$ are generated independently from $\Ber(p)$, Theorem \ref{thm:main} continues to hold under the self-loop (resp. no self-loop), i.e. if we set $\bA_{ii} = 1$ (resp. $\bA_{ii} = 0$) and generate off-diagonal elements from $\Ber(p)$, with different constant $c'$ instead of $c$. In the proof, we have pointed out the required modifications precisely to adapt proof when i) we have self-loop, and ii) $p$ is unknown and estimated from the data.  
\end{remark}